\documentclass[11pt]{elsarticle}
\usepackage{bm}
\usepackage[margin=1in,papersize={8.5in,11in}]{geometry}
\usepackage{times,xcolor,hyperref}
\usepackage[export]{adjustbox}
\usepackage{array}
\usepackage{amssymb,amsfonts,amsmath,amsthm, mathrsfs,mathtools}
\usepackage{caption}
\usepackage{graphicx}
\usepackage{multicol}
\usepackage{multirow}
\usepackage{nomencl}
\makenomenclature

\usepackage{etoolbox}
\usepackage{booktabs}
\usepackage{subcaption}
\usepackage[most]{tcolorbox}

\usepackage{floatrow}
\DeclareFloatFont{small}{\small}
  \newcommand{\R}{\mathbb{R}}
  
  \newcommand{\layerOutvec}{\bm{r}}
  \newcommand{\layer}{\bm{\mathcal{F}}}
  
  \newcommand{\Wparm}{\bm{W}}
  \newcommand{\bparm}{\bm{b}}
  \newcommand{\params}{\bm{\Theta}}
  \newcommand{\outwts}{\bm{A}}
  \newcommand{\Cbeta}{\frac{1}{(2\beta)^{N_x}}}
  
  \newcommand{\Cexpmarg}{\frac{1}{2\pi}}
  
  \newcommand{\betacube}{[ -\beta, \beta ]^{N_x}}
  \newcommand{\neu}{{\phi}}
  \newcommand{\mean}{\bm{\mu}}
  \newcommand{\noise}{\bm{z}}

\journal{ArXiv}

\renewcommand{\textcolor}[2]{#2}

\begin{document}

\begin{frontmatter}

\title{Uncertainty propagation in feed-forward neural network models}

\author[author]{Jeremy Diamzon}
\ead{jdiamzon@ucsc.edu}

\author[author]{Daniele Venturi\corref{correspondingAuthor}}
\cortext[correspondingAuthor]{Corresponding author}

\ead{venturi@ucsc.edu}
\address[author]{Department of Applied Mathematics, UC Santa Cruz, Santa Cruz, CA 95064}

\begin{abstract}
We develop new uncertainty propagation methods for feed-forward neural network architectures with leaky ReLU activation functions subject to random perturbations in the input vectors.  In particular, we derive analytical expressions for the probability density function (PDF) of the neural network output and its statistical moments as a function of the input uncertainty and the parameters of the network, i.e., weights and biases. A key finding is that an appropriate linearization of the leaky ReLU activation function yields accurate statistical results even for large perturbations in the input vectors. This can be attributed to the way information propagates through the network.  We also propose new analytically tractable Gaussian  copula surrogate models to approximate the full joint PDF of the neural network output. To validate our theoretical results, we conduct Monte Carlo simulations and a thorough error analysis on a multi-layer neural network representing a nonlinear integro-differential operator between two polynomial function spaces. Our findings demonstrate excellent agreement between the theoretical predictions and Monte Carlo simulations.
\end{abstract}

\end{frontmatter}

\section{Introduction}

In the past decade, the field of Scientific Machine Learning (SciML) has made remarkable advancements influencing a wide range of disciplines, including physics, engineering, and medicine \cite{abdar_review_2021,gawlikowski_survey_2023,Panos}.
However, despite these advancements,  an effective framework to assess the reliability and trustworthiness of solutions generated by neural network models has lagged behind their rapidly evolving capabilities. This gap is particularly pronounced in applications involving partial differential equations (PDEs) \cite{Raissi,GK2020}, and operator mappings 
between infinite-dimensional function spaces \cite{psaros_uncertainty_2023, zou_uncertainty_2025,VenturiSpectral,venturi2018numerical,NNFDE2024}, where even 
small uncertainties in input variables or parameters can be amplified by 
the neural network, resulting in inaccurate predictions.  
%
%
%
Uncertainty in neural network models may be categorized into {\em aleatoric} uncertainty, which arises from inherent variability or noise within the data, and {\em epistemic} uncertainty, which stems from model limitations or insufficient knowledge \cite{abdar_review_2021,gawlikowski_survey_2023,VenturiBook}. Quantifying the effects of these uncertainties is critical in fields where reliable predictions are essential, and the consequences of inaccurate outcomes can be severe. Nonetheless, despite significant progress in incorporating uncertainty quantification (UQ) methods into data-driven and physics-informed learning frameworks, their effective integration into the broader field of  SciML is still in its early stages.

In this paper,  we develop new uncertainty propagation methods for feed-forward neural network architectures representing a nonlinear operator between two function spaces \cite{zou_uncertainty_2025}. We write such operator as 
\begin{equation}
g(y) = \mathcal{N}\left(f(x)\right), \label{eq:intdifop}
\end{equation}
and assume that $f(x)$ is a random function with given statistical properties. 
The goal is to study the statistical properties of the output function $g(y)$, such as its mean, covariance, and multi-point probability density functions (PDFs), when the operator $\mathcal{N}$ is approximated using a {\em deterministic} deep neural network with leaky ReLU activation functions\footnote{\textcolor{blue}{We choose the leaky ReLU activation function over the standard ReLU to avoid the formation of point masses (atoms) at zero in the output of each layer. Nonetheless, the theoretical framework developed in this work is general and can be applied to standard ReLU activation function as well.}}. 
\textcolor{blue}{In contrast to UQ frameworks that introduce randomness in the neural network model of \( \mathcal{N} \) or its output \( g(y) \) while treating the input \( f(x) \) as deterministic, we consider a deterministic neural network subject to uncertainty in the input function \( f(x) \), and characterize the resulting uncertainty in the output through deterministic propagation, i.e., we study the ``forward'' UQ problem \cite{DongbinBook}.}
\textcolor{blue}{A real-world application of this setting arises, for example., in climate modeling. Modern climate simulations rely on complex and computationally intensive physical models to predict variables such as temperature, precipitation, and sea level over large spatial 
and temporal scales. To reduce computational cost, surrogate models based on deep neural networks are increasingly used to approximate the input--output mappings of these physical systems \cite{Ashesh1,Ashesh2}. While the neural networks are trained deterministically, the input data such as solar radiation, greenhouse gas concentrations, and ocean surface temperatures often contain significant uncertainty due to measurement errors, limited observational coverage, and inherent variability. In this setting, solving the forward UQ problem for climate is essential for characterizing how input uncertainty propagates through the neural network and affects the predicted outputs. Rather than modeling uncertainty in the neural network itself (i.e., model uncertainty), the objective is to understand the distribution of the output as induced by uncertain but structured input distributions. This enables sensitivity analysis and risk assessment, such as quantifying the impact of uncertain ocean temperature fields on predicted precipitation variability, thereby providing valuable guidance for climate policy and decision-making.}

\begin{figure}[!t]
  \centering
  \begin{subfigure}[c]{0.40\textwidth}
  \vspace{-1.2cm}
      \caption{Linearization of leaky ReLUs\vspace{0.5cm}}
      \hspace{0.5cm}\vspace{-0.5cm}
  \includegraphics[width=0.9\hsize]{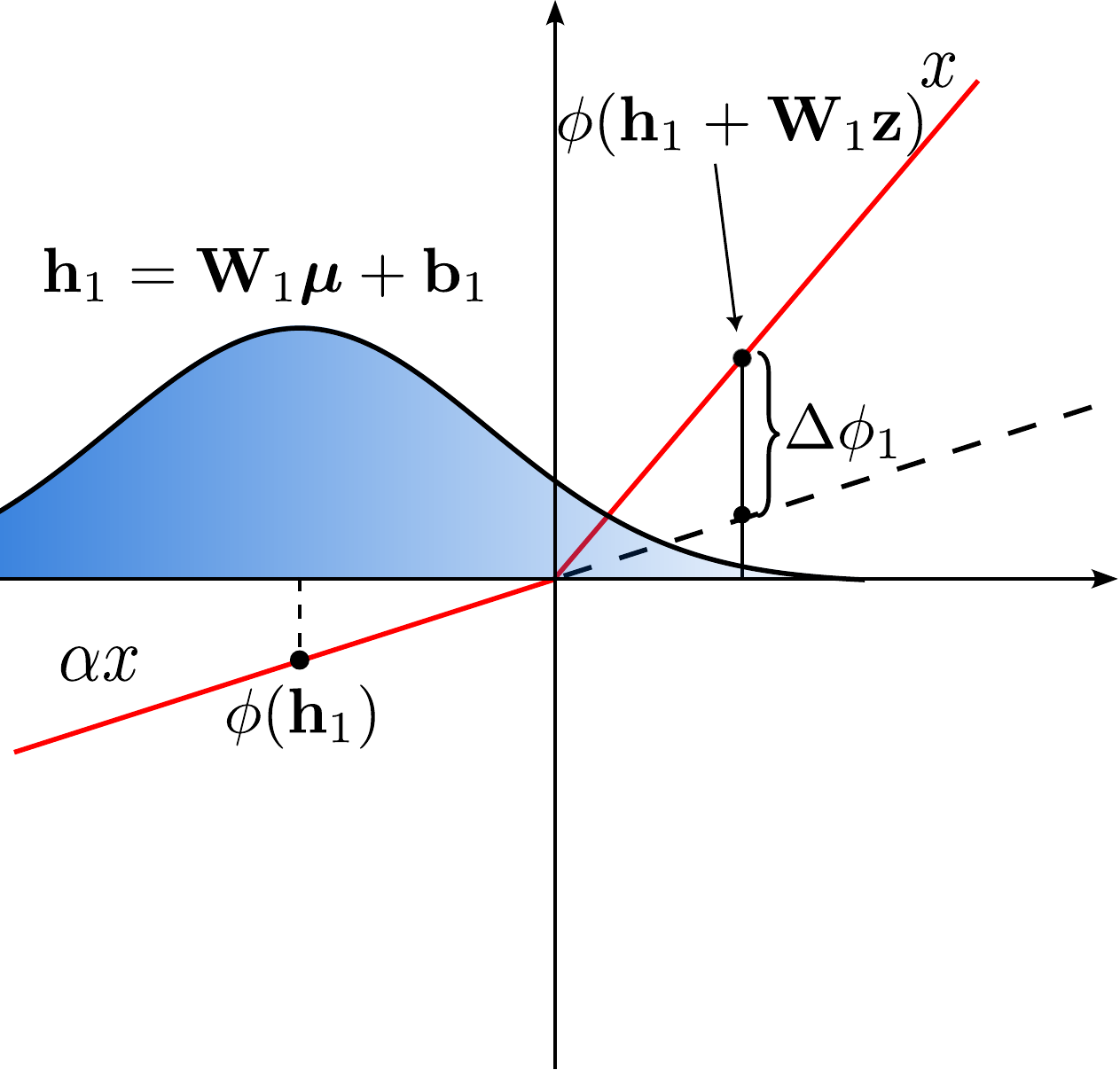}
  \end{subfigure}
  \quad
  \tikz[baseline]{
    \node[inner sep=0pt, anchor=base] (X) {
      \hspace{1ex}
      \tikz{\draw[ultra thick,->](0,0)--(1,0);}
    };
    \useasboundingbox (X.base);
  }
  \hspace{1.0ex}
  \begin{subfigure}[c]{0.40\textwidth}
     \caption{PDFs of the neural network output}
    \includegraphics[width=\hsize]{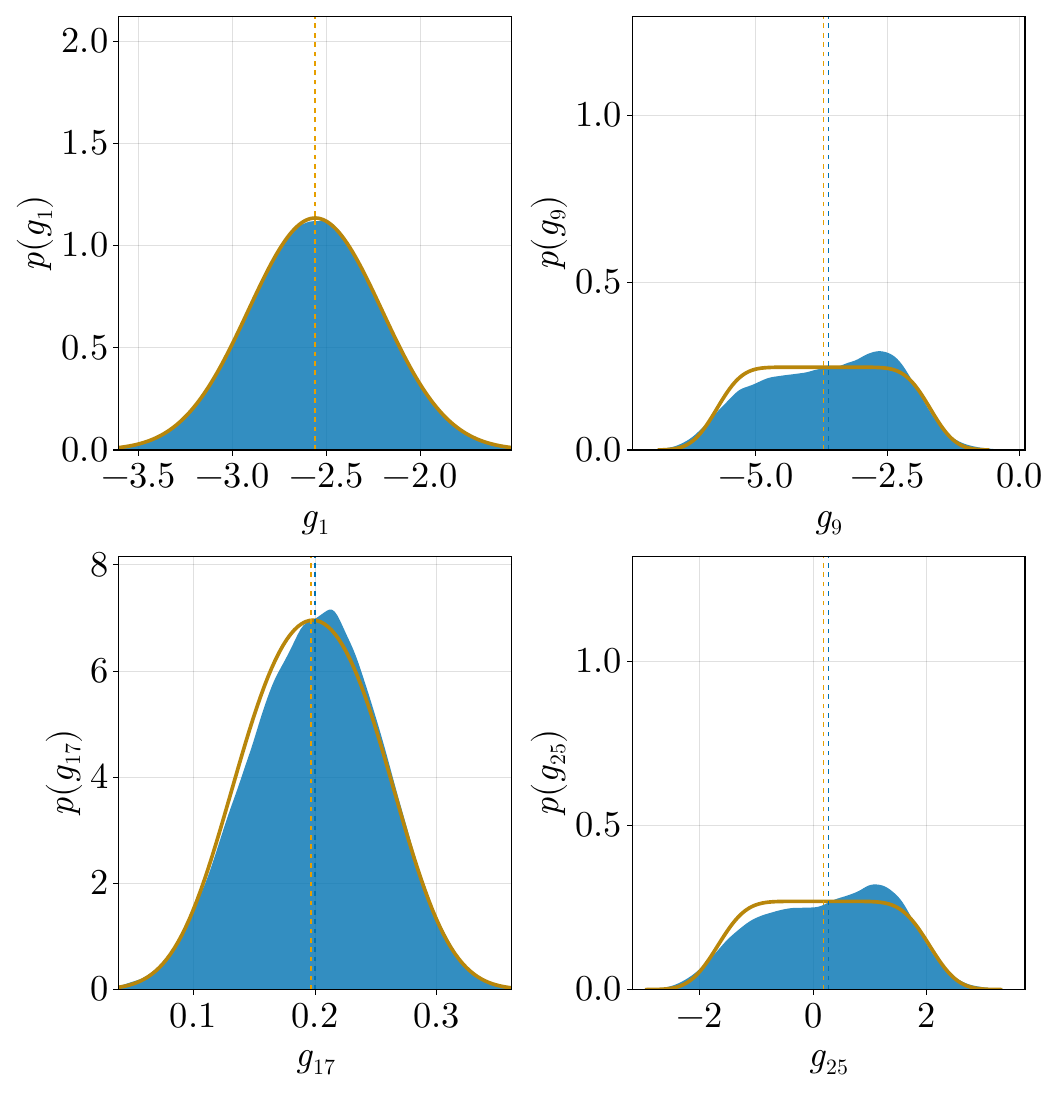}
  \end{subfigure}
     \caption{Illustration of the linearization of the leaky ReLU activation function $\phi(\cdot) $ for a one-layer neural network. The neuron input \( \bm h_1 = \bm W_1\bm \mu+\bm b_1 \) is perturbed to \( \bm h_1 + \bm W_1\bm z \), where \(\{\bm{W}_1,\bm b_1\} \) are weights and biases of the first layer, and \( \bm{z} \) is vector representing the random perturbation of the input function  \( f(x) \) (see Eq. \ref{f}). The PDF of $\bm h_1 + \bm W_1\bm z $  is sketched as a Gaussian in Figure. The leaky ReLU function is linearized at \( \bm h_1 \). Its extension to the positive real axis is represented by the black dashed line. If the samples of  $\bm h_1 + \bm W_1\bm z \rangle$ are negative, there is no difference between the output of the linearized leaky ReLU and the ReLU function. However, if the samples of $\bm h_1 + \bm W_1\bm z $ are positive then the linearized leaky ReLU approximation introduces and error \( \Delta_1 \phi (\bm h_1)\). The cumulative effect of these errors tends to cancel out for deep nets as they propagate through the network, preserving accuracy in the one-point PDFs of the neural network output function $g(y)$. \textcolor{blue}{The image on the right compares the PDF of the neural network output \( g \) evaluated at specific spatial locations \( y_i \), i.e., \( p(g_i) \), obtained using the proposed linearized leaky ReLU approximation (yellow line) and Monte Carlo simulation (blue shaded region) for the prototype nonlinear integro-differential operator \eqref{nonlinear} . Details about the specific simulation settings, e.g., perturbation amplitudes, number of layers, and other neural network parameters can be found in Section \ref{sec:results}. }}
     \label{fig:NonlinPerturbComparison}
\end{figure}

A key finding of our study is that a ``linearization'' of the leaky ReLU activation function in the deep net representing \eqref{eq:intdifop} allows us to obtain {\em computable analytical expressions} for the PDF of the output function $g(y)$ and its statistical moments (see Figure \ref{fig:NonlinPerturbComparison}). Such analytical expressions turn out to be accurate for {\em large perturbations} in the input function $f(x)$. 
Furthermore, we show that multi-point PDFs of the output function $g(y)$ can be accurately approximated by a {\em Gaussian copula} \cite{Czado}, which can be constructed based on the analytical formulas we obtain for the marginal distribution of $g(y_i)$ (one-point PDFs), and the correlation 
function $\mathbb{E}\{g(y_i)g(y_j)\}$. In other words, the uncertainty 
in the output function $g(y)$ can be rigorously quantified using analytical expressions for its PDFs and statistical moments, for any given fully connected feed-forward neural network, and a statistical characterization of the random input function $f(x)$.

This paper is organized as follows. In Section \ref{sec:problemformulation} we formulate the operator approximation problem, including the discretization of the input-output functions and the mapping between the input and the output. In Section~\ref{sec:methodology}, we model fully connected feed-forward neural networks as discrete dynamical systems and set the mathematical foundations for our approximation methods. In Section~\ref{sec:PDFoutput}  we derive analytical expressions for the PDF of the neural network output. In Section~\ref{sec:moments}, we extend the analysis to compute statistical moments such as the mean, variance, and covariance of the network output. To obtain these expressions, we introduce a suitable linearization of the leaky ReLU activation function, which makes the analytical formulas for both the PDF and the moments tractable and computable. 
In Section~\ref{sec:copula} we introduce a new Gaussian copula model to approximate the full joint distribution of the network output based on the derived analytical expressions for the marginal PDFs and the correlation function. \textcolor{blue}{In Section \ref{sec:comparison} we compare the uncertainty propagation framework developed in this paper to existing UQ methods.} In Section \ref{sec:results} we present numerical results that validate our theoretical predictions across a range of scenarios, including both linear and nonlinear operators. The main findings of this study are summarized in Section \ref{sec:summary}. Additional technical details, including the derivation of the PDF of the neural network output for linearized leaky ReLU activations functions and an error analysis for linearized leaky ReLU networks are presented in  \ref{app:PDF} and \ref{app:error}, respectively. \textcolor{blue}{In \ref{app:ResNets} we show how to apply the uncertainty propagation framework developed in this paper to a neural network architecture with skip connections, specifically ResNet \cite{ResNets}.}
 \begin{figure}[t]
   \centerline{   \includegraphics[scale=0.66]{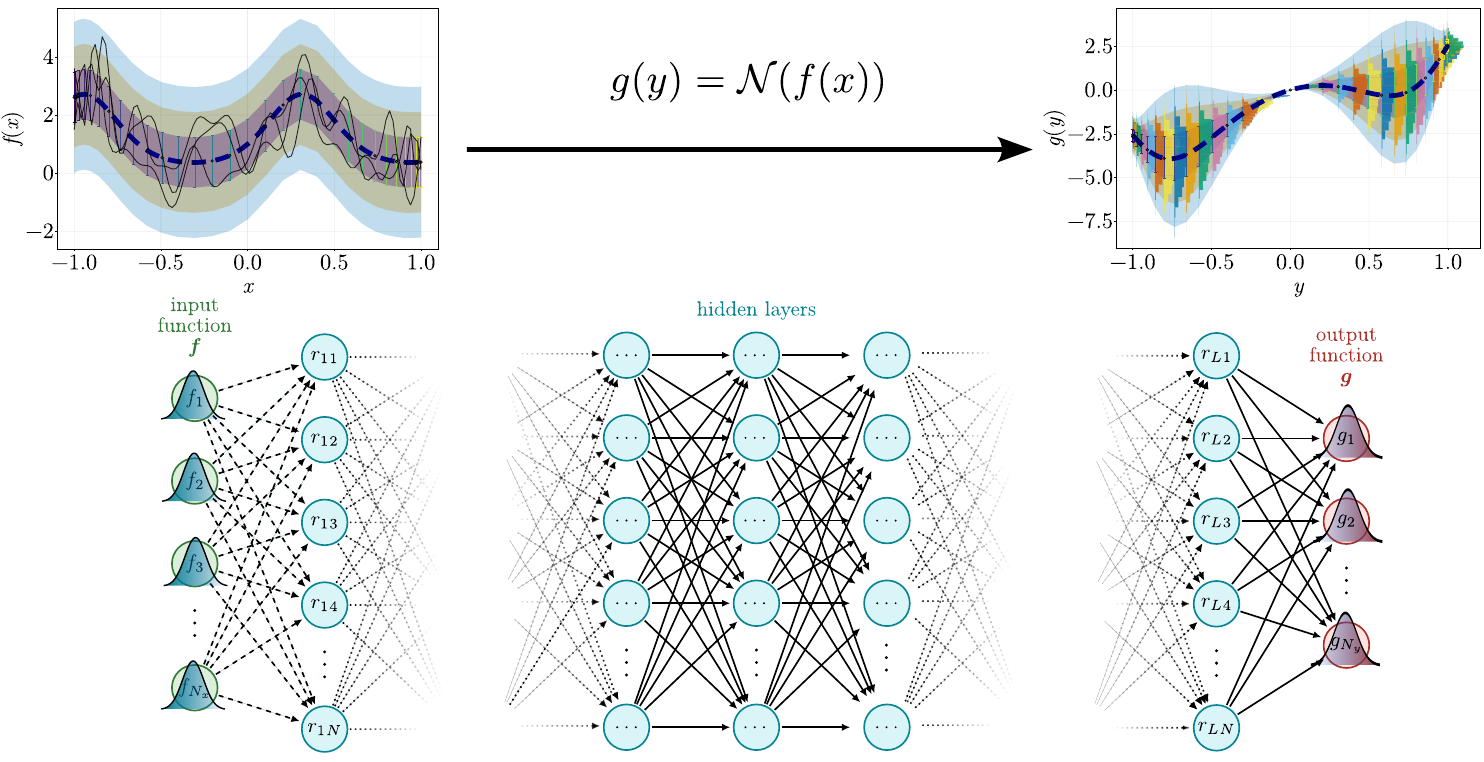}}
     \caption{Schematic of the feed-forward neural network approximating the non-linear (non-local) operator $g(y)=\mathcal{N}(f(x))$. The input function is assumed to be random, with known probability distribution. This makes the output function $g(y)$ random.  The goal is to study the statistical properties of $g(y)$ given statistical information on $f(x)$ and the parameters defining the neural net. \textcolor{blue}{The plots at the top illustrate how uncertainty in the random input function \( f(x) \) propagates to uncertainty in the output function \( g(y) \) via the deterministic nonlinear operator \( \mathcal{N} \). The neural network sketch below shows a discretized version of this setup, where \( f(x) \) is represented by the random input vector \( \bm{f} \), which is mapped by a deterministic MLP to the random output vector \( \bm{g} \).}}
     \label{fig:ProbForm}
\end{figure}

\section{Problem Formulation}
\label{sec:problemformulation}

We are interested in computing the statistical properties, e.g. statistical moments and PDF of the output, of a neural network approximating the nonlinear operator \eqref{eq:intdifop}, given statistical information on the input. 
To this end, we first select two appropriate polynomial 
function spaces defined on two intervals that are large enough 
to accommodate all input-output function pairs $\{f(x),g(y)\}$ 
of interest. Such input/output functions are then 
represented as Lagrangian interpolants at shifted 
Gauss-Legendre-Lobatto \cite{Hesthaven} 
grid points  $\{x_1,\ldots, x_{N_x}\}$ 
and  $\{y_1,\ldots, y_{N_y}\}$. This reduces the 
input-output function pairs $\{f(x),g(y)\}$ 
to a set of input-output vectors 
\begin{equation}
\bm f = \begin{bmatrix}
f_1 & \cdots & f_{N_x}
\end{bmatrix}, 
\qquad  
\bm g = \begin{bmatrix}
g_1 & \cdots & g_{N_y}
\end{bmatrix},
\label{fg}
\end{equation}
with components
\begin{equation}
f_i = f(x_i) \quad (i=1,\ldots, N_x) \qquad \text{and }\qquad g_j = g(y_j)\quad (j=1,\ldots, N_y). 
\end{equation}
This reduces the problem of learning a nonlinear operator 
between two Banach spaces to the problem of learning a nonlinear map between two finite-dimensional vector spaces, i.e., a map from $\mathbb{R}^{N_x}$ to $\mathbb{R}^{N_y}$.
In what follows, it is convenient to model the random input vector \( \bm{f} \) as the sum of a mean vector \( \bm{\mu} \) and a superimposed zero-mean random perturbation \( \bm{z} \), i.e.,
\begin{equation}
\bm f= \bm \mu + \bm z.
\label{f}
\end{equation} 
The perturbation \( \bm{z} \) may or may not be small and requires a probabilistic description. For instance, \( \bm{z} \) may be modeled as a zero-mean jointly Gaussian vector with a specified covariance matrix, or as a vector with independent and identically distributed (i.i.d.) components following a uniform distribution on the interval \( [-\beta, \beta] \). In the latter case, the joint PDF of $\bm z$ is given by
\begin{equation}
p(\bm z) =
 \begin{cases}
\displaystyle \frac{1}{(2\beta)^{N_x}} & \text{if} \quad \bm z \in [-\beta,\beta]^{N_x}\vspace{0.2cm}\\
0 &  \text{otherwise}
\end{cases}
\label{PDF}
\end{equation}
where \( \beta \) controls the amplitude of the perturbation in the input vector, which may be large (see Figure~\ref{fig:variation_of_input_perturbations}).
\textcolor{blue}{In Figure \ref{fig:ProbForm} we provide a sketch of the problem formulation. The plots at the top represent how uncertainty in the random input function $f(x)$  maps to the uncertainty in the output function $g(y)$ via the nonlinear operator $\mathcal{N}$.}  

\textcolor{blue}{
We emphasize that the theoretical framework developed hereafter can be applied to general finite-dimensional input-output pairs $\{\bm f,\bm g\}$, where $\bm f=\bm \mu +\bm z$ and $\bm z$ is distributed as in \eqref{PDF}. The use of a polynomial function space,  namely the space of Lagrange interpolants based on a finite set of Gauss-Legendre-Lobatto nodes, is introduced  as a practical tool for approximating the nonlinear operator \eqref{eq:intdifop}, as well as for implementing numerical experiments.}

\section{Modeling feed-forward neural networks as discrete dynamical systems}
\label{sec:methodology}

\begin{figure}[t]
    \centering
    \includegraphics[scale=0.9]{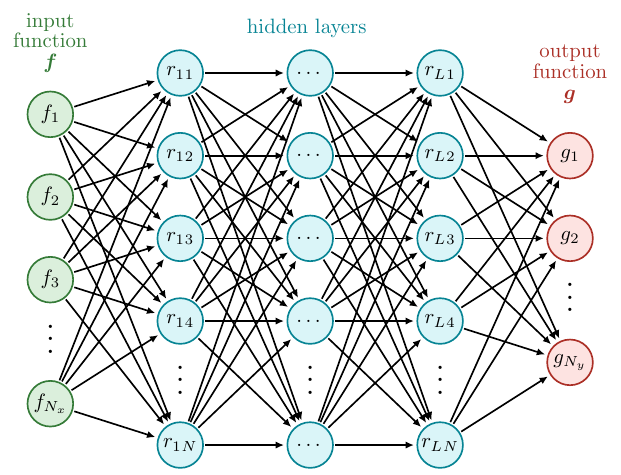}
     \caption{Schematic of the fully connected feed-forward (MLP) neural network used in this study.}
     \label{fig:NNDiagram}
\end{figure}

The input-output map representing the nonlinear operator \eqref{eq:intdifop} is chosen as a fully connected  feed-forward neural network with $L$ hidden layers. As is well known \cite{Venturi_MZDeep_learning,Li2022}, such neural network can be modeled as a discrete dynamical system via the following recurrence relation (see Figure \ref{fig:NNDiagram})
\begin{equation}
\layerOutvec_1 =  \layer_1(\bm f; \params_1), \qquad\ldots \qquad \layerOutvec_{n} = \layer_{n}(\layerOutvec_{n-1}; \params_{n}), \hspace{0.5cm} n=2,\ldots, L.
\end{equation}
 Here, $\bm f\in \mathbb{R}^{N_x} $ is the vector representing the random input function $f(x)$, \(\params_n \) are the set of neural network parameters for the \( n \)-th layer, usually consisting of weight and bias parameters \( \{\bm {W}_n, \bm{b}_n\} \), where \( \bm{W}_n \in \mathbb{R}^{N \times N} \), \(\bm{W}_1 \in \mathbb{R}^{N \times N_x}\),  \( \bm{b}_n \in \mathbb{R}^N \), and \( \layerOutvec_n \in \mathbb{R}^N \) ($n=1,\ldots L$). 

The operator  \( \layer_n \)  represents the mapping from the  hidden state $\bm r_{n-1}$ to $\bm r_n$. Its form varies depending 
on the type of neural network being considered, e.g., fully connected networks (MLPs), convolutional neural networks (CNNs), and residual networks (ResNets), among others.
In the simple fully-connected feed-forward net  setting we consider here,  \(\layer_n\) takes the following recurrent form across all \(L\) layers
  \begin{equation}
 \layer_n(\layerOutvec_{n-1}; \params_n) := \neu_n(\Wparm_n \layerOutvec_{n-1} + \bparm_n)\qquad n=1,\ldots, L,
 \label{eq:MLP_Sys}
\end{equation}
where  \(\neu_n : \R \to \R\) is the activation function at the \(n\)-th layer. In this paper, we assume that  the activation function is the same across all \textcolor{blue}{hidden} layers, i.e. we set $\neu_n=\neu$ for all $n=1, \ldots, L$, and select a leaky rectified linear unit \textcolor{blue}{(leaky ReLU)}, a piecewise-linear variant of the standard ReLU function, defined as\footnote{We set $\alpha=0.01$ in \eqref{eq:leakyReLU}. }
 \begin{equation}
    \phi(x) = \begin{cases}
    	x & x \geq 0\\ 
       \alpha x & \text{otherwise}
    \end{cases}. \label{eq:leakyReLU}
   \end{equation}
Note that  $\phi(x)$ can be expressed in terms of the Heaviside step function \( H(x) \) as
   \begin{align}
   \phi(x)   = H(x) x + \alpha (1 - H(x)) x.
  \end{align}
The network output is represented by a linear layer 
represented by a matrix $\bm A\in \mathbb{R}^{N_y\times N}$, with trainable entries, where $N_y$ is the number 
of grid points we use to represent the output function $g(y)$. This allows us to write the fully discrete input-output map as 
  \begin{align}
 \bm g =& \outwts \layer_{L}(\layerOutvec_{L-1}; \params_{L})\nonumber\\
    =& \outwts \layer_L \circ \layer_{L-1} \circ \cdots \circ \layer_2 \circ \layer_1(\bm f; \params_{1}).
\label{themap}
  \end{align}
Equation \eqref{themap} can be written explicitly for one-, two-, and $L$-layers as follows 
\begin{align}
 \bm g =&\bm A \phi(\bm W_1 \bm f +\bm b_1),\label{g1}\\
  \bm g =&\bm A \phi(\bm W_2  \phi(\bm W_1 \bm f +\bm b_1) +\bm b_2), \label{g2}\\
  \vdots&\nonumber\\
   \bm g =&\bm{A} \neu\left( \bm{W}_L \neu\left( \bm{W}_{L-1}\neu\left( \cdots \neu\left( \bm{W}_1 \bm{f} + \bm{b}_1 \right) \cdots\right) + \bm{b}_{L-1} \right) + \bm{b}_L \right).\label{g3}  
\end{align}

\section{PDF of the neural network output}
\label{sec:PDFoutput}
The sequence of vectors 
$\{\bm f,\bm r_1,\bm r_2, \ldots,\bm r_L, \bm g\}$ 
forms a {discrete Markov process}. 
Hence, the joint probability density function (PDF) 
of the random vectors $(\bm g,\bm r_{L},\ldots,\bm r_1,\bm f)$, 
i.e., joint PDF of the state of the entire neural network (including input, output, and hidden states), can be factored out as
\begin{equation}
p(\bm g,\bm r_{L},\ldots,\bm r_1,\bm f)=p(\bm g|\bm r_{L}) p(\bm r_L|\bm r_{L-1}) p(\bm r_{L-1}|\bm r_{L-2})\cdots p(\bm r_1|\bm f)p(\bm f),
\label{jointPDFtransition}
\end{equation}
where $p(\bm f)$ is the joint probability density function of the random input vector $\bm f$.  The transition densities 
$p(\bm g|\bm r_{L})$,  $p(\bm r_{n}|\bm r_{n-1})$ and 
$p(\bm r_{1}|\bm f)$ can be easily expressed as in terms 
of multivariate Dirac delta functions as 
\begin{align}
p(\bm g|\bm r_{L}) = & \,\delta (\bm g - \bm A\bm r_L),\\
p(\bm r_n|\bm r_{n-1}) = &\, \delta (\bm r_n -\phi(\bm W_n \bm r_{n-1} +\bm b_n)),\\
p(\bm r_{1}|\bm f) = &\, \delta (\bm r_1 -\phi(\bm W_1 \bm f +\bm b_1)).
\end{align}
By integrating \eqref{jointPDFtransition} relative to the phase variables $\{\bm f,\bm r_1,\ldots, \bm r_{L}\}$ we can easily express the PDF of the output $\bm g$ in terms for the PDF of the input $\bm f$. 
For instance, for a one-layer network we have 
\begin{align}
p(\bm g) = & 
\int_{\mathbb{R}^{N_x}}
\delta (\bm g - \bm A\phi(\bm W_1 \bm f +\bm b_1))p(\bm f) d\bm f . 
\label{15}
\end{align}
Furthermore, we can write the multivariate Dirac delta at  the right hand side of \eqref{15} in a Fourier form as 
\begin{equation}
\delta (\bm g - \bm A\phi(\bm W_1 \bm f +\bm b_1)) = \frac{1}{(2\pi)^{N_y}}\int_{\mathbb{R}^{N_y}} e^{i\bm a\cdot\left[\bm g-\bm A\phi(\bm W_1 \bm f +\bm b_1)\right]}d\bm a.
\end{equation}
Substituting this expression into \eqref{15}  yields the following input-output PDF map  
\begin{align}
p(\bm g) = \frac{1}{(2\pi)^{N_y}} 
 \int_{\mathbb{R}^{N_y}} \int_{\mathbb{R}^{N_x}} e^{i\bm a\cdot \left[\bm g-\bm A\neu(\Wparm_1 \bm f + \bparm_1)\right]}
p(\bm f) d\bm f d\bm a.
\label{16}
\end{align}
Note that the PDF of $\bm g$  depends on the weights and biases of the first layer, the leaky ReLU activation function $\phi(\cdot)$, and the PDF of the input $p(\bm f)$. Similarly, the input-output PDF map for a network with two layers can be written as  
\begin{align}
p(\bm g) = \frac{1}{(2\pi)^{N_y}} 
 \int_{\mathbb{R}^{N_y}} \int_{\mathbb{R}^{N_x}}e^{i\bm a\cdot \left[\bm g-\bm A\neu(\Wparm_2 \neu(\Wparm_1 \bm f + \bparm_1) + \bparm_2)\right]}p(\bm f) d\bm f d\bm a.
\label{17}
\end{align}
This can be easily generalized to networks with $L$ layers as
\begin{align}
p(\bm g) = \frac{1}{(2\pi)^{N_y}} 
 \int_{\mathbb{R}^{N_y}} \int_{\mathbb{R}^{N_x}}e^{i\bm a\cdot \left[\bm g-\bm{A} \neu\left( \bm{W}_L \neu\left( \bm{W}_{L-1}\neu\left( \cdots \neu\left( \bm{W}_1 \bm{f} + \bm{b}_1 \right) \cdots\right) + \bm{b}_{L-1} \right) + \bm{b}_L \right)
 )\right]}p(\bm f) d\bm f d\bm a.
\label{17_1}
\end{align}

\subsection{Linearization of leaky ReLUs}
\label{sec:linerizedLeakyReLU}
While equations \eqref{16}-\eqref{17_1} are {\em exact}, their 
analytical computation, i.e., the computation of the integrals, may be intractable. Hence, we introduce an approximation that significantly simplifies the analysis, i.e., 
we linearize the leaky ReLU activation functions of the network.
\textcolor{blue}{The leaky ReLU activation function  \eqref{eq:leakyReLU} is piecewise linear, and therefore the notion of ``linearization'' should be interpreted in context. Specifically, \eqref{eq:leakyReLU} has slope 1 for $x > 0$ and slope $\alpha$ for $x < 0$, with a discontinuity in the derivative at $x = 0$. Hence, linearizing $\phi(x)$ at $x=a$ corresponds to approximating $\phi(x)$ with the linear function $\phi'(a) x$ for all $x\in\mathbb{R}$.} 

For a network with one layer we obtain
\begin{align}
\phi(\bm W_1( \bm \mu+\bm z) +\bm b_1) \simeq  \phi(\bm W_1 \bm \mu +\bm b_1) + 
 \phi'(\bm W_1 \bm \mu +\bm b_1)\boxdot\bm W_1 \bm z,
 \label{KR}
\end{align}
where \( \boxdot \) is a variant of the Khatri-Rao 
product known as the {\em face-splitting product}\footnote{Given an \(m \times n \) matrix \(\bm{X}\) (with rows $\bm X_i$) and an \(m \times p \) matrix \(\bm{Y}\) (with rows $\bm Y_i$)  
 \begin{equation}
    \bm{X} = \begin{bmatrix}
\bm X_1\\
\vdots\\
\bm X_m
\end{bmatrix},
\qquad 
    \bm{Y}  = \begin{bmatrix}
\bm Y_1\\
\vdots\\
\bm Y_m
\end{bmatrix},
  \end{equation}
the face-splitting product between \(\bm{X}\) and \(\bm{Y}\) is defined as their corresponding row-wise Kronecker product. This results in the following \(m \times np\) matrix
\begin{equation}
\bm{X} \boxdot \bm{Y} =\begin{bmatrix}
\bm{X}_1 \otimes\bm{Y}_1\\
\vdots\\
\bm{X}_m \otimes\bm{Y}_m
\end{bmatrix}
  \end{equation}
},  $\phi'(\cdot)$ is the first derivative of the leaky ReLU activation function \eqref{eq:leakyReLU}
\begin{equation}
\phi'(x) = \begin{cases}
1 & x > 0 \\
\alpha & x<0 
\end{cases},
\end{equation}
 and $\bm z$ is the random component in the 
input vector $\bm f$  (see Eq. \eqref{f}). Equation \eqref{KR}  can be 
written in compact form as 
\begin{equation}
\phi(\bm W_1( \bm \mu+\bm z) +\bm b_1) \simeq  \phi(\bm W_1 \bm \mu +\bm b_1) +\bm J_1 \bm z, 
\label{P1}
\end{equation}
where 
\begin{equation}
\bm J_1= \phi'(\bm W_1 \bm \mu +\bm b_1)\boxdot\bm W_1.
\label{J1}
\end{equation}
The advantage of using this notation is that it holds for multiple layers as well. For instance, it can be shown that for a network with two layers we have 
\begin{equation}
\phi(\bm W_2\phi(\bm W_1( \bm \mu+\bm z) +\bm b_1) +\bm b_2) \simeq  \phi(\bm W_2\phi(\bm W_1 \bm \mu + \bm b_1) +\bm b_2)+\bm J_2\bm z, 
\label{P2}
\end{equation}
where 
\begin{equation}
\bm J_2  = \left[\phi'(\bm W_2 \phi(\bm W_1 \bm \mu +\bm b_1)  +\bm b_2)\boxdot\bm W_2\right] \phi'(\bm W_1 \bm \mu +\bm b_1)\boxdot \bm W_1.
\label{J2}
\end{equation}
These expressions can be readily extended to networks with \( L \) layers by induction. 
In essence, $\bm J_L$ is a product of neural network Jacobians (readily computed via backpropagation) mutiplied by \( \bm{W}_j \) through the face-splitting product, i.e.,
\begin{align}
\bm J_L = & \left[\neu'\left( \bm{W}_L \neu\left( \bm{W}_{L-1}\neu\left( \cdots \neu\left( \bm{W}_1 \bm{\mu} + \bm{b}_1 \right) \cdots\right) + \bm{b}_{L-1} \right) + \bm{b}_L \right)\boxdot \bm W_L\right] \cdots \nonumber\\ 
  & \left[ \phi'(\bm W_2 \phi(\bm W_1 \bm \mu +\bm b_1)  +\bm b_2) \boxdot \bm W_2\right]\phi'(\bm W_1 \bm \mu +\bm b_1)\boxdot \bm W_1.
  \label{JL}
\end{align}

\subsection{Approximation of the PDF of the neural network output}
\label{sec:PDFapprox}
With the linearized composition of leaky ReLU available, we can now compute analytically the approximated PDF of the $j$-th component of the output $g_j$. In \ref{app:PDF} we show that for uniformly distributed inputs vectors $\bm f=\bm \mu+\bm z$ with $p(\bm z)$  given in \eqref{PDF}, the PDF $p(g_j)$ can be approximated as a {\em convolution of rectangular functions}. Specifically,  
 \begin{align}
p(g_j) \simeq \frac{1}{2\pi}\left(\prod_{n= 1}^{N_x}\frac{2\pi}{q_{jn}}\right) \Pi_{\beta q_{j1}}[\eta(g_j)]*\cdots * \Pi_{\beta q_{jN_x}}[\eta(g_j)]
 \label{eq:FINAL_PDF}
\end{align}
where 
\begin{equation}
\bm q_j = \bm A_j\bm J_L, 
\label{qj}
\end{equation}
is a row vector with $N_x$ components, $\bm A_j$ is the $j$-th row of the output matrix $\bm A$, $\bm J_L$ is the matrix \eqref{JL}, 
\begin{equation}
\eta(g_j) = g_j - \bm{A}_j \cdot \neu\left( \bm{W}_L \neu\left( \bm{W}_{L-1}\neu\left( \cdots \neu\left( \bm{W}_1 \bm{\mu} + \bm{b}_1 \right) \cdots\right) + \bm{b}_{L-1} \right) + \bm{b}_L \right),
\end{equation}
$\beta$ is the ``perturbation amplitude'' (see Eq. \eqref{PDF}), $*$ is standard convolution operation, and $\Pi_{\beta q_{j1}}(\cdot)$ is the rectangular function (see Eq. \eqref{r}).
In practice, one may compute \( p(g_j) \) using its inverse Fourier transform representation, which involves a product of $\text{sinc}(\cdot)$ functions as in \eqref{eq:t}, rather than through the iterated convolution of rectangular functions in \eqref{eq:FINAL_PDF}, which can be more computationally expensive.

In \ref{app:error} we develop a thorough error analysis to quantify the effects of linearizing leaky ReLU activation functions on the neural network output. This analysis provides insight into why the approximate analytical formula \eqref{eq:FINAL_PDF} and the formulas for the statistical moments derived hereafter in Section \ref{sec:moments}, remain remarkably accurate even in the presence of large perturbations in the input vector.

\section{Statistical moments}
\label{sec:moments}
The linearized leaky ReLU approximation discussed 
in Section \ref{sec:linerizedLeakyReLU} allows us to derive simple 
approximate expressions for the mean and the covariance matrix of the neural 
network output $\bm g$. To this end, consider the 
``unperturbed'' network output 
\begin{equation}
m_j = \bm{A}_j \cdot \neu\left( \bm{W}_L \neu\left( \bm{W}_{L-1}\neu\left( \cdots \neu\left( \bm{W}_1 \bm{\mu} + \bm{b}_1 \right) \cdots\right) + \bm{b}_{L-1} \right) + \bm{b}_L \right).
 \end{equation}
Recalling the linearized leaky ReLU approximation (e.g., Eq, \eqref{P1}) we 
can write the neural network output linearized around the input 
vector $\bm \mu$ as
\begin{equation}
g_j \simeq m_j + \bm q_j \cdot \bm z,
\end{equation}
where the vector $\bm q_j$ is defined in \eqref{qj}. Recalling that $\bm z$ is a zero-mean vector, from this expression it follows that 
\begin{align}
\mathbb{E}\{g_j\} \simeq &m_j + \bm q_j \cdot \mathbb{E}\{\bm z\}  \nonumber\\
 = &m_j.\label{mean}
\end{align}
Similarly,
\begin{equation}
\mathbb{E}\{g_ig_j\} \simeq m_i m_j  + \bm q_i \bm Q \bm q^T_j,
\end{equation}
where the matrix $\bm Q$ has entries
\begin{equation}
Q_{ln} = \mathbb{E}\{z_lz_n\} \qquad \text{(covariance of the perturbation vector $\bm z$).}
\end{equation}
If we assume that $\bm z$ has i.i.d uniform components in $[-\beta,\beta]^{N_x}$ (see Eq. \eqref{PDF}) then 
\begin{equation}
\mathbb{E}\{g_ig_j\} \simeq  m_i m_j  +\frac{\beta^2}{3} \bm q_i  \cdot \bm q_j. 
\label{corr}
\end{equation}
In this case, the variance of $g_j$ and the covariance of $g_i$ 
and $g_j$ can be written as 
\begin{align}
\text{Var}(g_j) \simeq & \frac{\beta^2}{3} \left\|\bm q_j\right\|^2_2, \label{var}\\
\text{Cov}(g_i,g_j) \simeq & \frac{\beta^2}{3} \bm q_i\cdot \bm q_j\label{cov}.
\end{align}

\section{Gaussian copula approximation of the PDF of the neural network output}
\label{sec:copula}
%
Let us transform  each component $g_i$ of the random vector $\bm g$ representing the output of the neural network to a uniform 
random variable $u_i$ via the probability mapping 
\begin{equation}
u_i=F_i(g_i) \qquad i=1,\dots, N_y,
\label{CC}
\end{equation}
 where $F_i$ is the cumulative distribution function (CDF) of $g_i$.
A {\em copula} is defined as the joint cumulative distribution 
function of $\bm u=[u_1,\dots,u_{N_y}]$, i.e., $F_{\bm u}(u_1,\dots, u_{N_y})$. 
Note that  such copula is a multivariate CDF defined on 
the unit cube \( [0,1]^n \), which encodes all information about 
the dependence structure among the components 
of the random vector \( \bm{g} \).
By using the definition \eqref{CC} it can be shown that
 \begin{align}
 F_{\bm u}(u_1,\dots, u_{N_y})  =F_{\bm g}\left(F^{-1}_1(u_1),\dots, F^{-1}_{N_y}(u_{N_y})\right),
  \label{Sklar}
  \end{align}
Where   $\bm F_{\bm g}$ is the joint CDF of the random vector $\bm g$. 
 This equation can be reversed to obtain
  \begin{equation}
  F_{\bm g}(g_1,\dots,g_{N_y})=F_{\bm u}\left(F_1(g_1),\dots, F_{N_y}(g_{N_y})\right).
  \label{FFF}
  \end{equation}
 Indeed, it is always possible write $F_{\bm g}(g_1,\dots,g_{N_y})$ 
 in terms of a copula $F_{\bm u}(u_1,\dots, u_{N_y})$ 
 and the individual CDFs $F_i$ of each component $g_i$. This result 
 is known as {\em Sklar's theorem} \cite[Theorem 1.9]{Czado}. 
By differentiating \eqref{FFF} with respect to $(z_1,\dots,z_n)$ 
yields the copula representation of the PDF of $\bm g$, i.e., 
\begin{equation}
  p_{\bm g}(g_1,\dots,g_{N_y}) = p_{\bm u}(F_1(g_1),\dots, F_{N_y}(g_{N_y}))
  p_1(g_1)\cdots p_{N_y}(g_{N_y}), 
  \label{PDFNNOUT}
  \end{equation}
where $p_{\bm u}(u_1,\dots, u_{N_y})$ is the PDF of the copula, and $p_j(g_j)$ 
is the PDF of $g_j$.
  
\subsection{Gaussian copulas}
Gaussian copulas are constructed from a multivariate 
normal distribution  by using simple one-dimensional probability mappings.
For a given correlation matrix $\bm R$ the Gaussian copula is defined as 
\begin{equation}
F_{\bm u}(u_1,\dots,u_{N_y}) = F^{\text{Gauss}}_{\bm R} \left(G^{-1}(u_1),\dots,G^{-1}(u_{N_y})\right),
\label{GCP} 
\end{equation}
where $F^{\text{Gauss}}_{\bm R}$ is the CDF of a zero-mean Gaussian 
with correlation matrix $\bm R$, and $G$ are one-dimensional 
CDFs of standard Gaussian variables with zero mean and variance one, i.e., 
\begin{equation}
G(u) = \frac{1}{2}\left[1+\text{erf}\left(\frac{u}{\sqrt{2}}\right)\right].
\label{Fg}
\end{equation}
By differentiating \eqref{GCP} with respect to 
$\{u_1,\dots, u_n\}$ we obtain\footnote{Recall that 
  \begin{equation}
  \frac{d G^{-1}(G(x))}{dx} = 0 \quad \Rightarrow\quad  \frac{d G^{-1}(u)}{du}= \frac{1}{G'( G^{-1}(u))}= \frac{1}{p\left(G^{-1}(u)\right)}.
  \end{equation}
  This explains the term $\bm z_u\bm I\bm z_u^T$ in \eqref{Gcop}.
 }
  \begin{equation}
  p_{\bm u}(u_1,\dots,u_{N_y})=\frac{1}{\sqrt{\det(\bm R)}}
  \exp\left[\frac{1}{2}\bm z_{\bm u}\left(\bm I-\bm R^{-1}\right)\bm z_{\bm u}^T\right],
  \label{Gcop}
  \end{equation}
  where 
  \begin{equation}
  \bm z_{\bm u} = \left[G^{-1}(u_1),\dots,G^{-1}(u_{N_y})\right].
  \end{equation}
Substituting~\eqref{Gcop} into~\eqref{PDFNNOUT} yields the Gaussian copula approximation of the PDF of the neural network output \( \bm{g} \). Note that this approximation depends only on the marginal PDFs of the individual components \( g_j \) (see~\eqref{eq:FINAL_PDF}) and the copula correlation matrix $\bm R$ is  which can be computed, e.g., from~\eqref{corr}. To this end, let  
\begin{equation}
C_{ij}=\mathbb{E}\{g_ig_j\}.
\end{equation}
By using the copula PDF representation we have 
\begin{equation}
C_{ij} = \int_{[0,1]^2} F^{-1}_i(u_i)F^{-1}_j(u_j)p_{\bm u}(u_i,u_j;\rho_{ij})du_idu_j\qquad i,j=1,\ldots, N_y
\end{equation}
  where 
  \begin{equation}
  p_{\bm u}(u_i,u_j;\rho_{ij}) = \frac{1}{\sqrt{1-\rho_{ij}^2}}\exp\left[\frac{1}{2(1-\rho_{ij}^2)}\left[2\rho_{ij}G^{-1}(u_i)G^{-1}(u_j)-\rho_{ij}^2\left(G^{-1}(u_j)^2+G^{-1}(u_j)^2\right)\right]\right]
  \end{equation}
  are 2D marginals\footnote{Recall that, by definition, the 
  covariance of a Gaussian copula has ones along 
  the diagonal (variances are one). This implies that 
  the off-diagonal entries of $\bm R$ coincide with 
  the correlation coefficients $\rho_{ij}$, i.e., 
  \begin{equation}
  R_{ij} =\begin{bmatrix}
  1         & \rho_{ij}\\ 
  \rho_{ij} & 1
  \end{bmatrix} 
  \end{equation}
 } of \eqref{Gcop}, and $\rho_{ij}$ is the correlation coefficient. 
Given a target correlation matrix $C_{ij}$, e.g., the analycal correlation \eqref{corr}, we can campute the copula correlation coefficients $\rho_{ij}$ by solving the following sequence of 1D optimization problem 
  \begin{equation}
  \min_{\rho_{ij}\in[-1,1]} \left| C_{ij} -\int_{[0,1]^2} F^{-1}_i(u_i)F^{-1}_j(u_j)p_{\bm u}(u_i,u_j;\rho_{ij})du_idu_j \right| \qquad j>i.
  \end{equation}
  Once all off-diagonal entries of $\bm R$ are computed in this way, 
  we can assemble the symmetric matrix 
  \begin{equation}
  \bm R = \begin{bmatrix}
  1 & \rho_{12}& \cdots & \rho_{1n}\\
  \rho_{12} & 1 & \cdots & \rho_{2n}\\
  \vdots & & \ddots & \vdots\\
  \rho_{1n} & \rho_{2n}& \cdots & 1
  \end{bmatrix}.
  \end{equation}
Of course  $\bm R$ may not be positive definite. The reason is that a greedy optimization procedure was used to compute each entry of \( \bm{R} \) independently. The closest positive definite matrix   \( \bm{R}^+ \)  to \( \bm{R} \) in the Frobenius norm can be easily obtained by performing a spectral decomposition of \( \bm{R} \) and setting all negative eigenvalues to zero.
  \begin{equation}
  \bm R = \bm V \bm \Lambda \bm V^T \quad \Rightarrow \quad \bm R^+ = \bm V \bm \Lambda^+\bm V^T\quad  \bm \Lambda^+=\max\{\bm \Lambda,\bm 0\}.
  \end{equation}
The transformation between correlated Gaussian copula variable $\bm u$ (with correlation $\bm R$) and the output vector $\bm g$ with given marginal distributions $p(g_j)$ and correlation $\bm C$ is known as {\em Nataf transformation}.

 \subsection {Sampling from Gaussian copulas} 

The procedure to sample from a Gaussian copula is very simple: 
\begin{enumerate}
\item We first sample realizations of a Gaussian vector with zero mean 
and unit variance, say $\bm h^{(j)}$.

\item Such sample vectors are then transformed
to a vector with correlation $\bm R$ (copula correlation) using the Cholesky 

factor\footnote{The matrix $\bm R$ is symmetric and 
positive definite. Hence it admits the Cholesky factorization $\bm R=\bm L\bm L^T$ 
where $\bm L$ is lower triangular with positive diagonal entries. If $\bm f$ is 
a Gaussian vector with zero mean and unit covariance then $\bm s=\bm L \bm f$ 
is a Gaussian vector with zero mean and covariance $\bm R$. In fact 
\begin{equation}
\mathbb{E}\left\{\bm s \bm s^T\right\}= \bm L \mathbb{E}\left\{\bm f \bm f^T\right\}\bm L^T = \bm L\bm L^T = \bm R.
\end{equation}} of $\bm R$. Call such samples $\bm s^{(j)}=\bm L \bm f^{(j)}$ ($\bm L$ is the lower-triangular Cholesky factor of $\bm R$).
  
\item Each component of the rotated vectors is then mapped to a uniform 
distribution via the Gaussian probability mapping $F_g$ (which is the same for all components), i.e., $\bm u^{(i)}=F_g(\bm s^{(j)})$. The samples $\bm u^{(j)}$ are samples of the Gaussian copula. 
  
\item At this point we generate a sample of the random vector 
of interest as $\bm g$ by using the marginal CDF mapping as 
\begin{equation}
\bm g^{(i)}=\left[F^{-1}_1\left(u^{(i)}_1\right),\dots,F^{-1}_{N_y}\left(u^{(i)}_{N_y}\right)\right].
\end{equation} 
Here, $F_i^{-1}$ is the inverse CDF of the random variable $g_i$, which we easily obtain from the PDF $p(g_i)$ (see Eq. \eqref{eq:FINAL_PDF}). 
\end{enumerate}

\textcolor{blue}{
\section{Comparison with other UQ methods}
\label{sec:comparison}
In this section, we compare the proposed uncertainty propagation method based on linearized leaky ReLU activation functions with other established methods for UQ in neural networks.
To this end, let us first recall that uncertainty in the neural network input is characterized by the PDF $p(\bm f)$ (e.g., Eq. \eqref{PDF}) while uncertainty in the output is described by the PDF $p(\bm g)$. In a Bayesian setting, the mapping from input to output (i.e., the model) is represented by the conditional density $p(\bm g|\bm f)$. Given both $p(\bm f)$ and $p(\bm g|\bm f)$, the PDF of the output can be obtained via marginalization
\begin{equation}
p(\bm g) = \int_{\mathbb{R}^{N_x}} p(\bm g|\bm f)p(\bm f) d\bm f.
\label{integ1}
\end{equation} 
For a deterministic MLP with $L$ layers, such as the one considered in this paper, the conditional density  $p(\bm g|\bm f)$ 
is obtained in closed form as\footnote{\textcolor{blue}{In Section \ref{sec:PDFoutput}, we derived a closed-form representation of  the output density \( p(\bm g) \) using the Fourier representation of the transition density \eqref{pgf} (see Eq.~\eqref{17_1})}.}
\begin{equation}
p(\bm{g} | \bm{f}) = \delta\left( 
\bm{g} - \bm{A} \neu\left( 
\bm{W}_L \neu\left( 
\bm{W}_{L-1} \neu\left( 
\cdots \neu\left( 
\bm{W}_1 \bm{f} + \bm{b}_1 
\right) \cdots 
+ \bm{b}_{L-1} 
\right) 
+ \bm{b}_L 
\right) 
\right) 
\right),
\label{pgf}
\end{equation}
where \( \delta(\cdot) \) denotes the Dirac delta function and \( \phi(\cdot) \) is the leaky ReLU activation function. This expression reflects the deterministic nature of the network, i.e., for a given input \( \bm{f} \), the output \( \bm{g} \) is {\em uniquely determined} by the nonlinear transformation \eqref{g3}.
}

\textcolor{blue}{
If the goal is to quantify uncertainty in an input--output map and we are not constrained to a fixed, deterministic, pre-trained network with random inputs, then a wide array of UQ techniques becomes available. 
In fact, rather than training a deterministic network, which results in the Dirac delta conditional density \eqref{pgf},  one can instead learn \( p(\bm{g} | \bm{f}) \) directly (or learn how to sample from it) using probabilistic modeling techniques within a supervised learning framework. Well-known examples of such techniques include Mixture Density Networks (MDNs)~\cite{bishop2006pattern}, Variational Bayesian Last Layers (VBLL)~\cite{VBLL}, Deep Ensembles  \cite{DeepEnsembles,deepEnsemblesBayes}, and more general Bayesian Neural Networks (BNNs)~\cite{murphy2023probabilistic,BNNVI}.
These methods aim to quantify the uncertainty in the mapping \( \bm{f} \mapsto \bm{g} \), and typically provide a distribution of possible outputs for each input \( \bm{f} \), either in the form of samples or an estimated PDF.
A comprehensive summary of these and other techniques, along with accessible implementations, can be found in~\cite{LightningUQ}. With the transition density  \( p(\bm{g} | \bm{f}) \) available in a tractable form, one 
can integrate \eqref{integ1} for given $p(\bm f)$ to compute the PDF of the output, or its statistical moments.}

\textcolor{blue}{ 
On the other hand, if we are given a fixed deterministic neural network and consider the problem of propagating input uncertainty through it -- as in the setting of this paper -- then the range of available techniques becomes significantly narrower. 
Given the high dimensionality of input--output maps typically encountered in neural networks, the only viable approaches are generally those based on random sampling, such as Monte Carlo (MC) and its low-discrepancy variants, e.g., quasi-Monte Carlo (qMC)~\cite{Lemieux,qmc}. These methods offer high asymptotic accuracy but suffer from slow convergence rates and significant computational cost, particularly when tight confidence bounds are required. Nevertheless, MC and qMC scale well with input dimension and are often regarded as ground truth for benchmarking other UQ methods, as we have done in this paper. Other sampling-based approaches that can potentially improve upon the convergence rates of MC and qMC for networks with smooth activation functions, such as Sparse Grids (SG)~\cite{Novak,Bungartz} or Probabilistic Collocation~\cite{Foo1}, quickly become computationally intractable as the dimension of $\bm f$ and $\bm g$ increases.}

\textcolor{blue}{An alternative strategy for propagating input uncertainty through a deterministic network is to use \emph{tractable surrogate approximations} of the input--output map \( \bm{f} \mapsto \bm{g} \). Such approximations often enable the derivation of closed-form expressions for key output statistics such as moments, uncertainty intervals, and PDFs. The linearized leaky ReLU approximation introduced in this paper falls into this category. Compared to sampling-based techniques like MC, qMC, or Bayesian methods that involve evaluating the posterior $p(\bm g)$ via Eq.~\eqref{integ1}, the linearized leaky ReLU approximation  is significantly more efficient from a computational standpoint. However, its accuracy may degrade when the input noise \( \bm{z} \) in Eq.~\eqref{f} becomes large, especially for input-output maps that exhibit strong nonlinearity. 
}

\section{Numerical results}
\label{sec:results}
In this section, we apply the uncertainty quantification methods developed in this paper to the {\em nonlinear} integro-differential operator
\begin{equation}
g(y) = \mathcal{N}(f) = \int_{-1}^{1} \left( f(x)y + f(x)f'(x)\sin(\pi y^2)\cos(x) \right) dx.
\label{nonlinear}
\end{equation}
To this end, we first approximate \( g(y) = \mathcal{N}(f(x)) \) using a feed-forward neural network trained on a large dataset of input-output pairs \( \{f(x), g(y)\} \) via a 
standard discrete \( L_2 \) performance metric. The input and output functions are discretized on a  Gauss-Legendre-Lobatto grid  \cite{Hesthaven} with \( N_x = N_y = 31 \) points in $[-1,1]$. This allows us to use high-order approximations of both the derivative and the integral appearing in \eqref{nonlinear}. The ensemble of input functions is constructed by sampling the $N_x$-dimensional input vector $\bm f$ representing $f(x)$ from a probability distribution with Gaussian  i.i.d. components.
We also consider a separate neural network approximating a simplified version of the previous operator, namely, a {\em linear} integro-differential operator
\begin{equation}
g(y) = \mathcal{L}f = \int_{-1}^{1} \left( f(x)y + f'(x)\sin(\pi y^2)\cos(x) \right) dx.
\label{linear}
\end{equation}
In Table \ref{tab:parameters} we summarize the set of optimization parameters we used to train both neural networks via the ADAM optimizer.
\begin{table}
 \centerline{ 
  \begin{tabular}{|p{2.5cm}||p{3cm}|p{2.5cm}|p{2.5cm}|p{2.5cm}|}
    \hline
    \multicolumn{5}{|c|}{Nonlinear Operator} \\
 \hline
    Layers & Number of Samples & Batch Size & Epochs & Learning Rate \\
    \hline
    1 Layer   & 1 M    & 2000      & 1000   & 0.01 \\
    5 Layers  & 1 M    & 1000      & 300    & 0.01 \\
    20 Layers & 1 M    & 2000      & 900    & 0.01\\
    \hline
\end{tabular}}
\vspace{0.5cm}
\centerline{
  \begin{tabular}{|p{2.5cm}||p{3cm}|p{2.5cm}|p{2.5cm}|p{2.5cm}|}
    \hline
    \multicolumn{5}{|c|}{Linear Operator} \\
    \hline
    Layers  & Number of Samples & Batch Size & Epochs & Learning Rate \\
    \hline
    1 Layer   & 1 M    & 1000      & 5000   & 0.001 \\
    5 Layers  & 1 M    & 1000      & 100    & 0.001 \\
    20 Layers & 1 M    & 1000      & 200    & 0.001 \\
    \hline
  \end{tabular}}
  \caption{Number of layers, number of input-output samples, and optimization parameters used to train the feed-forward neural networks approximating the operators \eqref{nonlinear} and \eqref{linear}. The ADAM optimizer was used for training. The number of neurons is fixed at \( N = 64 \) in each layer.}
\label{tab:parameters}
 \end{table}
\textcolor{blue}{The numerical results presented in this section were generated using  code developed upon the libraries \cite{makie,pal2023lux} and made publicly available at the GitHub repository \cite{diamzon2024uq_nn}.}

\subsection{Input functions perturbed by random noise}
\begin{figure}[t]
   \centering
    \begin{subfigure}[b]{.32\linewidth}
      \centering
        \caption{\(\beta=0.1\)}
      \includegraphics[width=\linewidth]{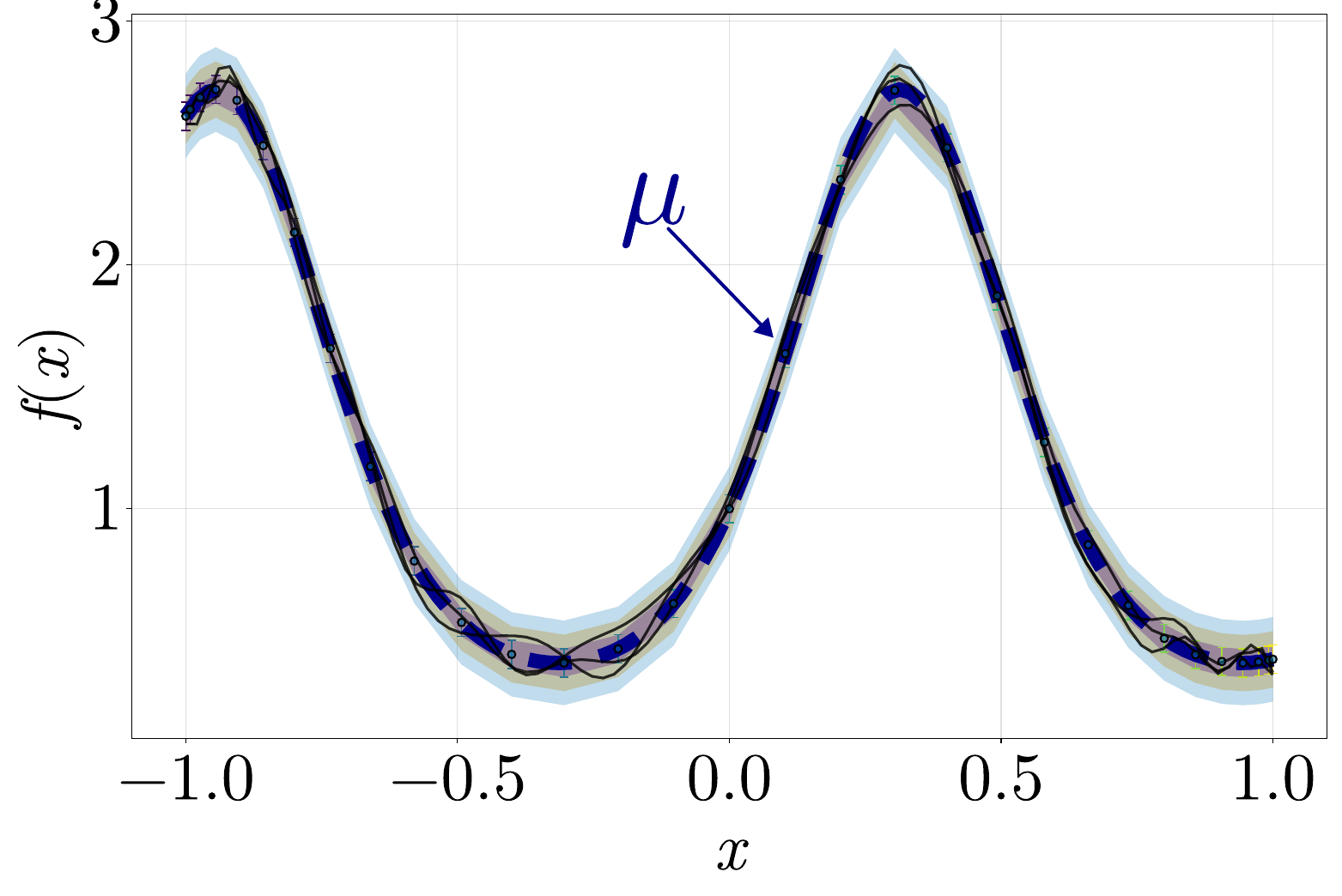}

    \end{subfigure}
    \hfill
    \begin{subfigure}[b]{.32\linewidth}
      \centering
      \caption{\(\beta=0.5\)}
      \includegraphics[width=\linewidth]{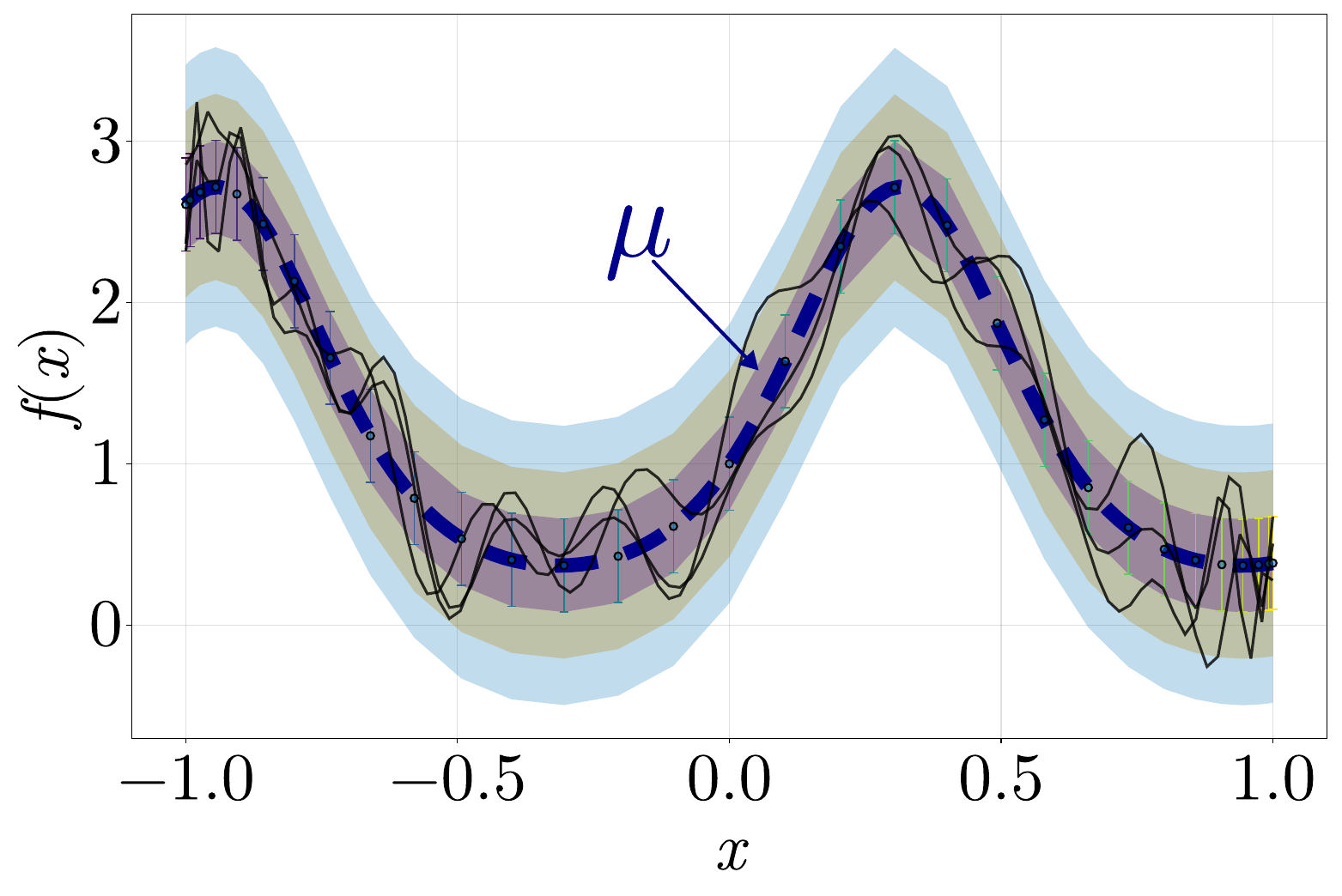}
    \end{subfigure}
    \hfill
    \begin{subfigure}[b]{.32\linewidth}
      \centering
      \caption{\(\beta=1.0\)}
      \includegraphics[width=\linewidth]{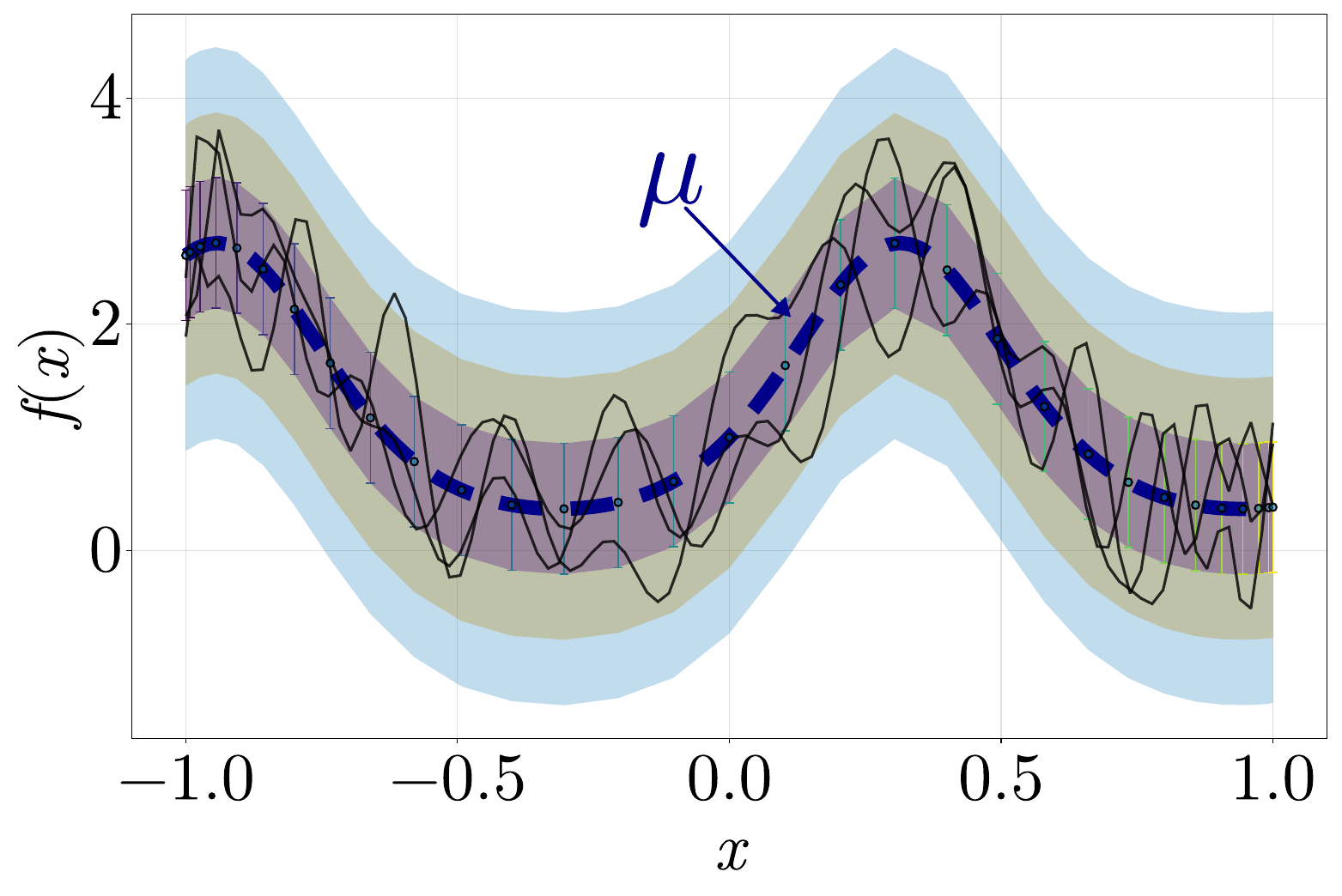}
    \end{subfigure}
    \\
    \begin{subfigure}[b]{.32\linewidth}
      \centering
      \caption{\(\beta=1.5\)}
      \includegraphics[width=\linewidth]{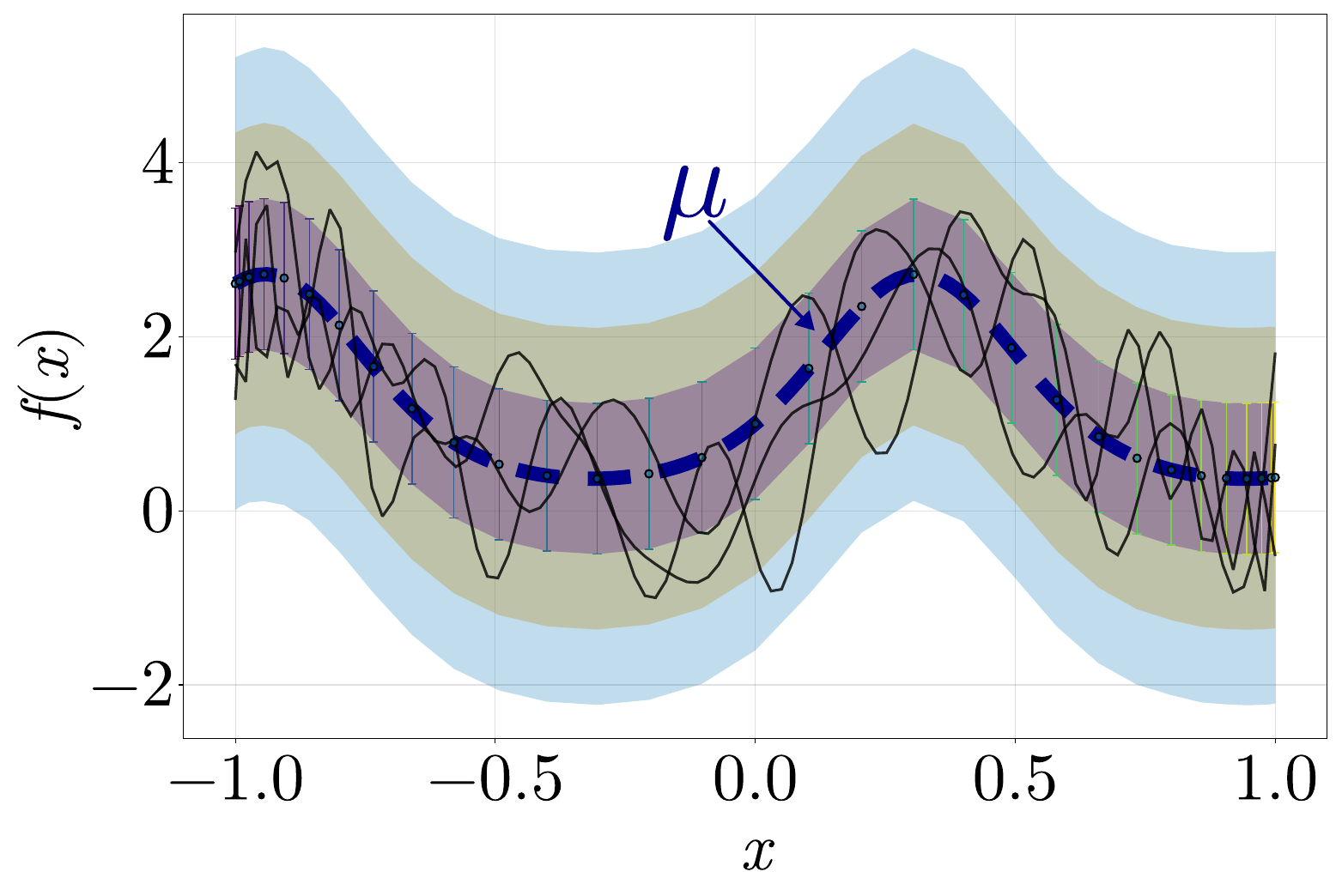}
    \end{subfigure}
    \hfill
    \begin{subfigure}[b]{.32\linewidth}
      \centering
      \caption{\(\beta=2.0\)} 
      \includegraphics[width=\linewidth]{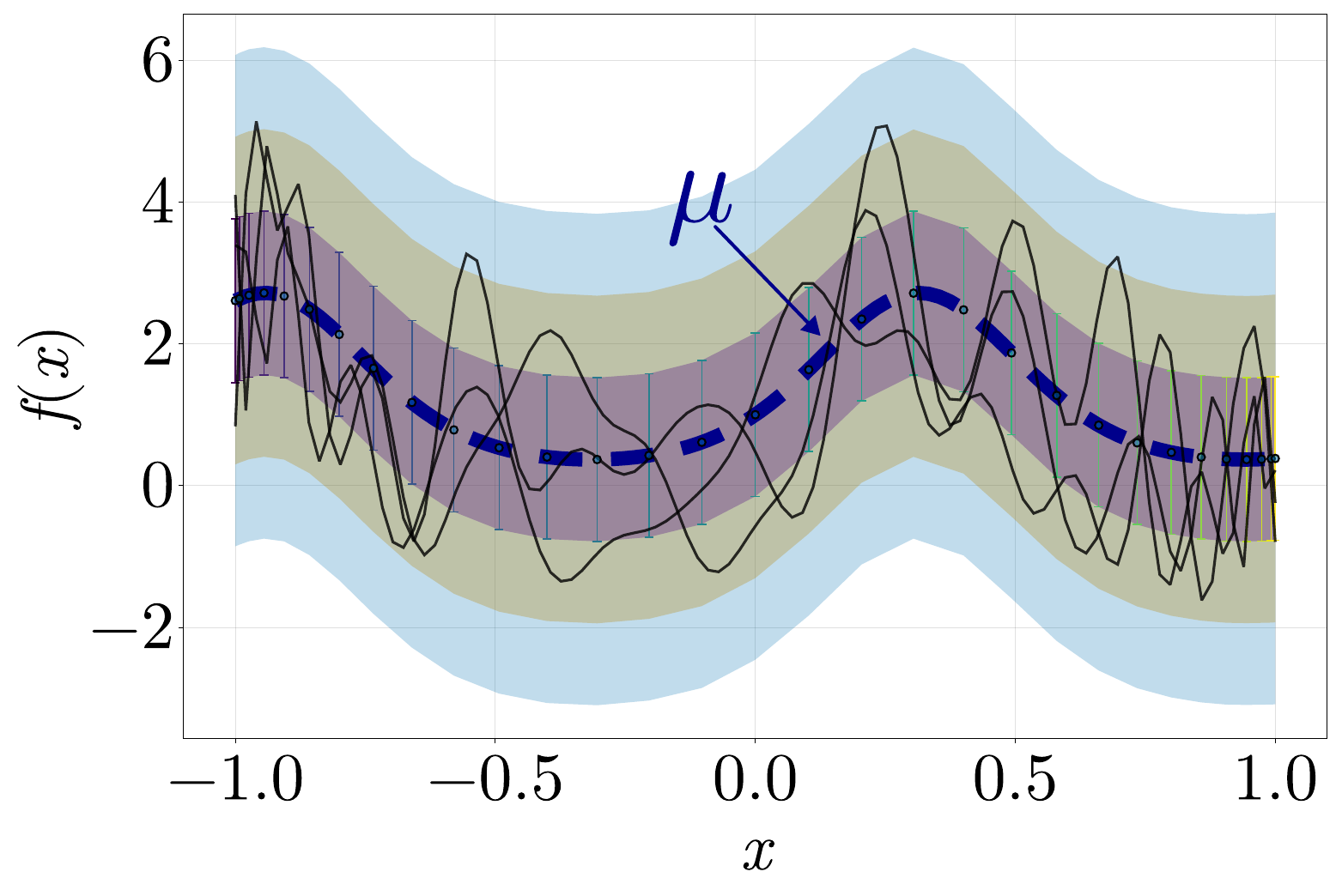}
    \end{subfigure}
    \hfill
    \begin{subfigure}[b]{.32\linewidth}
      \centering
       \caption{\(\beta=3.0\)}
      \includegraphics[width=\linewidth]{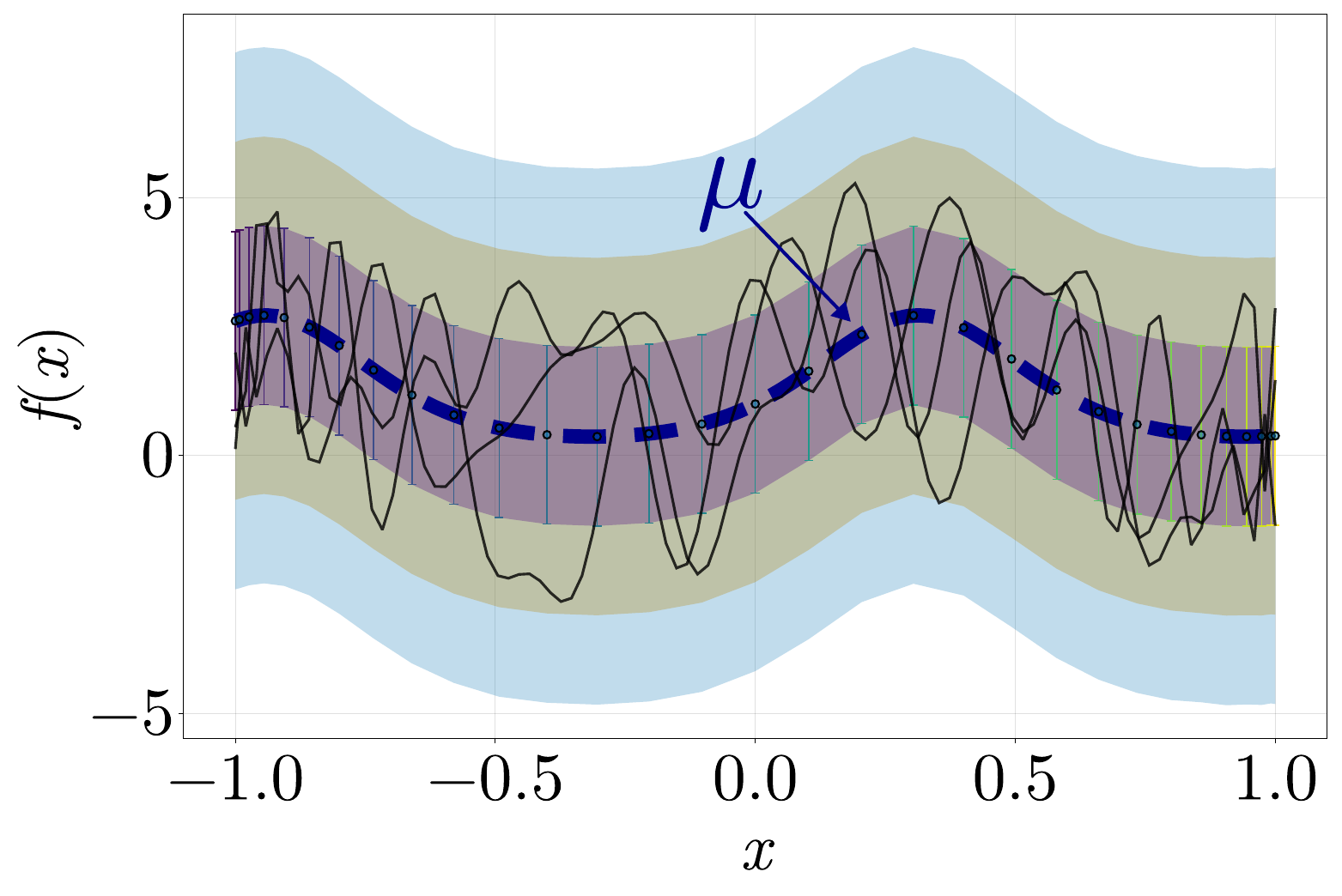}
    \end{subfigure}
    \hfill
       \caption{Samples of the random input function $f(x)$ for fixed mean and different perturbation amplitudes $\beta$. Shown are the Lagrange polynomial interpolating the mean vector \( \bm{\mu} \), along with several random realizations of \( f(x) \), i.e., Lagrange polynomials interpolating input vectors of the form \( \bm{\mu} + \bm{z} \), where $\bm z$ is a random vector with i.i.d.  uniform components in $[-\beta,\beta]$ (see Eq. \eqref{PDF}).}
       \label{fig:variation_of_input_perturbations}
  \end{figure}

With a fully trained neural network available, we now consider an input vector of the form \( \bm{f} = \bm{\mu} + \bm{z} \), where \( \bm{\mu} \) is {\em fixed} and \( \bm{z} \) is a random perturbation drawn from the uniform probability distribution defined in \eqref{PDF}, with varying perturbation amplitudes \( \beta \). In Figure \ref{fig:variation_of_input_perturbations}, we plot the Lagrange polynomial interpolating the mean vector \( \bm{\mu} \), along with several random realizations of \( f(x) \) i.e., Lagrange polynomials interpolating random input vectors of the form \( \bm{\mu} + \bm{z} \), for different values of \( \beta \).

\subsection{PDF of the neural network output and its statistical moments}
\begin{figure}[t]
    \centering
    \begin{subfigure}{0.9\textwidth}
      \centering
      {\includegraphics[width=0.85\linewidth]{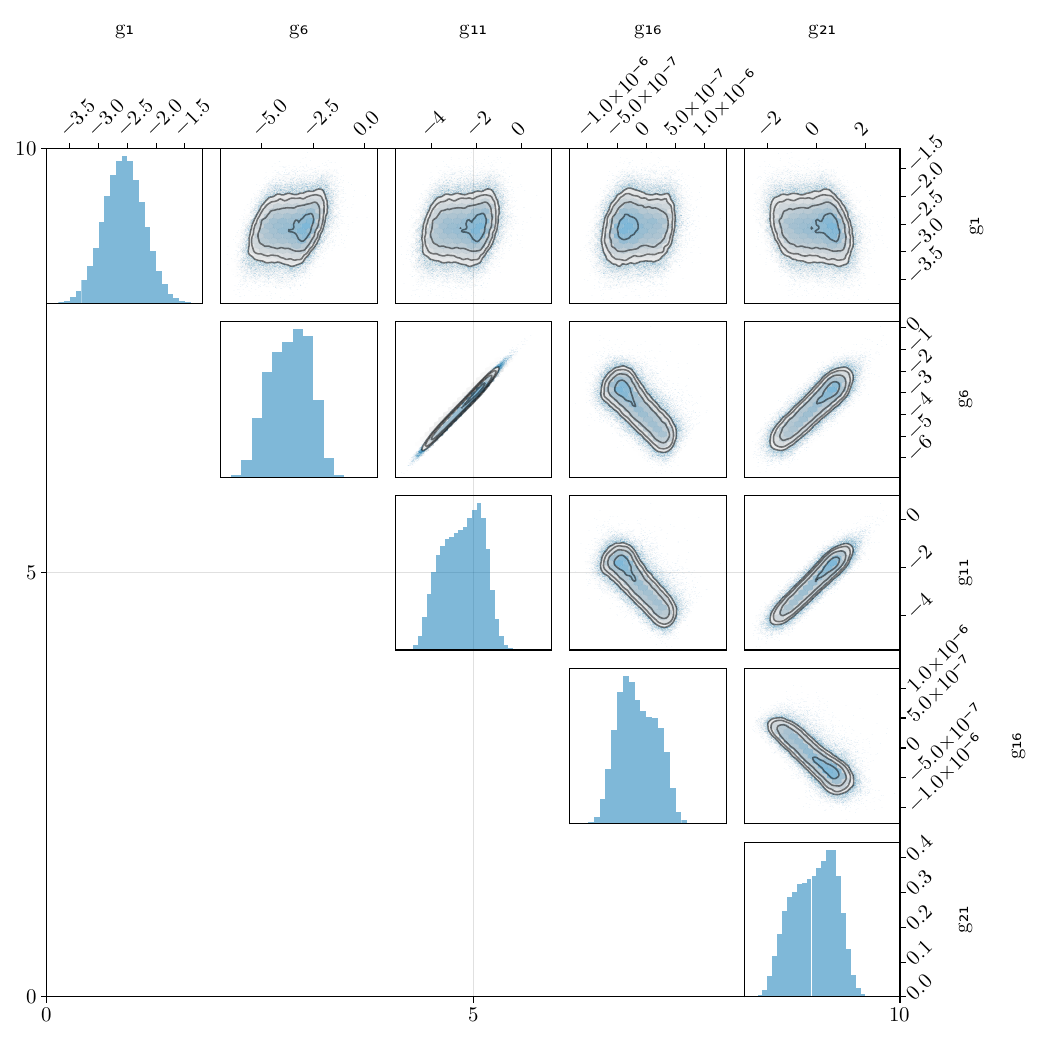}}
      \caption{}
    \end{subfigure}
       \caption{Nonlinear operator \eqref{nonlinear}. Corner plot of the one- and two-point PDFs of  \(g_1, g_6, g_{11}, g_{16}, g_{21}\) generated by MC simulation.}
       \label{fig:cornerplot_nonLin}
  \end{figure}

\begin{figure}[t]
    \centering
    \includegraphics[width=0.75\linewidth]{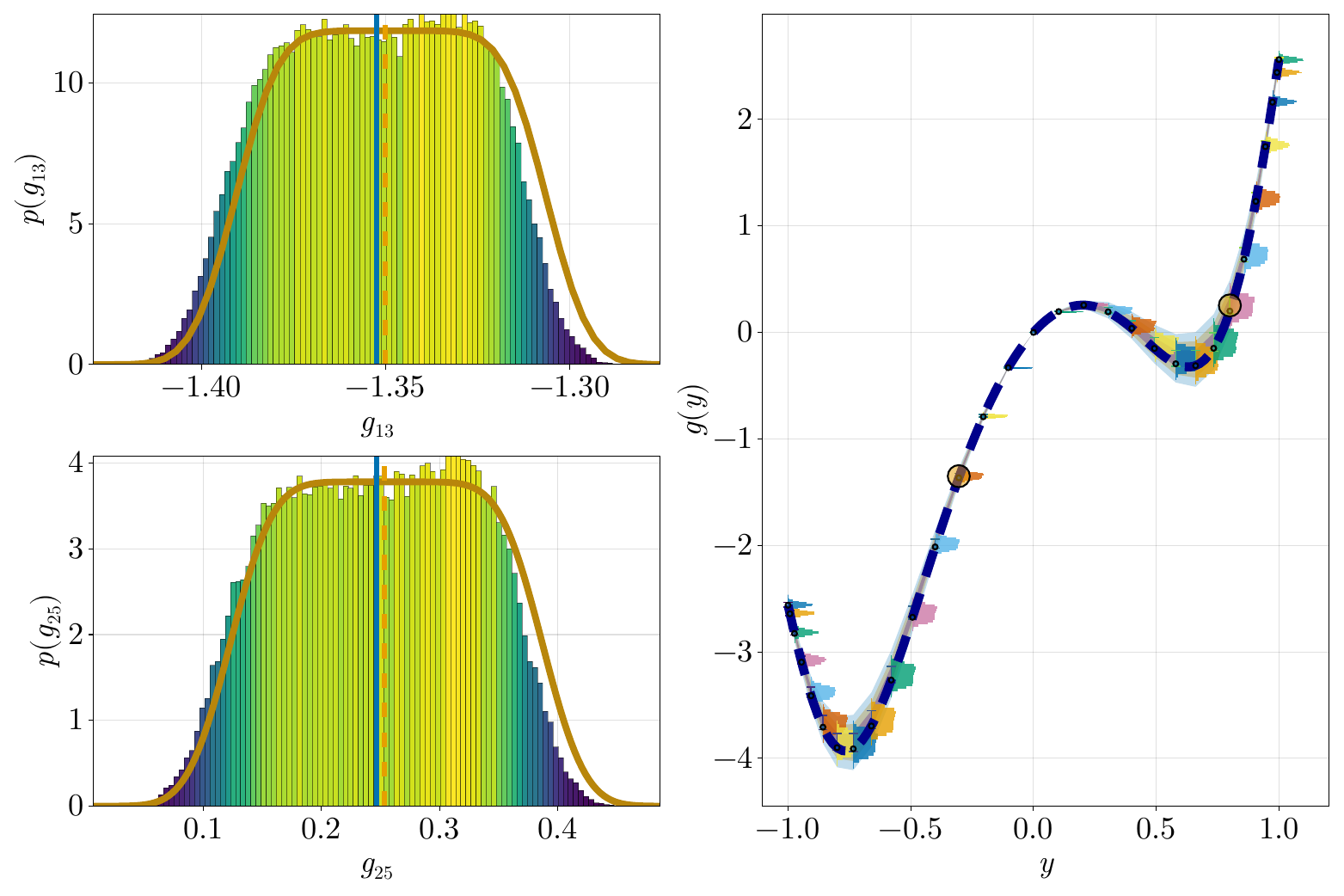}%
 \caption{Nonlinear operator \eqref{nonlinear}. Comparison between the PDFs of  $g_{13}$ and $g_{25}$ as predicted by \eqref{eq:FINAL_PDF} (yellow curve --   linearized leaky ReLU approximation) and the Monte Carlo benchmark. The perturbation amplitude in the input function here is set to $\beta=0.1$ (see Figure \ref{fig:variation_of_input_perturbations}).  We also plot the confidence intervals $\pm \kappa \sigma$  ($\kappa=1,2,3$) representing the variability of the output $g(y)$ relative to the mapped mean input function.}
       \label{fig:nonlin1}
  \end{figure}

\begin{figure}[t]
\centering
\includegraphics[width=0.75\linewidth]{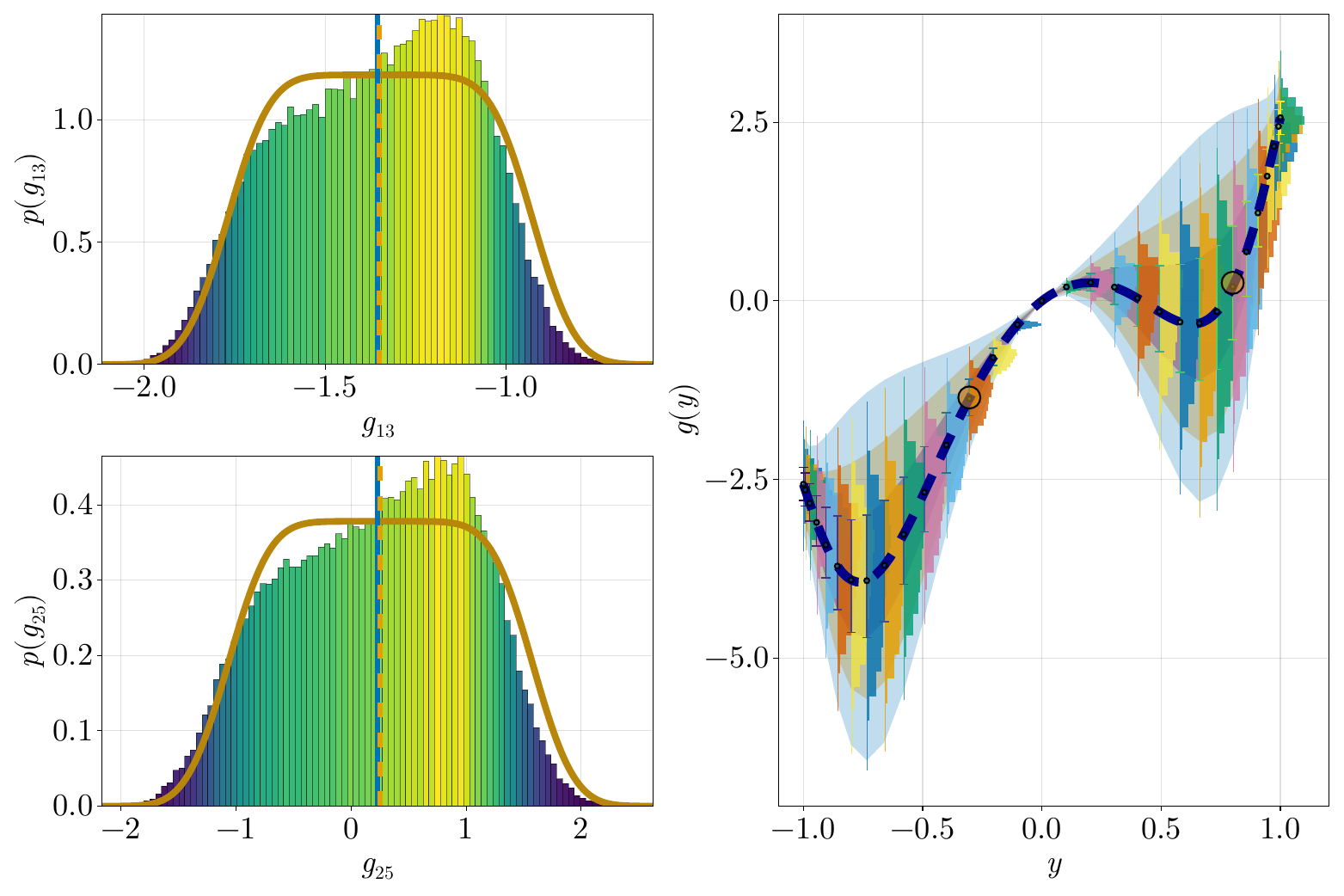}%
  \caption{Nonlinear operator \eqref{nonlinear}. Comparison between the PDFs of  $g_{13}$ and $g_{25}$ as predicted by \eqref{eq:FINAL_PDF} (yellow curve --   linearized leaky ReLU approximation) and the Monte Carlo benchmark. The perturbation amplitude in the input function here is set to $\beta=1$ (see Figure \ref{fig:variation_of_input_perturbations}).  We also plot the confidence intervals $\pm \kappa \sigma$  ($\kappa=1,2,3$) representing the variability of the output $g(y)$ relative to the mapped mean input function.}
 \label{fig:nonlin2}
  \end{figure}

\begin{figure}[t]
    \centering
    \begin{subfigure}[b]{0.9\linewidth}
      \centering
      \includegraphics[width=0.9\linewidth]{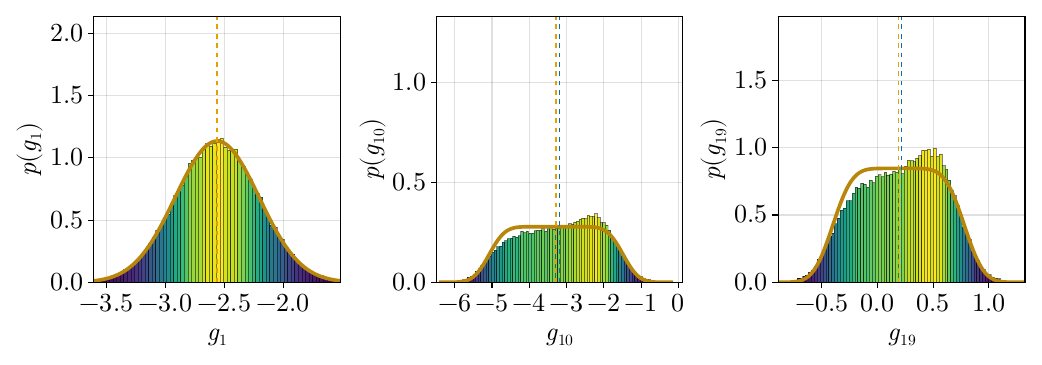}
    \end{subfigure}
    \begin{subfigure}[b]{0.9\linewidth}
      \centering
      \includegraphics[width=0.9\linewidth]{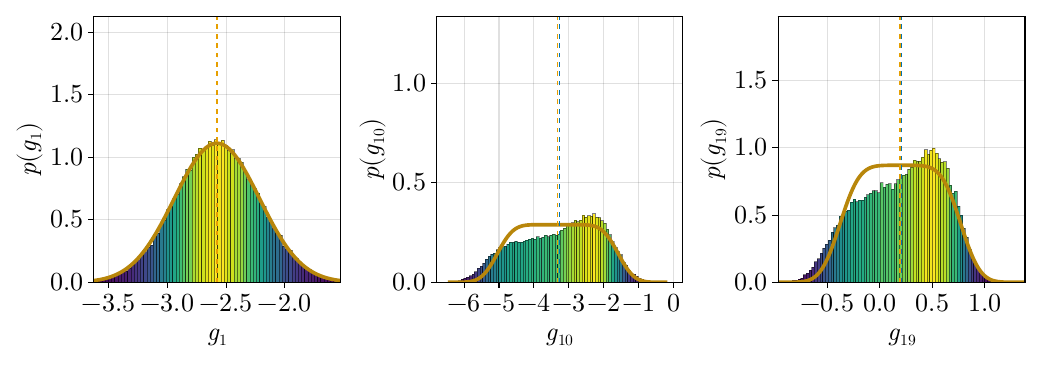}
    \end{subfigure}
       \caption{Nonlinear operator \eqref{nonlinear}. Marginal PDFs of the output function as a function of the number of layers. Shown are results for 1 layer (first row) and 20 layers (last row). The noise amplitude here is set to \(\beta=1.5\).}
       \label{fig:NonlinOps}
  \end{figure}

In this section, we study the accuracy of the linearized leaky ReLU approximation
of the PDF and the statistical moments. To this end, we first generate \( 100,000 \) realizations of the random input functions \( f(x) \) shown in Figure \ref{fig:variation_of_input_perturbations} (i.e., \( 100,000 \) samples of  
the random vector \( \bm{f} = \bm{\mu} + \bm{z} \)), and compute a Monte Carlo (MC) approximation of the statistical properties of the neural network output, for both nonlinear and linear operators in equations \eqref{linear} and \eqref{nonlinear}, respectively. In Figure \ref{fig:cornerplot_nonLin} we plot one- and two-point joint PDFs of  \(g_1, g_6, g_{11}, g_{16}, g_{21}\) (components of the output vector $\bm g$) generated by the benchmark MC simulation. 
In Figures~\ref{fig:nonlin1} and~\ref{fig:nonlin2}, we compare the Monte Carlo benchmark results with the approximate PDFs obtained from the linearized leaky ReLU approximation, i.e., Eq.~\eqref{eq:FINAL_PDF}, in the case of the nonlinear operator.  For noise amplitudes \( \beta\geq 1 \), we observe a slight discrepancy between MC and the analytical PDFs. Specifically, the distributions begin to deviate from a fully symmetric bell-shaped form and exhibit heavier tails in some regions, although in other regions the close agreement observed in the linear case still holds. These results are independent of the number of layers considered in the feed-forward neural network as demonstrated in Figure \ref{fig:NonlinOps}.

\begin{figure}[t]
    \centering
    \includegraphics[width=0.75\linewidth]{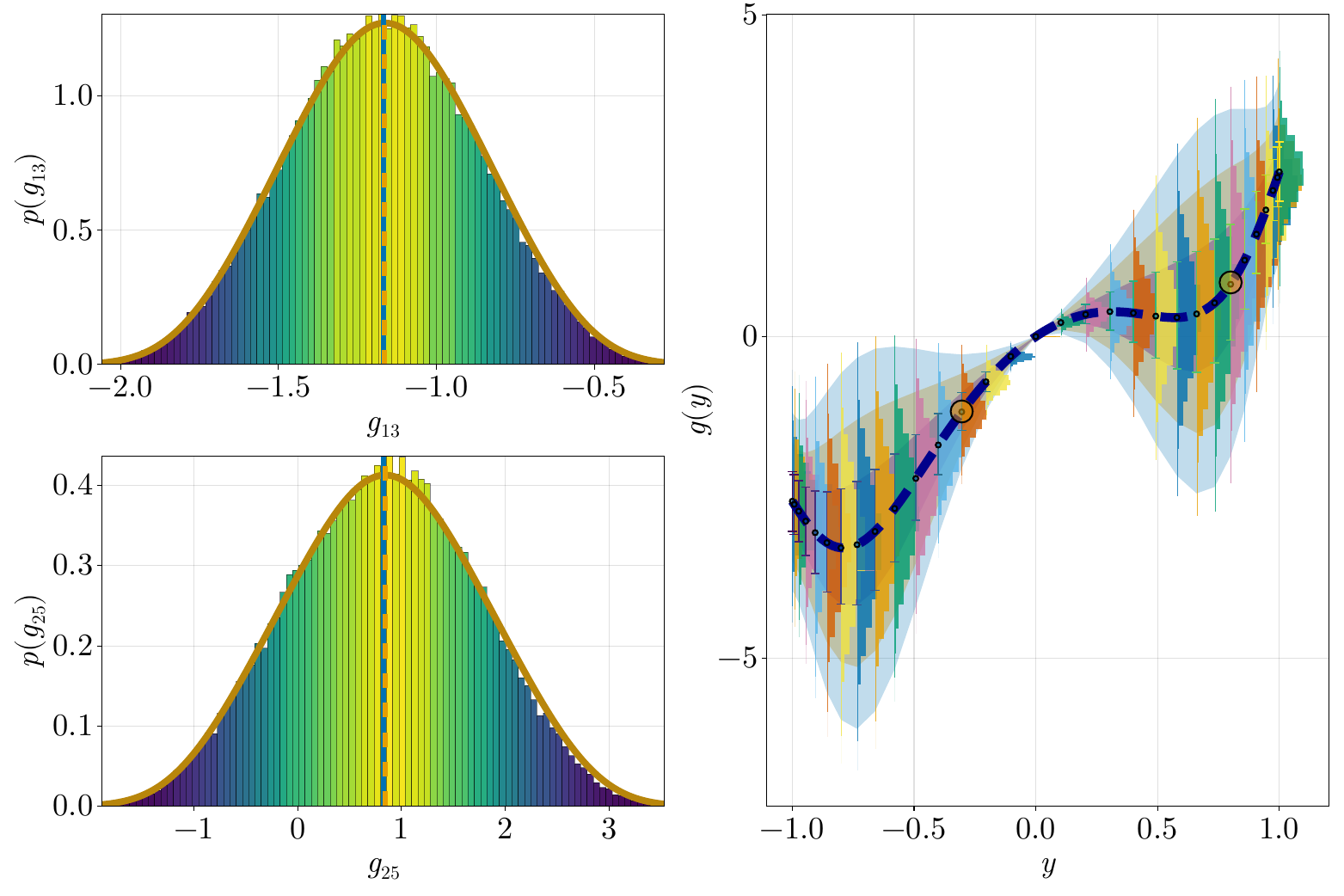}%
       \caption{Linear operator \eqref{linear}. Comparison between the PDFs of  $g_{13}$ and $g_{25}$ as predicted by \eqref{eq:FINAL_PDF} (yellow curve --   linearized leaky ReLU approximation) and the Monte Carlo benchmark. The perturbation amplitude in the input function here is set to $\beta=2$ (see Figure \ref{fig:variation_of_input_perturbations}).  We also plot the confidence intervals $\pm \kappa \sigma$  ($\kappa=1,2,3$) representing the variability of the output $g(y)$ relative to the mapped mean input function.}
       \label{fig:lin1}
  \end{figure}

Despite these slight deviations, the statistical moments of the output obtained from the linearized leaky ReLU approximation -- namely, the mean \eqref{mean}, variance \eqref{var}, and covariance \eqref{cov} -- remain highly accurate, exhibiting strong agreement with the corresponding moments computed with MC (see Table~\ref{tab:statisticalProp}).
 \begin{table}[H]
    \centering
    \begin{tabular}{| m{5em} | m{3cm}| m{3cm} | m{3cm} | m{3cm} | m{3cm} |}
      \hline
      \(\beta = 1.5\) & \(g_6\) & \(g_{11}\) & \(g_{16}\) & \(g_{21}\) \\
      \hline                                               
      \(g_1\)         & \(0.32 (0.32)\)  & \(0.21 (0.21)\)  & \(0.17  (0.15)\)   & \(-0.19  (-0.19)\)      \\
      \hline                                           
      \(g_6\)         &         & \(0.99  (0.99)\)    & \(-0.77  (-0.73)\)   & \(0.87  (0.87)\)        \\
      \hline                                          
      \(g_{11}\)      &         &         & \(-0.82  (-0.77)\)   & \(0.92  (0.92)\)         \\
      \hline                                                  
      \(g_{16}\)      &         &         &           & \(-0.89 (-0.84)\)         \\
      \hline                                                  
    \end{tabular}
    \caption{Nonlinear operator \eqref{nonlinear}. Correlation coefficients computed from analytical expression of moments derived from the linearized leaky ReLU approximation method compared to those numerically found from MC simulations (in parentheses). See Figure \ref{fig:cornerplot_nonLin} for a plot of two-point  PDFs of $g_i$ and $g_j$.}
\label{tab:statisticalProp}
\end{table}
On the other hand, in Figure \ref{fig:lin1} we observe excellent good agreement between the PDFs computed numerically and those obtained from the analytical approximation in the case of the linear operator, regardless of the number of layers used in training the neural network, and for significantly large perturbation amplitude. 

\begin{figure}[t]
    \centering
    \begin{subfigure}[b]{0.9\linewidth}
      \centering
     \caption{1 layer}
      \includegraphics[width=0.9\linewidth]{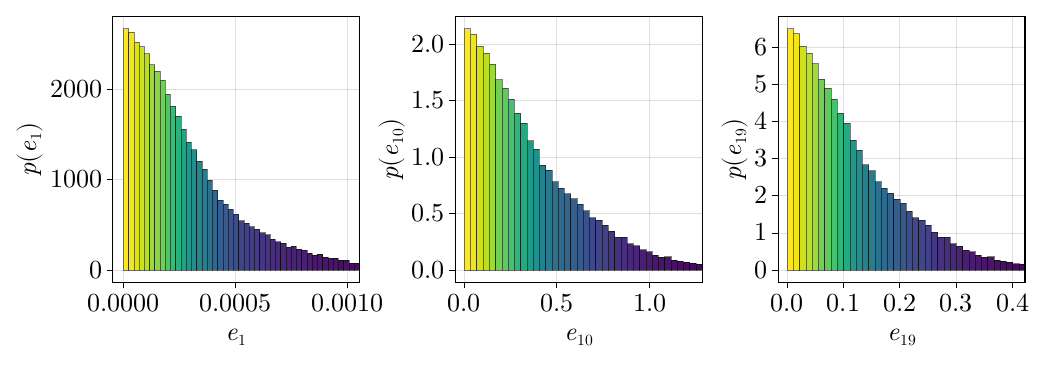}
    \end{subfigure}
    \hfill
    \begin{subfigure}[b]{0.9\linewidth}
      \centering
           \caption{5 layers}
      \includegraphics[width=0.9\linewidth]{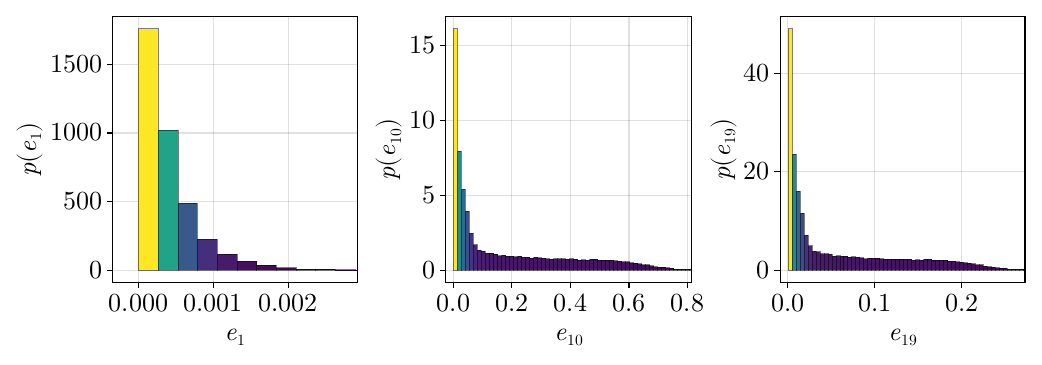}
    \end{subfigure}
    \caption{Nonlinear operator \eqref{nonlinear}. PDFs of the error \eqref{eq:error} for noise amplitude $\beta=1.5$ and different number of layers. It is seen that as the number of layer increases the PDF of the error tends to concentrate around zero. This suggests that linearized leaky ReLU approximations of MLP networks are most effective for deep networks. }
       \label{fig:NonlinOp_ErrResid}
  \end{figure}

The reason behind the surprisingly accurate PDFs and statistical moments obtained from the linearized leaky ReLU network approximation lies in the manner in which the approximation error propagates through the layers of the network (see \ref{app:error}). In Figure~\ref{fig:NonlinOp_ErrResid}, we plot the PDF of the error defined in~\eqref{eq:error}, obtained via Monte Carlo sampling. 
As expected, the numerical results show that the error is proportional to the perturbation amplitude in the input vector. Furthermore, the error seems to decrese with the number of neural network layers $L$. This suggests that linearized leaky ReLU approximations of MLP networks are 
most effective for deep networks. 

\subsection{Gaussian copula PDF surrogates}
\begin{figure}[t]
    \centering
    \begin{subfigure}[h]{0.32\linewidth}
      \centering
      \includegraphics[width=\linewidth]{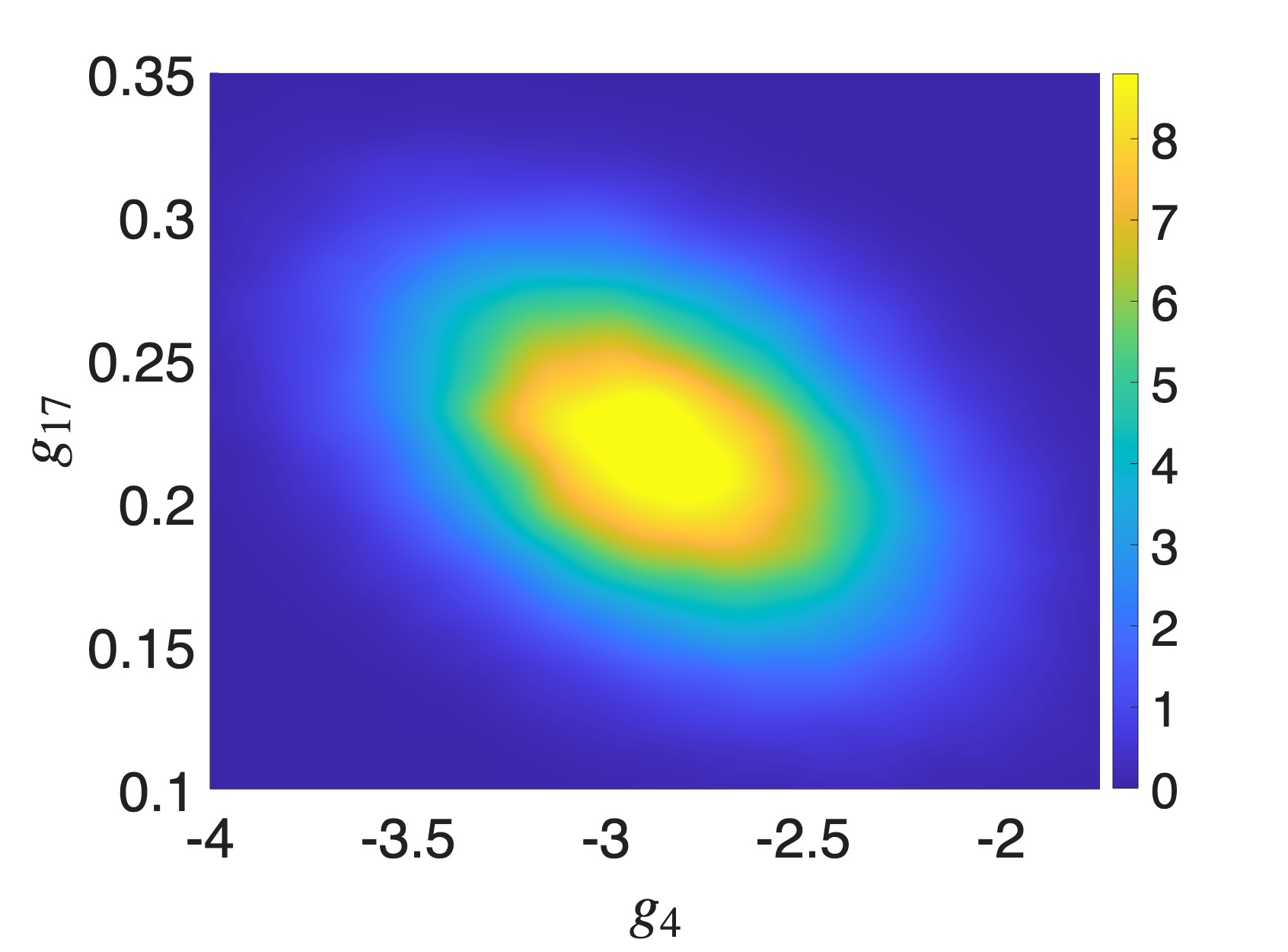}
    \end{subfigure}
    \hfill
    \begin{subfigure}[h]{0.32\linewidth}
      \centering
      \includegraphics[width=\linewidth]{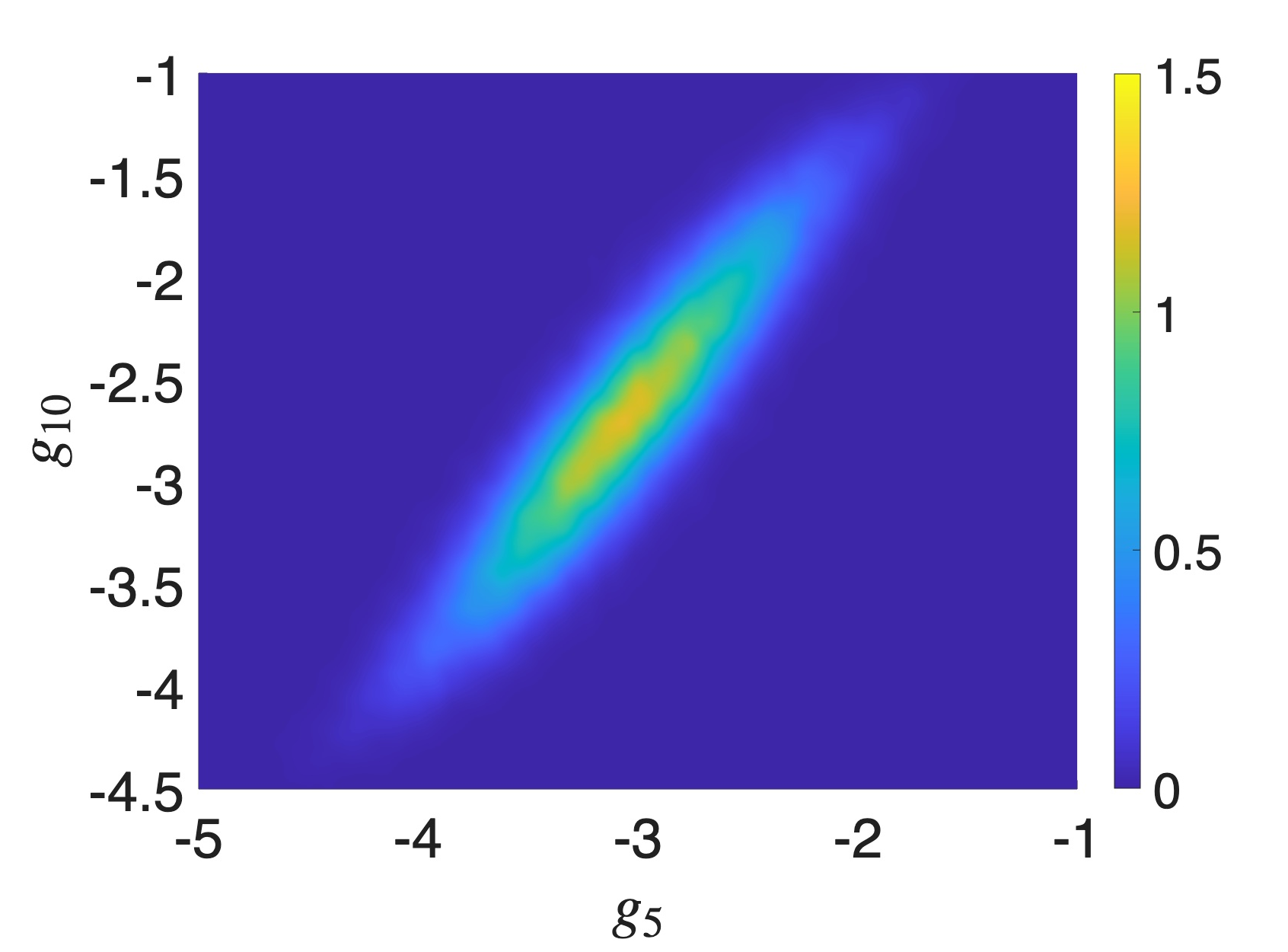}
    \end{subfigure}
    \hfill
    \begin{subfigure}[h]{0.32\linewidth}
      \centering
      \includegraphics[width=\linewidth]{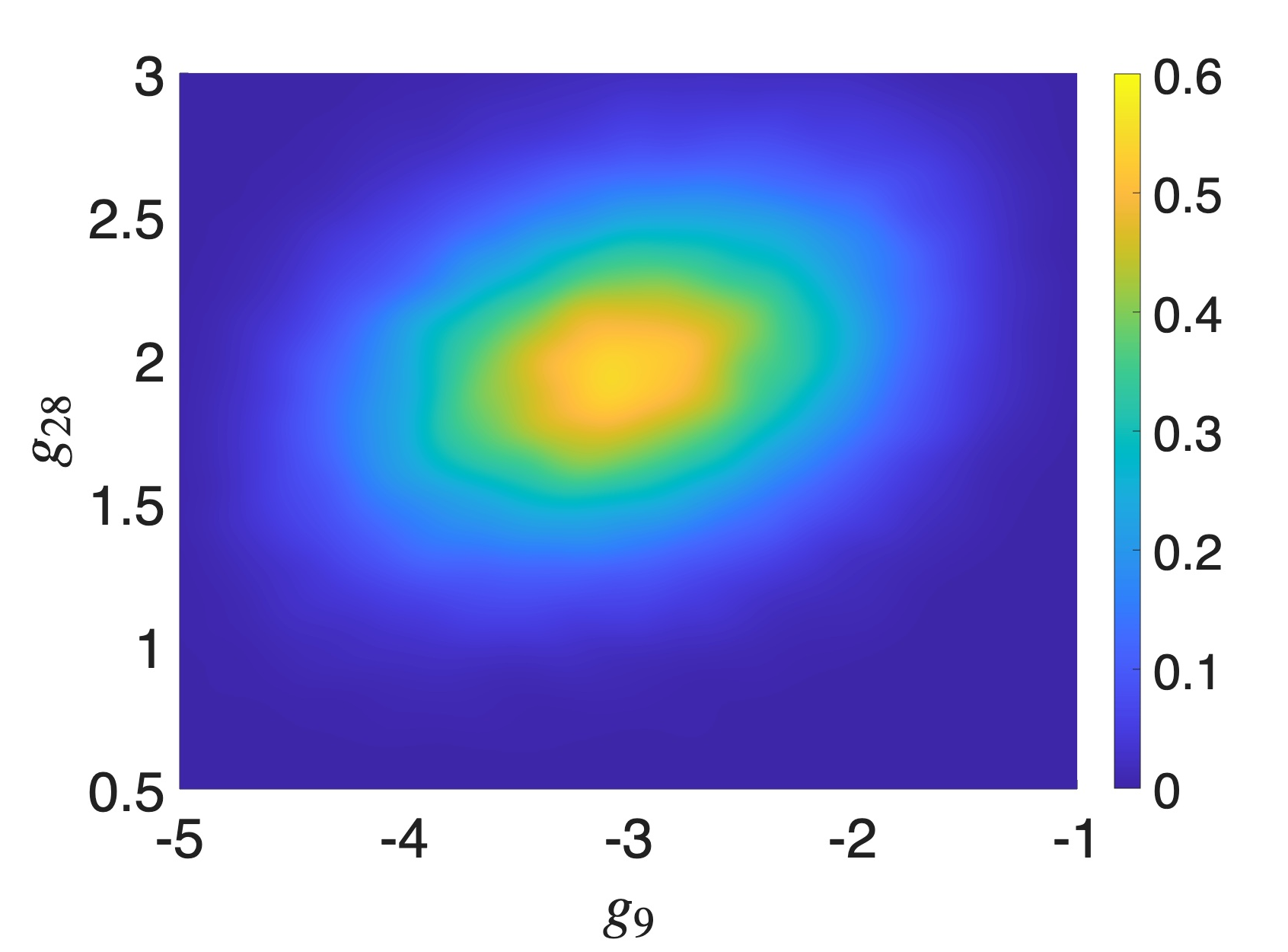}
    \end{subfigure}
    \\
    \begin{subfigure}[h]{0.32\linewidth}
      \centering
      \includegraphics[width=\linewidth]{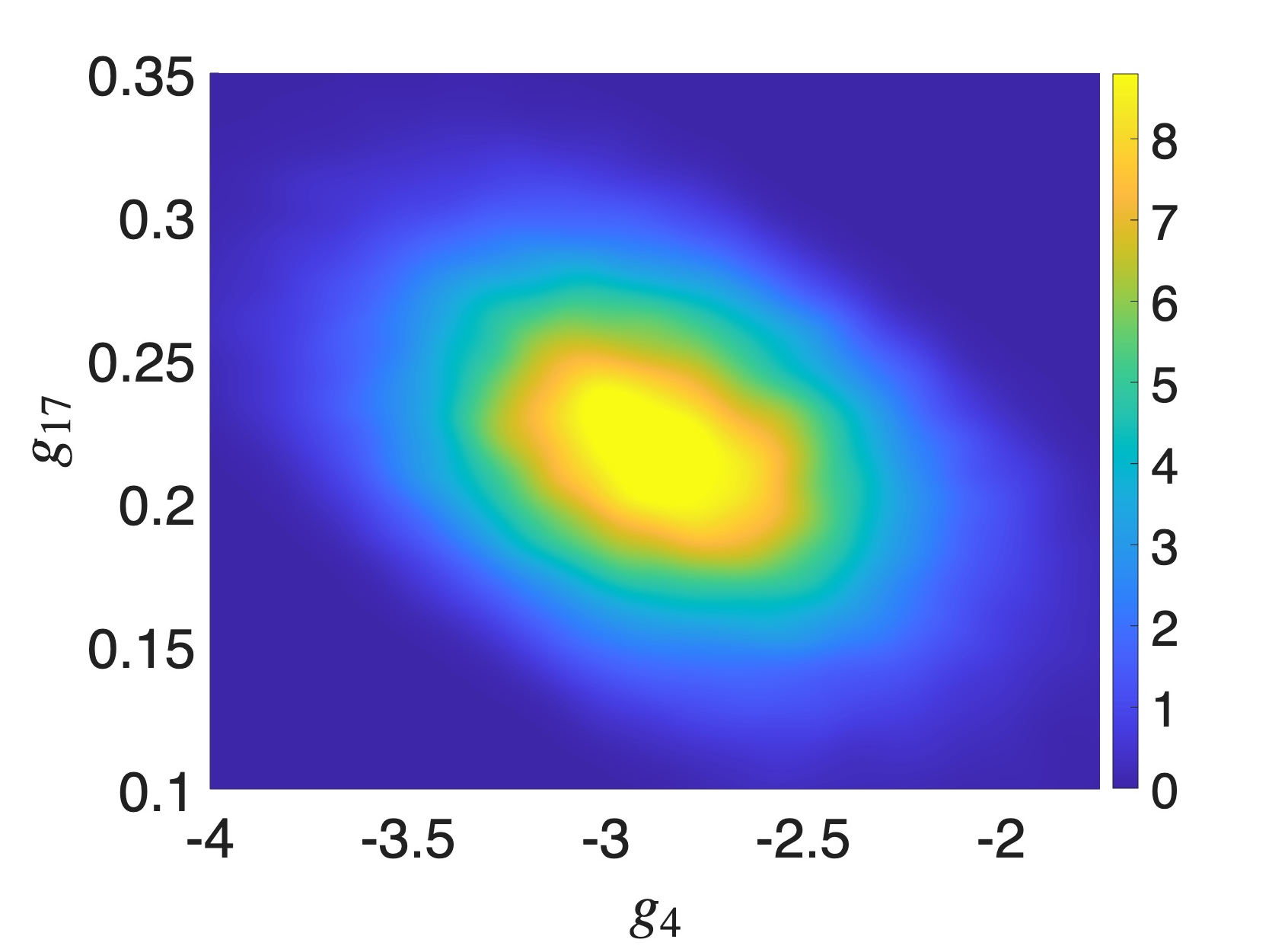}
    \end{subfigure}
    \hfill
    \begin{subfigure}[h]{0.32\linewidth}
      \centering
      \includegraphics[width=\linewidth]{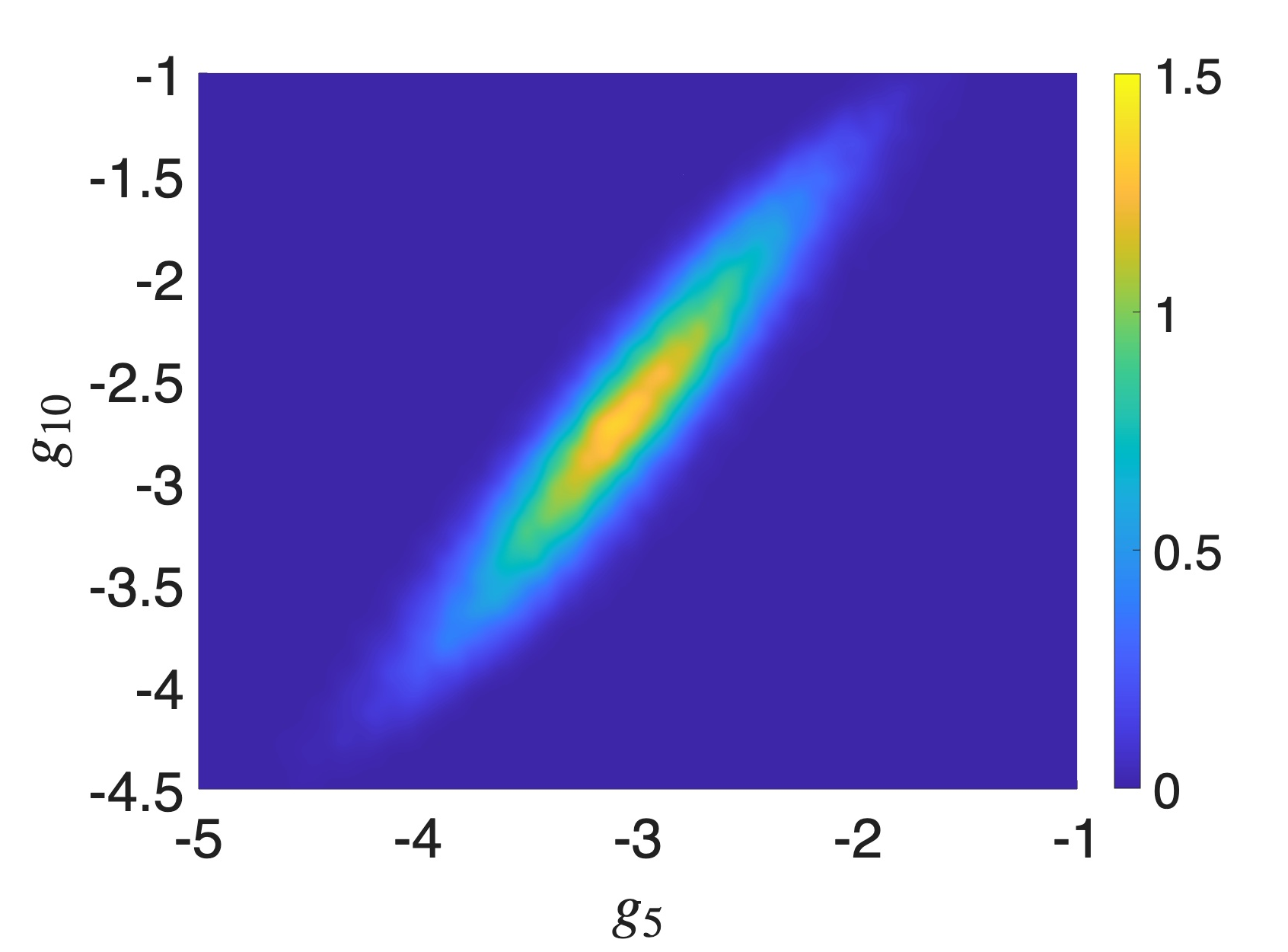}
    \end{subfigure}
    \hfill
    \begin{subfigure}[h]{0.32\linewidth}
      \centering
      \includegraphics[width=\linewidth]{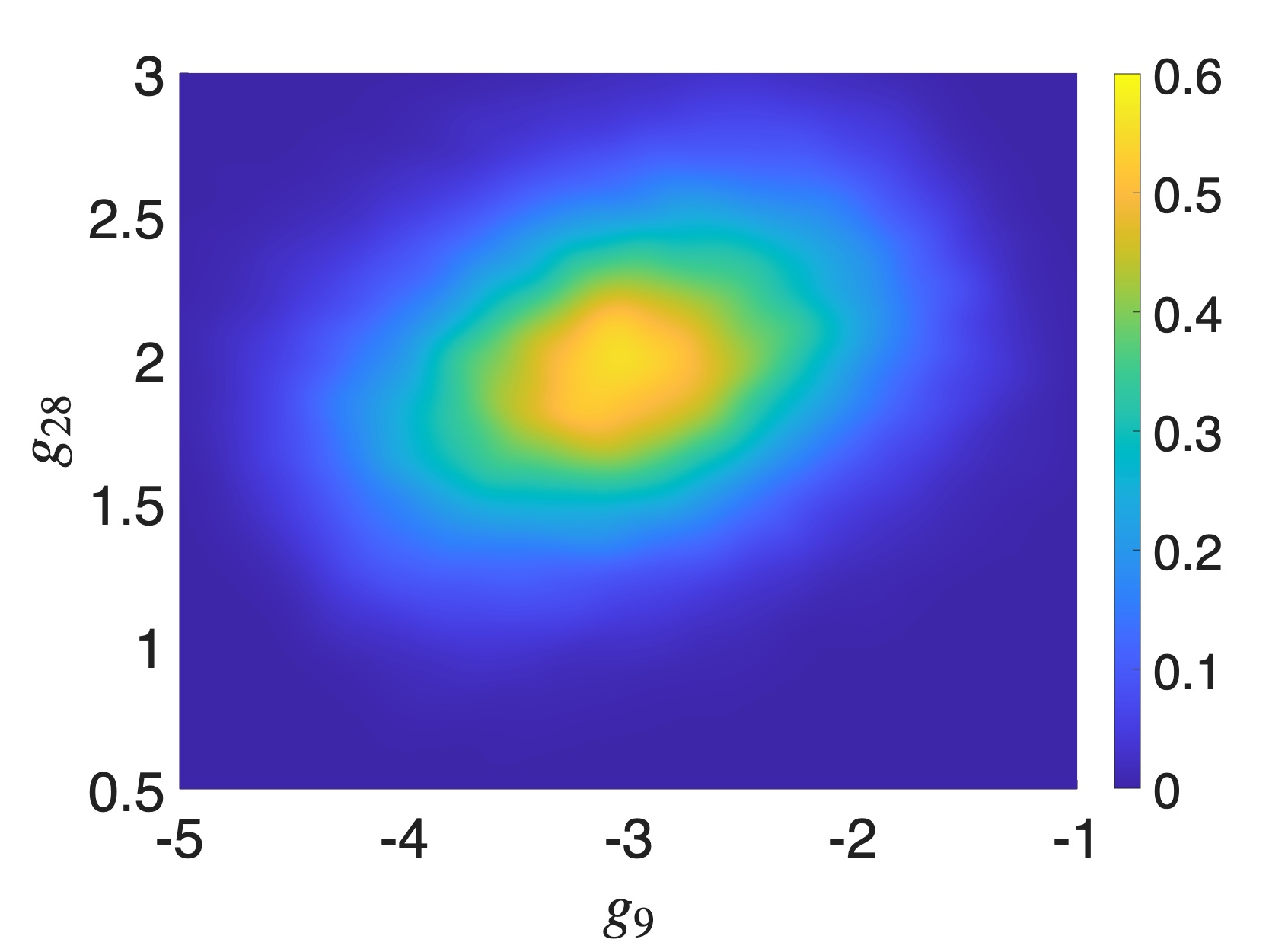}
    \end{subfigure}
    \\
       \caption{Linear Operator \eqref{linear}. Comparison of the 2-point PDFs generated by Gaussian copula (top) vs. the  benchmark 2-point PDFs obtained from MC. }
       \label{fig:2-point_PDF_results_Linop}
  \end{figure}

\begin{figure}[t]
    \centering
    \begin{subfigure}[h]{0.32\linewidth}
      \centering
      \includegraphics[width=\linewidth]{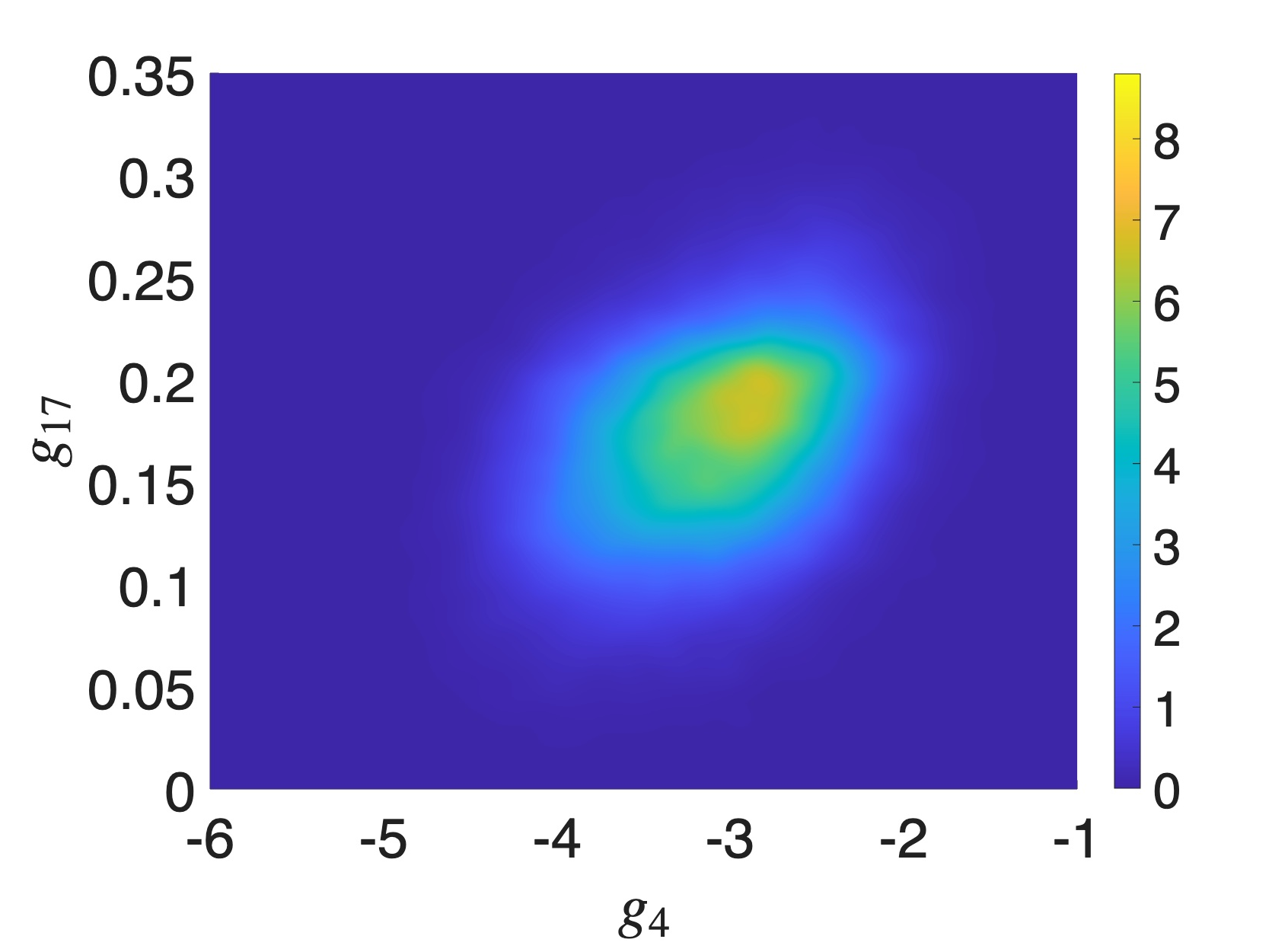}
    \end{subfigure}
    \hfill
    \begin{subfigure}[h]{0.32\linewidth}
      \centering
      \includegraphics[width=\linewidth]{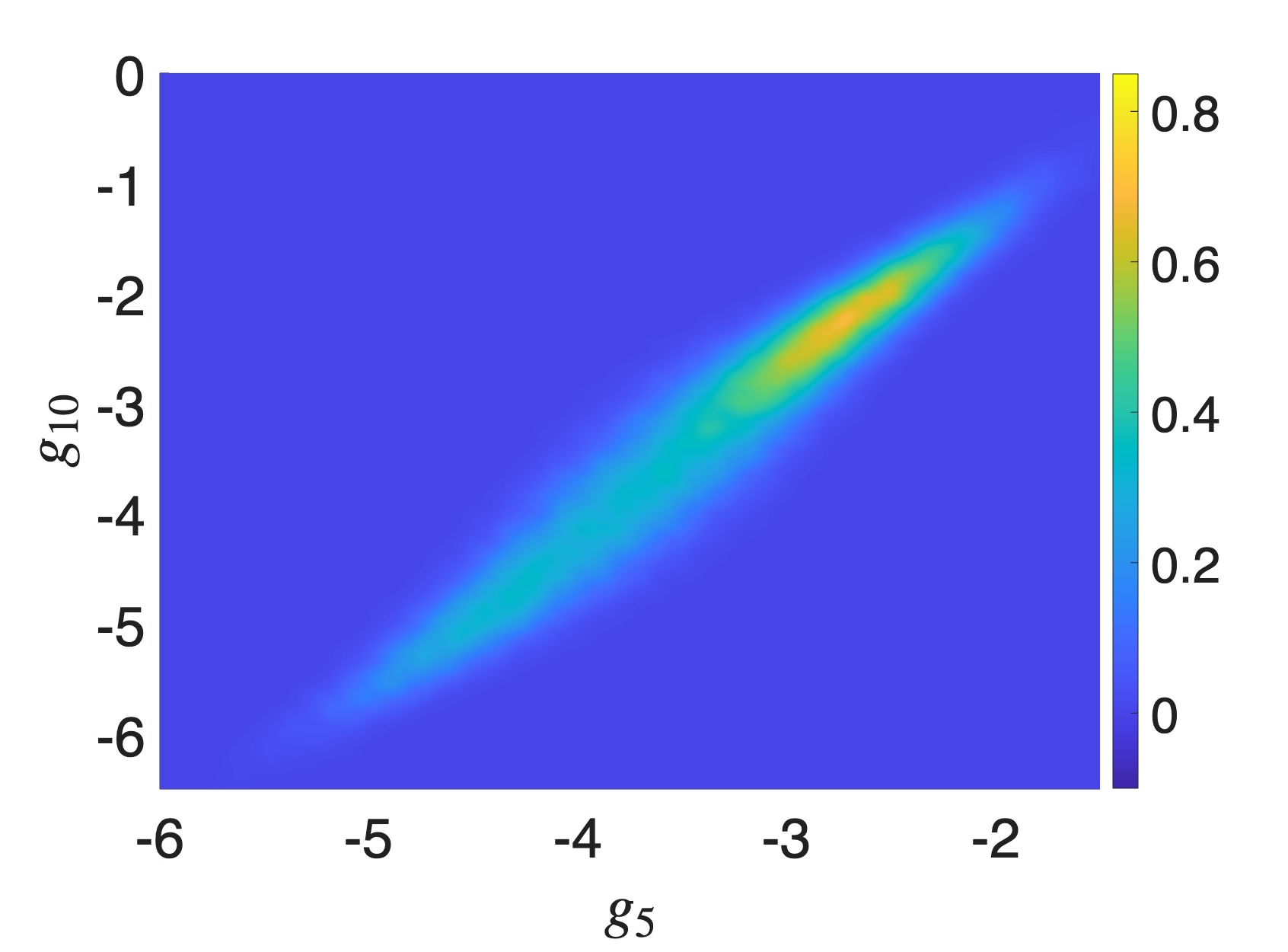}
    \end{subfigure}
    \hfill
    \begin{subfigure}[h]{0.32\linewidth}
      \centering
      \includegraphics[width=\linewidth]{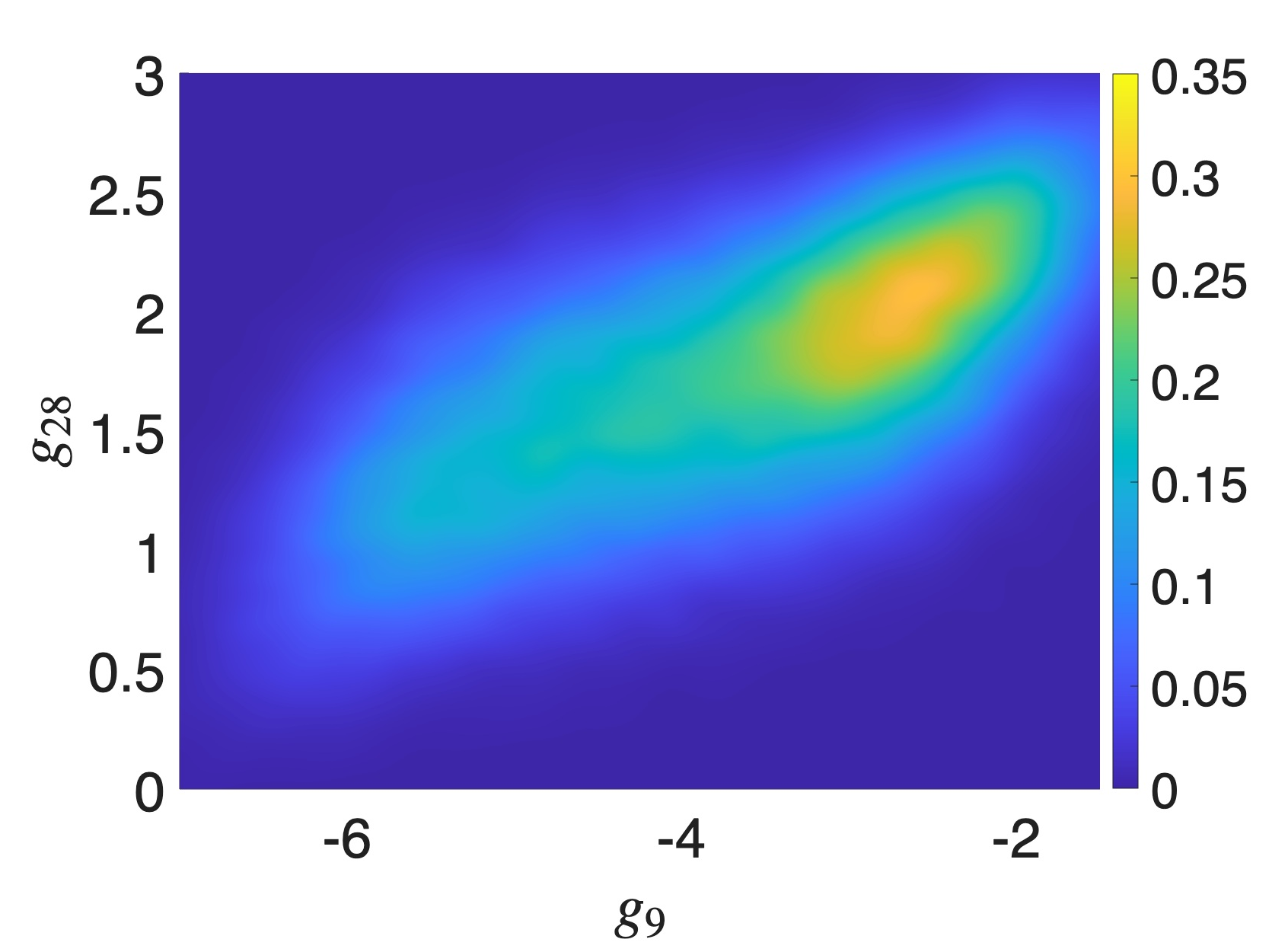}
    \end{subfigure}
    \\
    \begin{subfigure}[h]{0.32\linewidth}
      \centering
      \includegraphics[width=\linewidth]{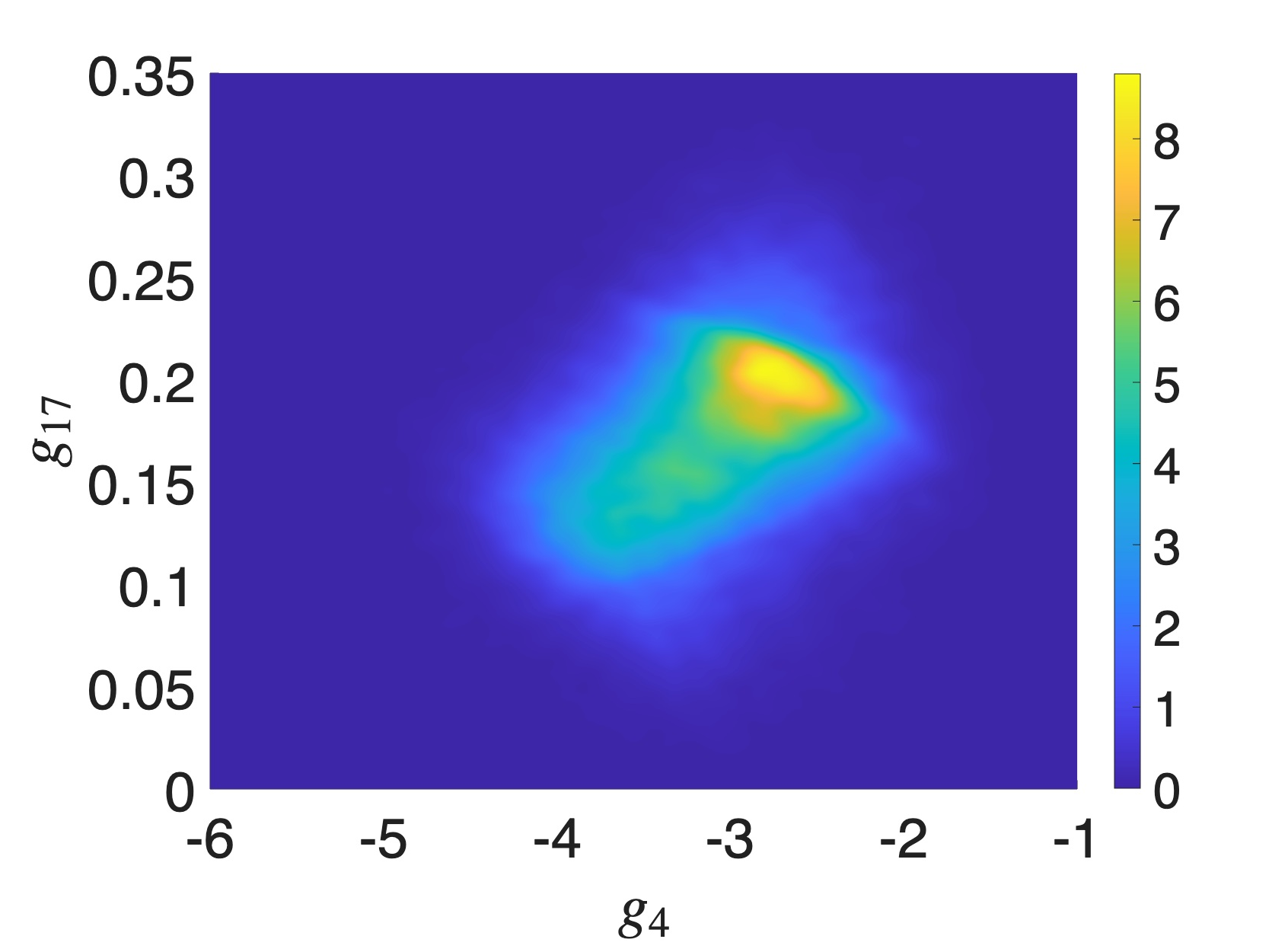}
    \end{subfigure}
    \hfill
    \begin{subfigure}[h]{0.32\linewidth}
      \centering
      \includegraphics[width=\linewidth]{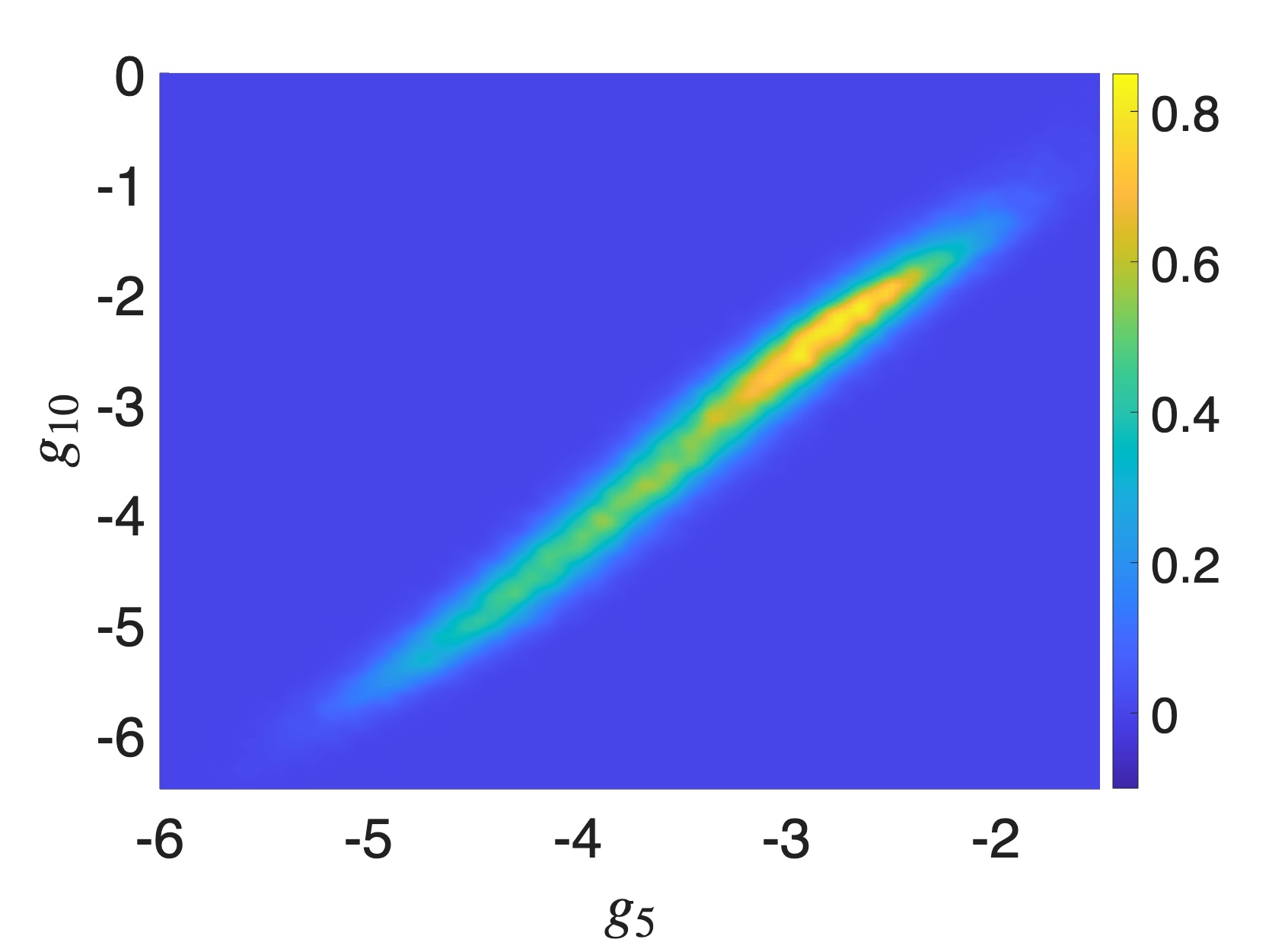}
    \end{subfigure}
    \hfill
    \begin{subfigure}[h]{0.32\linewidth}
      \centering
      \includegraphics[width=\linewidth]{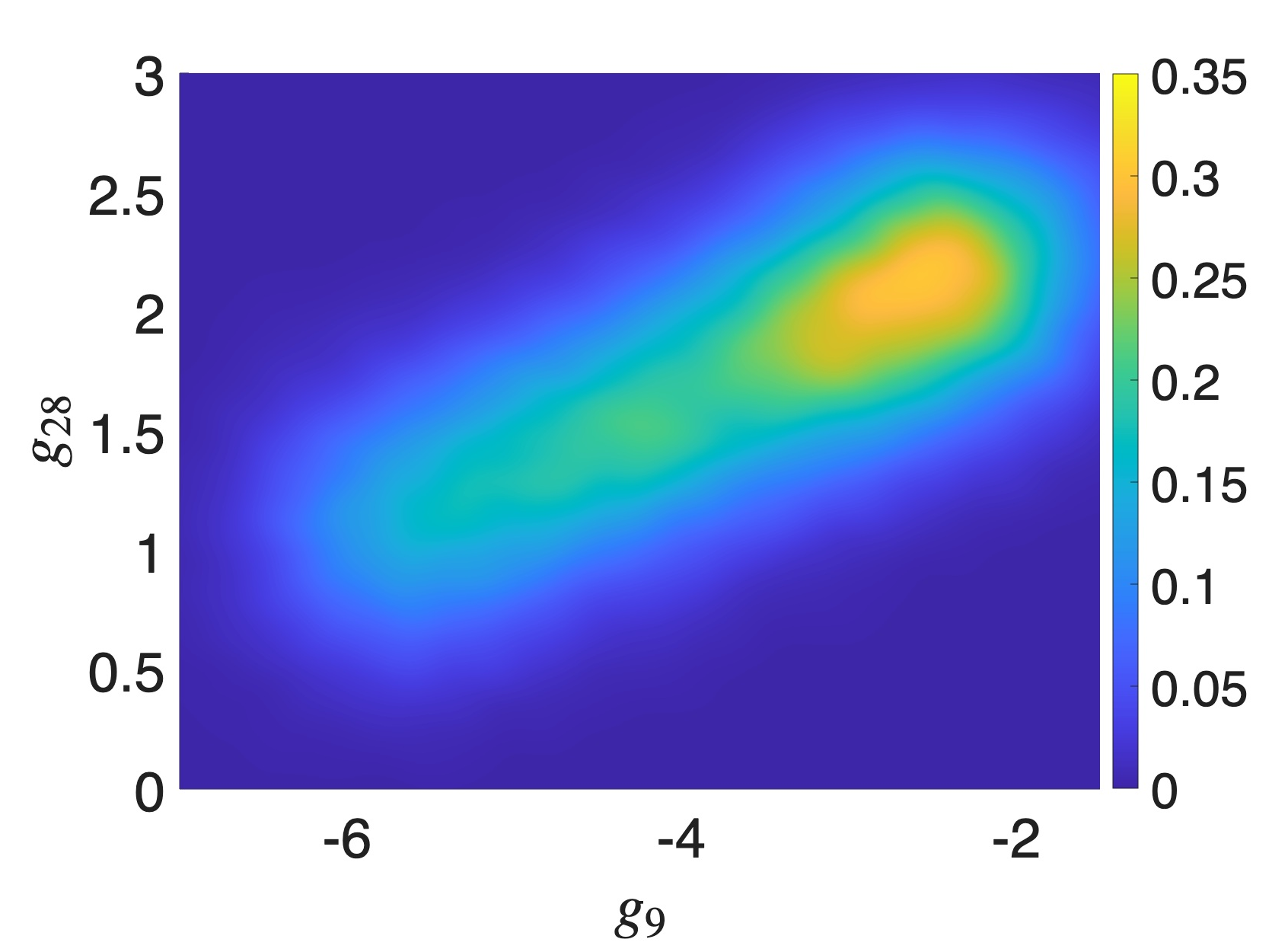}
    \end{subfigure}
    \\
       \caption{Nonlinear operator \eqref{nonlinear}. Comparison of the 2-point PDFs generated by Gaussian copula (first row) vs. the  benchmark 2-point PDFs obtained from MC (second row). }
       \label{fig:2-point_PDF_results_Nonlinop}
  \end{figure}

In this section we study the accuracy of Gaussian copulas to approximate the 
{\em full} joint PDF $p(\bm g)$ of the neural network output. As discussed in 
section \ref{sec:copula}, a Gaussian copula can be constructed using the 
marginal PDFs  $p(g_j)$ and the correlation function $\mathbb{E}\{g_ig_j\}$. 
These quantities can accurately approximated by the analytical expressions 
in \eqref{eq:FINAL_PDF} and \eqref{corr}. 
 \begin{figure}[t]
    \centering
    \begin{subfigure}[b]{0.4\linewidth}
      \centering
       \caption{Gaussian Copula}
      \includegraphics[width=\linewidth]{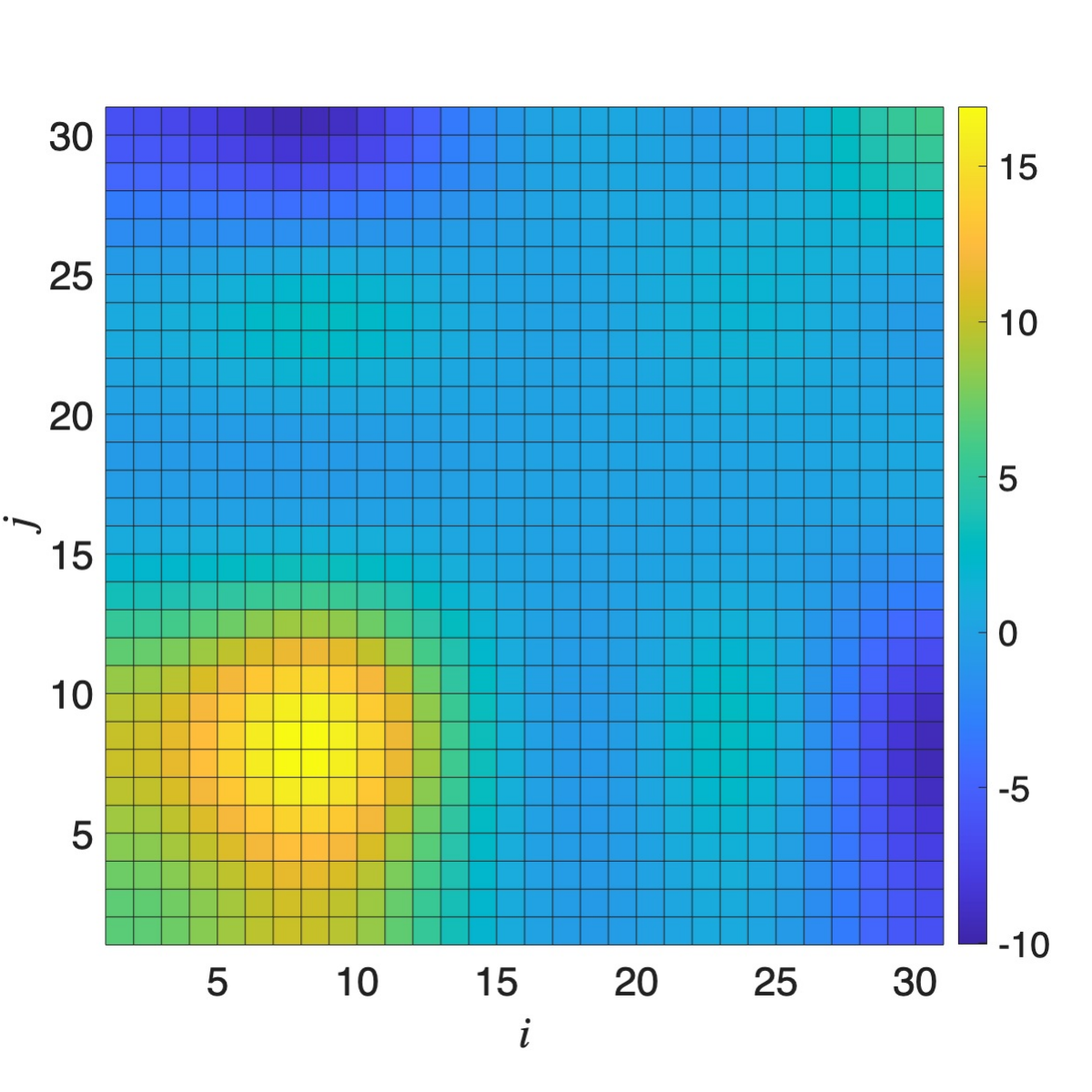}
      \label{fig:CopulaCorrCoeffs_Nonlinop}
    \end{subfigure}
    \hspace{1cm}
    \begin{subfigure}[b]{0.4\linewidth}
      \centering
       \caption{Benchmark}
      \includegraphics[width=\linewidth]{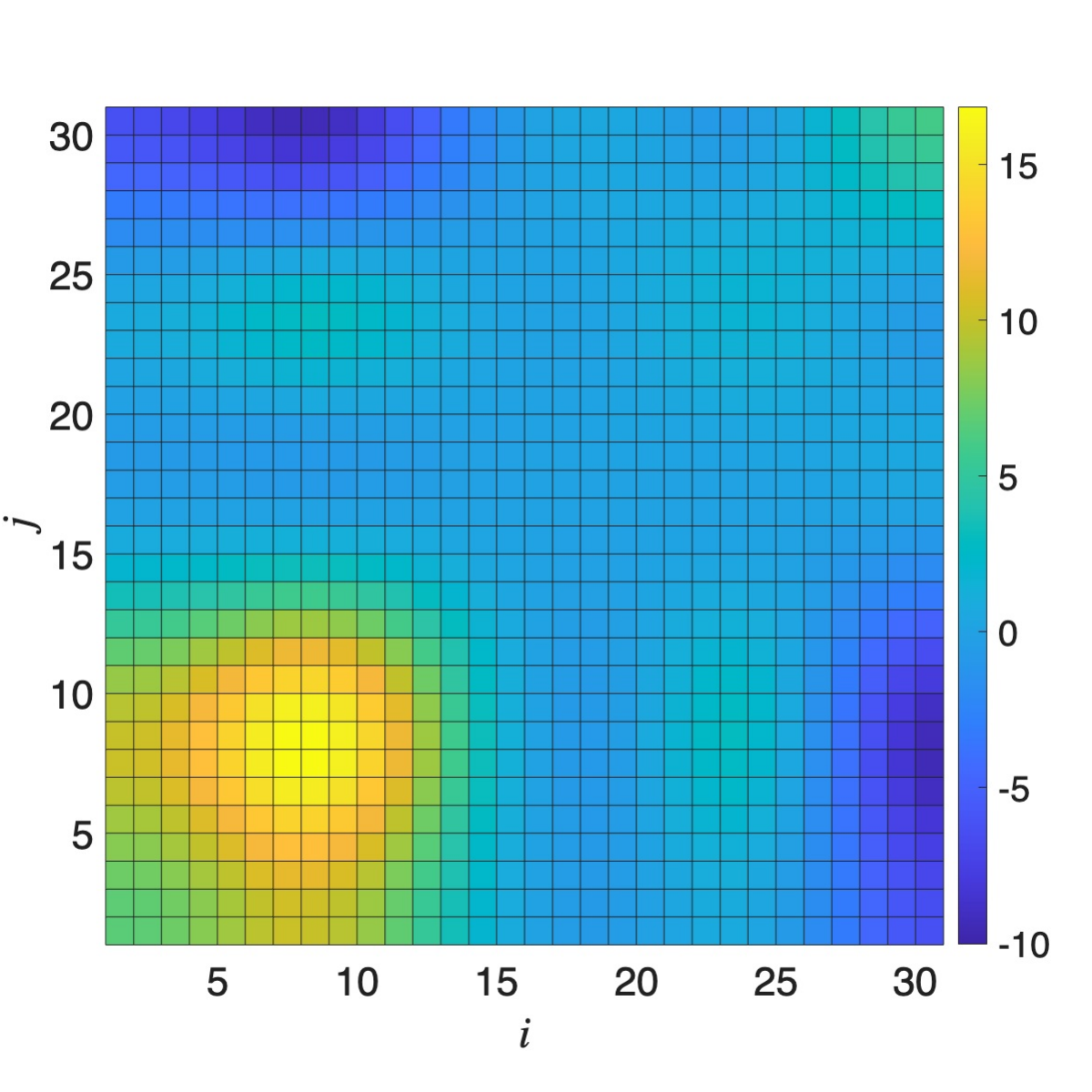}
      \label{fig:BenchmarkCorrCoeffs_Nonlinop}
    \end{subfigure}
    \hfill
       \caption{Nonlinear Operator \eqref{nonlinear}. Comparison between the correlation matrix $\mathbb{E}\{g_ig_j\}$ obtained from the Gaussian copula vs. the benchmark correlation obtained from MC.}
       \label{fig:CorrCoeffsComparison_Nonlinop}
  \end{figure}
In Figure \ref{fig:CorrCoeffsComparison_Nonlinop}, we show that the entries of the correlation matrix $\mathbb{E}\{g_ig_j\}$ obtained by the MC benchmark and the Gaussian copula for the nonlinear operator are nealy identical, suggesting proper training of the copula. In Figures~\ref{fig:2-point_PDF_results_Linop} and~\ref{fig:2-point_PDF_results_Nonlinop}, we compare various two-point PDFs of the neural network output obtained from the Monte Carlo benchmark and the Gaussian copula approximation, for the linear and nonlinear operators, respectively. It is seen that such PDFs are in excellent agreement. 

\begin{figure}[t]
    \centering
    \begin{subfigure}[h]{0.32\linewidth}
      \centering
      \includegraphics[width=\linewidth]{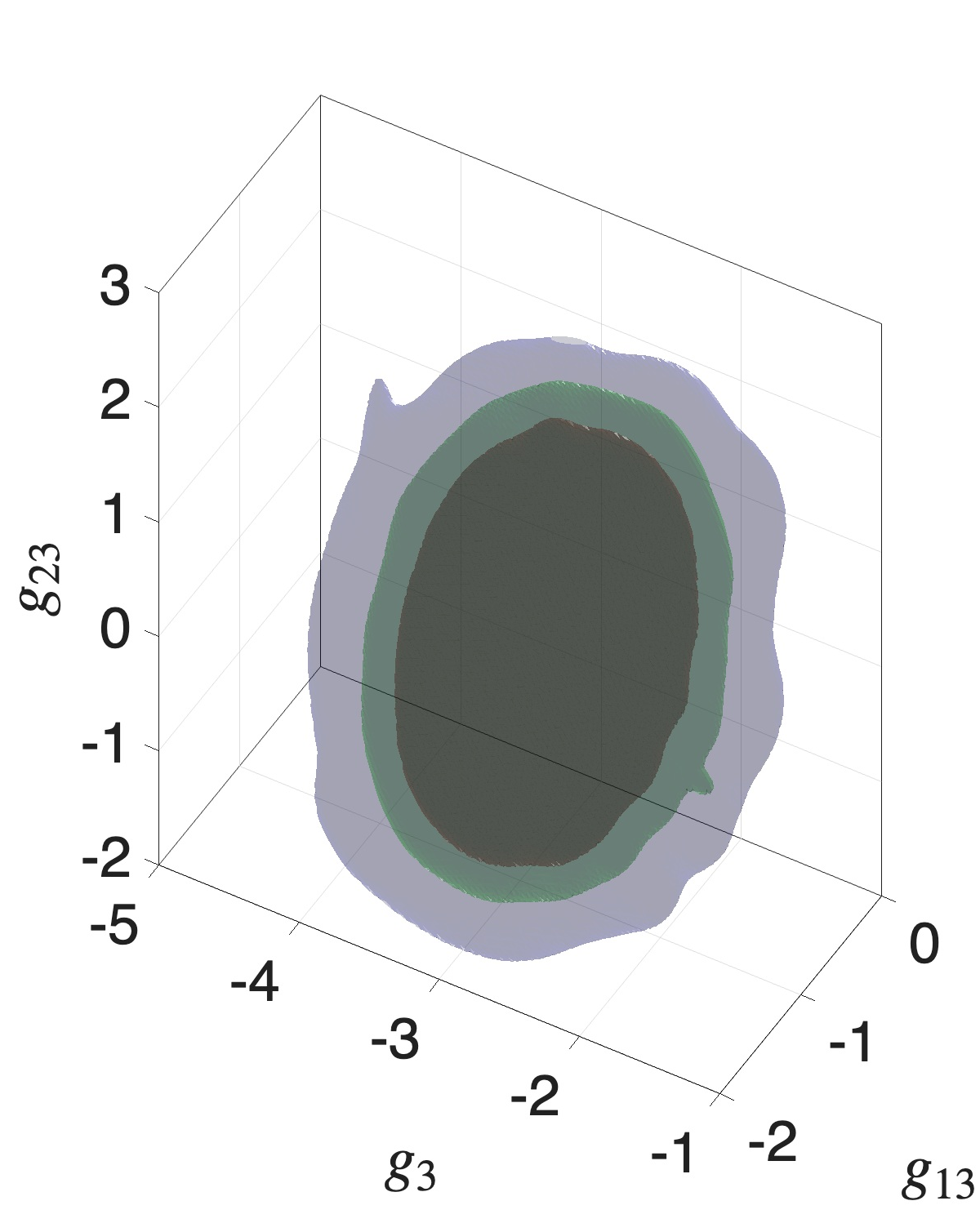}
    \end{subfigure}
    \hfill
    \begin{subfigure}[h]{0.32\linewidth}
      \centering
      \includegraphics[width=\linewidth]{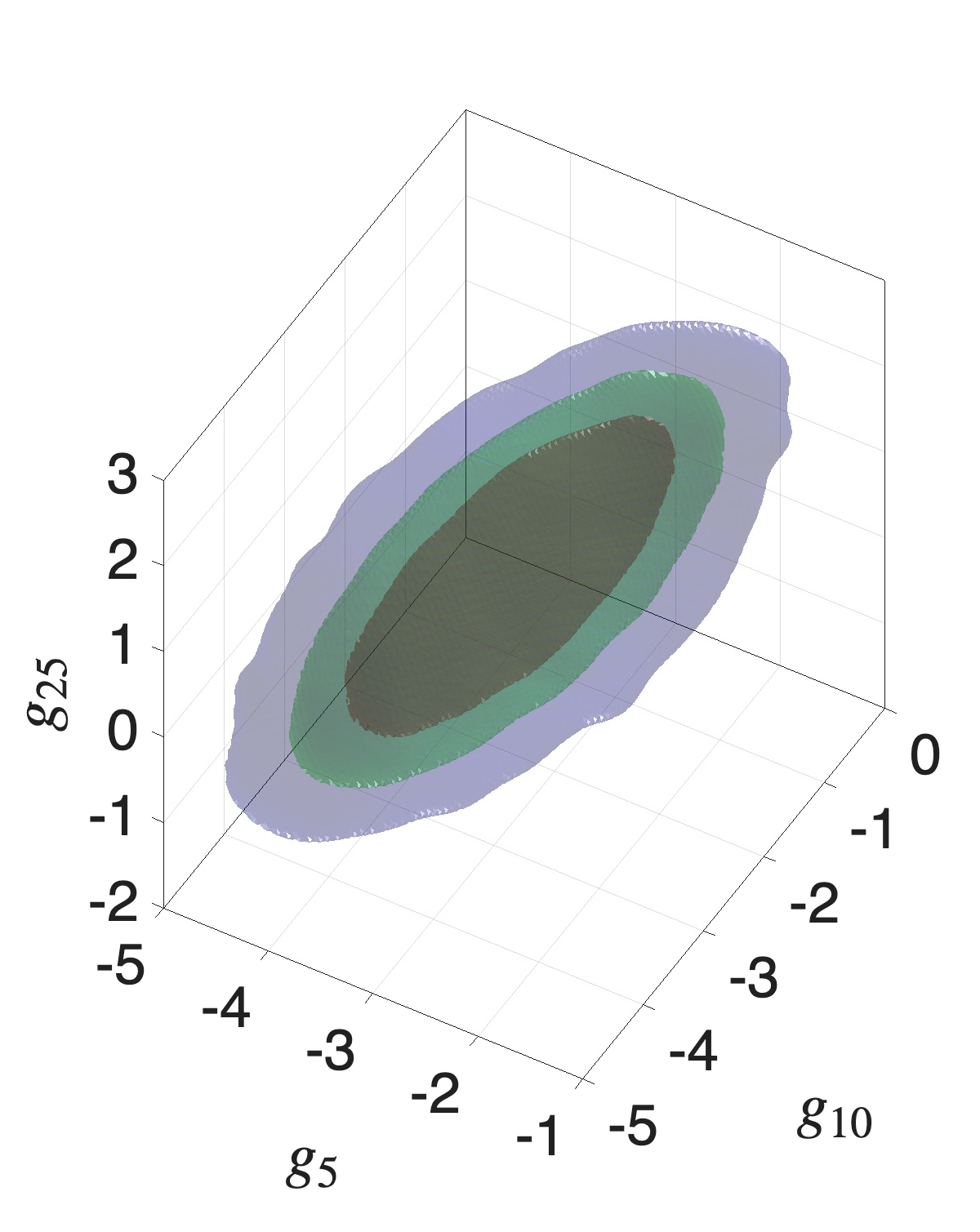}
    \end{subfigure}
    \hfill
    \begin{subfigure}[h]{0.32\linewidth}
      \centering
      \includegraphics[width=\linewidth]{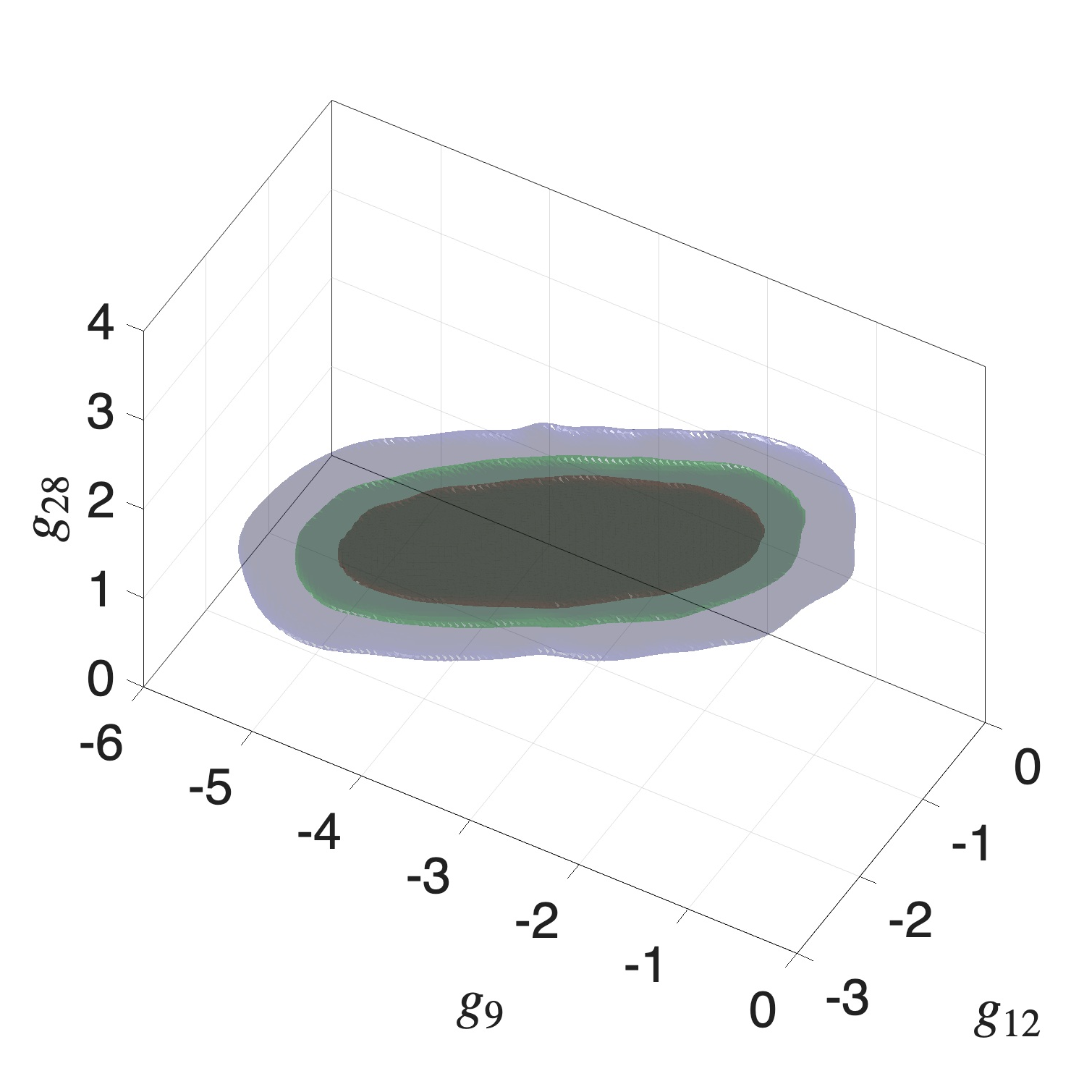}
    \end{subfigure}
    \\
    \begin{subfigure}[h]{0.32\linewidth}
      \centering
      \includegraphics[width=\linewidth]{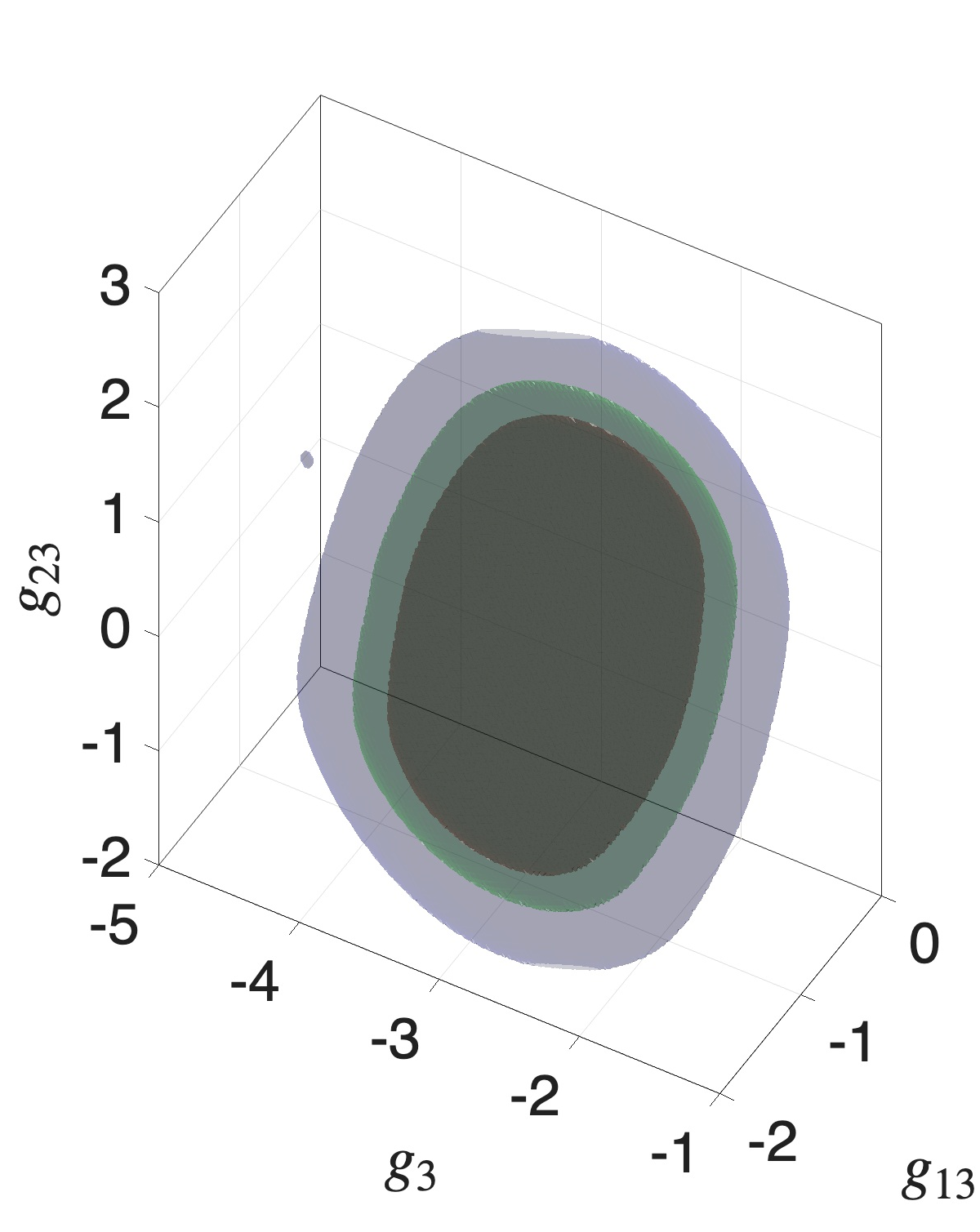}
    \end{subfigure}
    \hfill
    \begin{subfigure}[h]{0.32\linewidth}
      \centering
      \includegraphics[width=\linewidth]{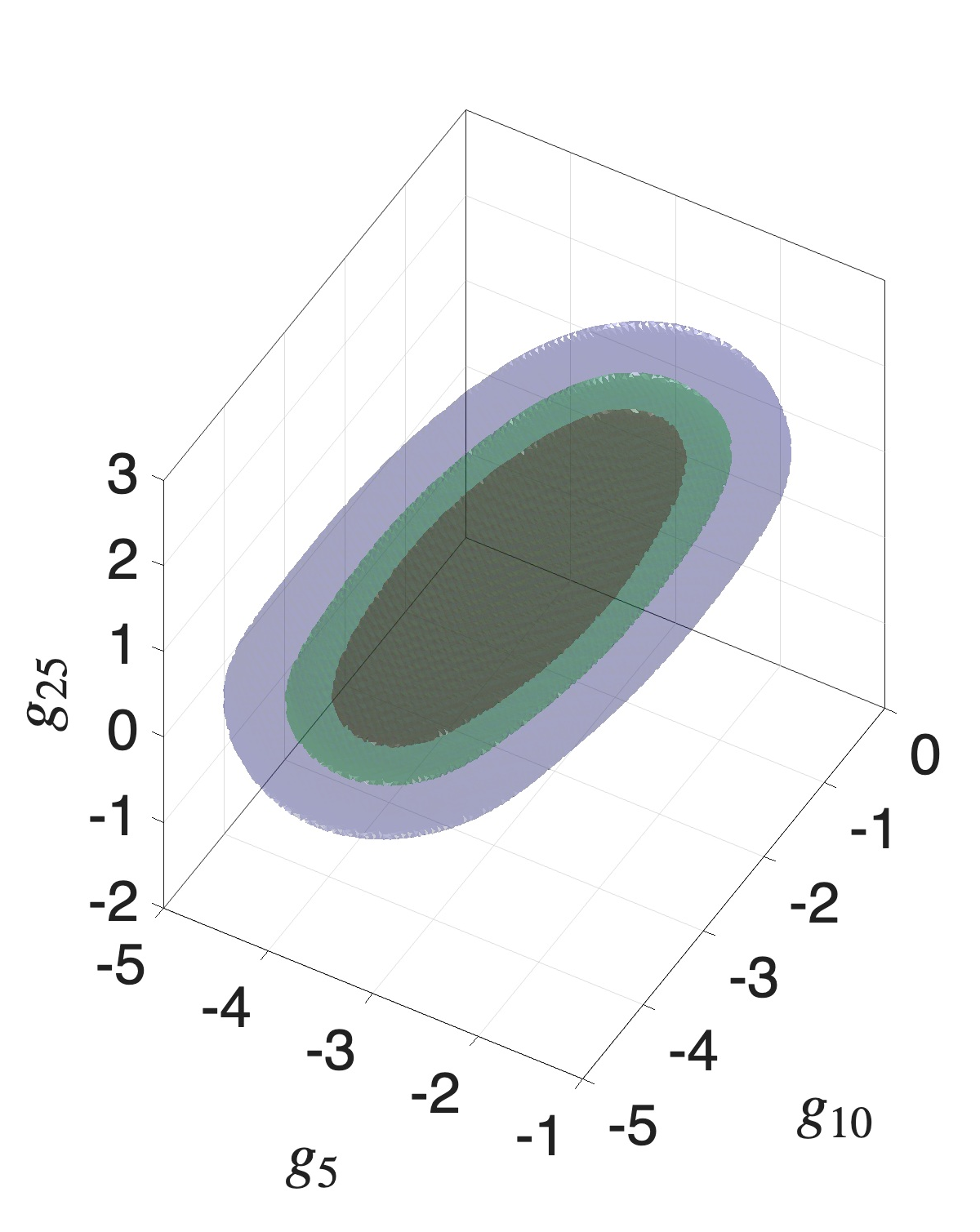}
    \end{subfigure}
    \hfill
    \begin{subfigure}[h]{0.32\linewidth}
      \centering
      \includegraphics[width=\linewidth]{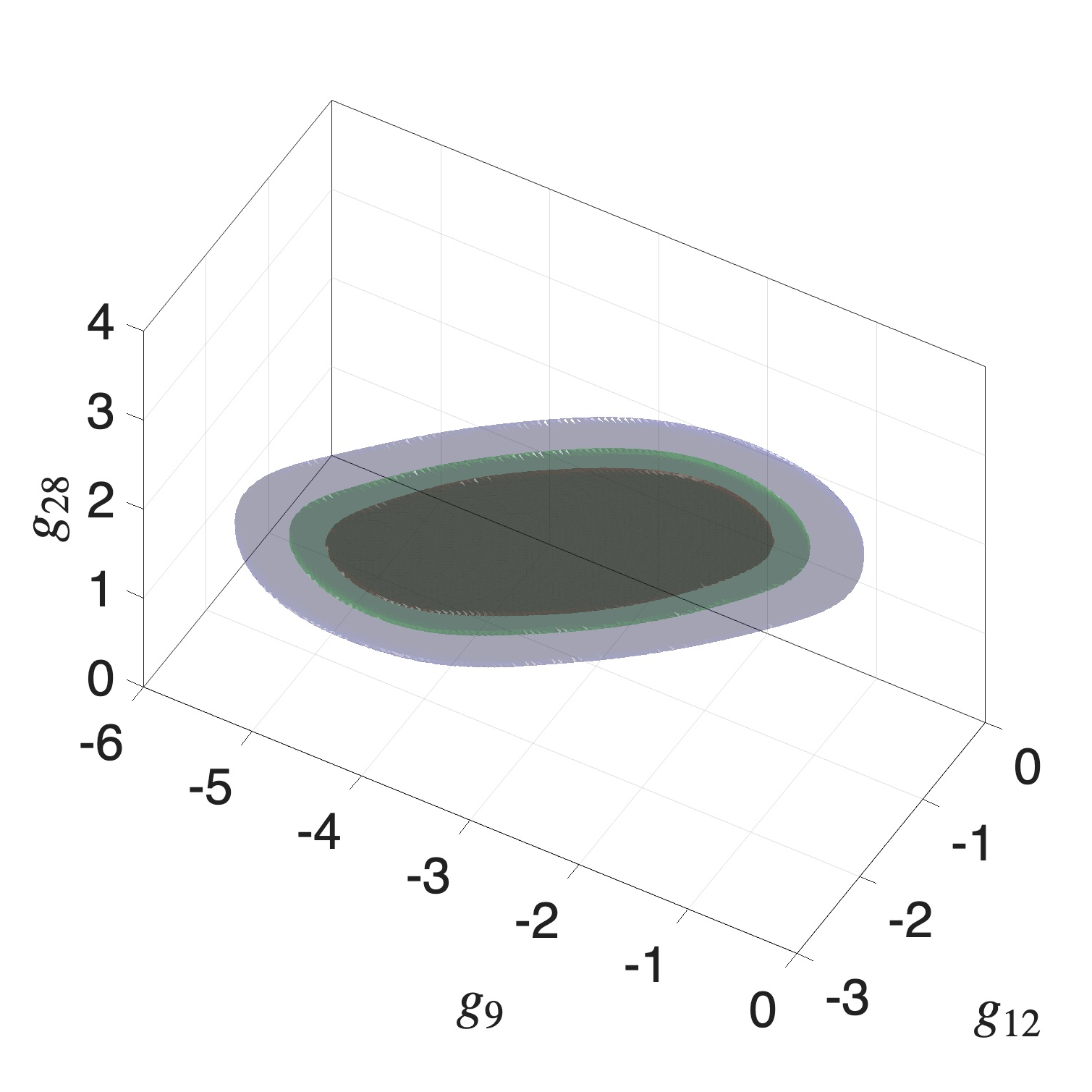}
    \end{subfigure}
    \caption{Linear operator \eqref{linear}. Comparison of the 3-point PDFs generated by Gaussian copula (first row) vs. the  benchmark 3-point PDFs obtained from MC (second row).  Shown are $\{0.005, 0.1, 0.4\}$ level sets. }
    \label{fig:3-point_PDF_results_Linop}
  \end{figure}

\begin{figure}[h]
    \centering
    \begin{subfigure}[h]{0.31\linewidth}
      \centering
      \includegraphics[width=\linewidth]{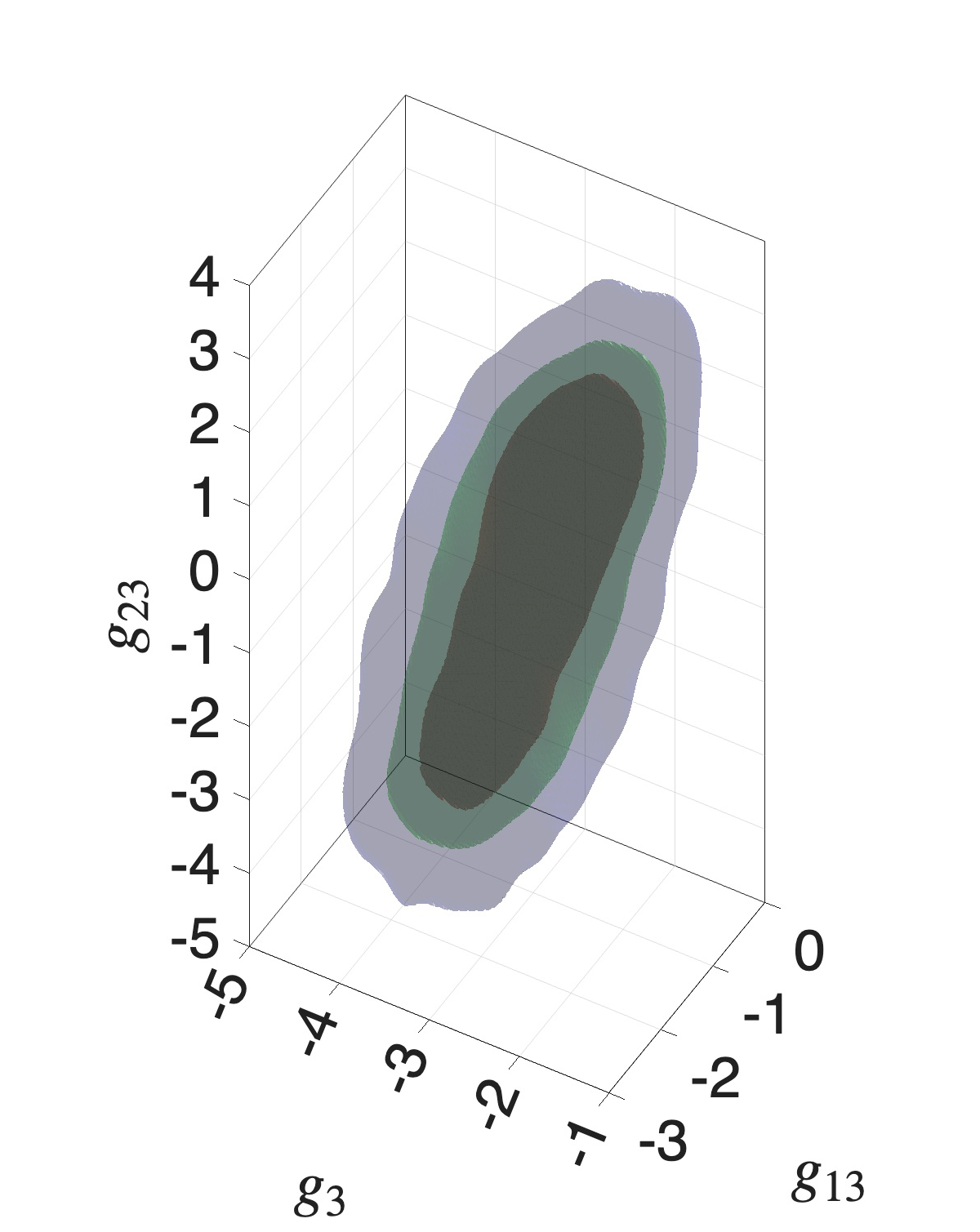}
    \end{subfigure}
    \hfill
    \begin{subfigure}[h]{0.31\linewidth}
      \centering
      \includegraphics[width=\linewidth]{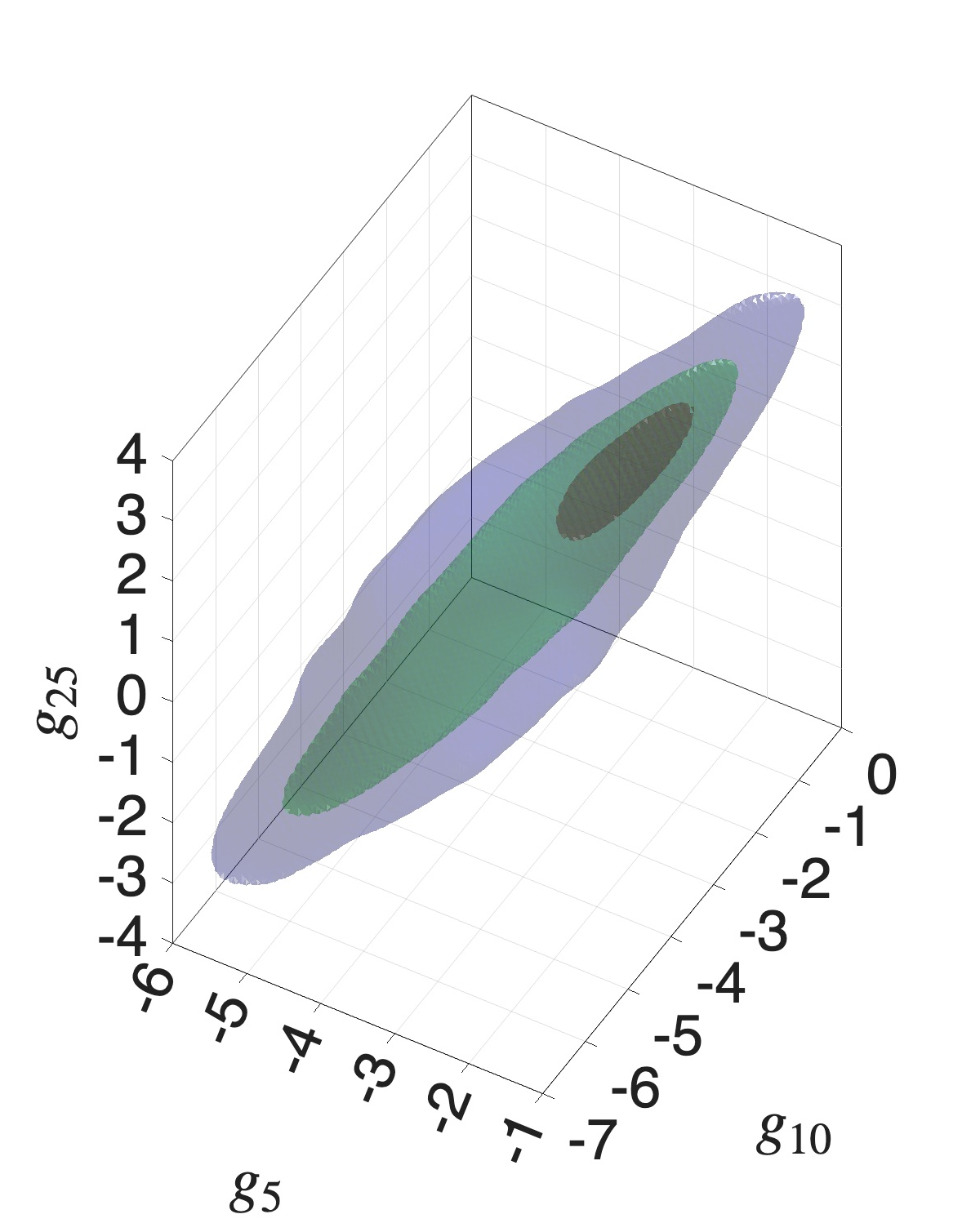}
    \end{subfigure}
    \hfill
    \begin{subfigure}[h]{0.32\linewidth}
      \centering
      \includegraphics[width=\linewidth]{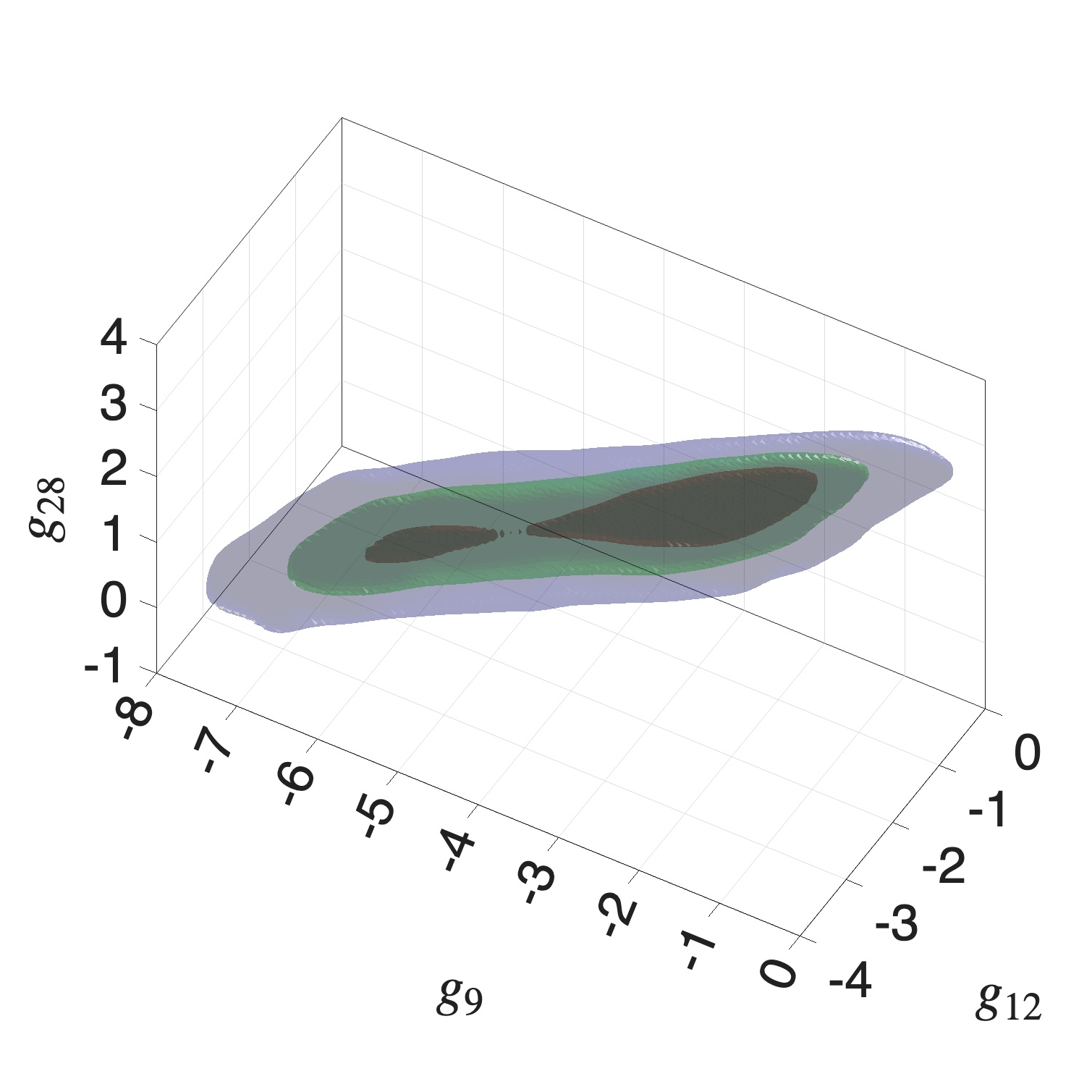}
    \end{subfigure}
    \\
    \begin{subfigure}[h]{0.32\linewidth}
      \centering
      \includegraphics[width=\linewidth]{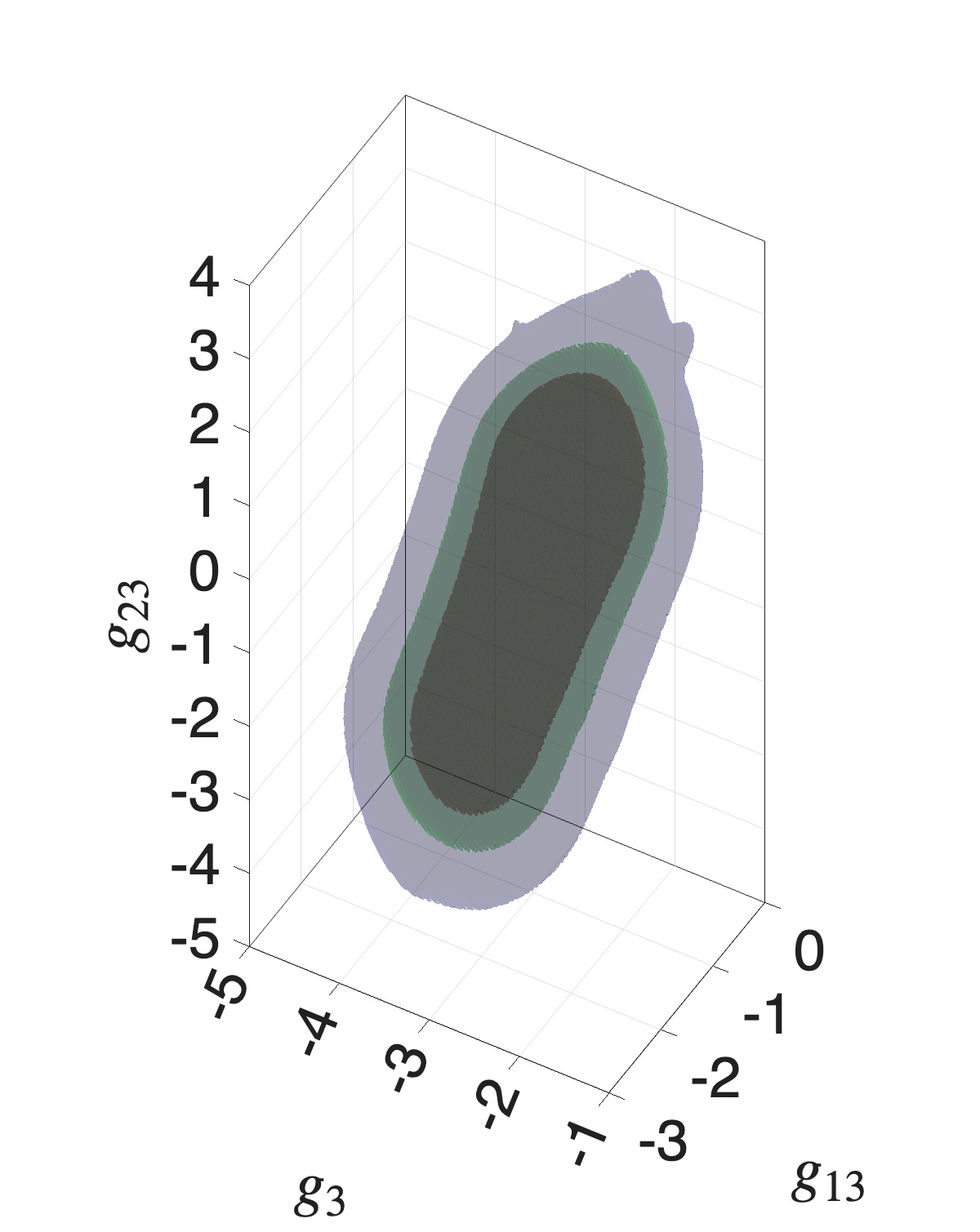}
    \end{subfigure}
    \hfill
    \begin{subfigure}[h]{0.32\linewidth}
      \centering
      \includegraphics[width=\linewidth]{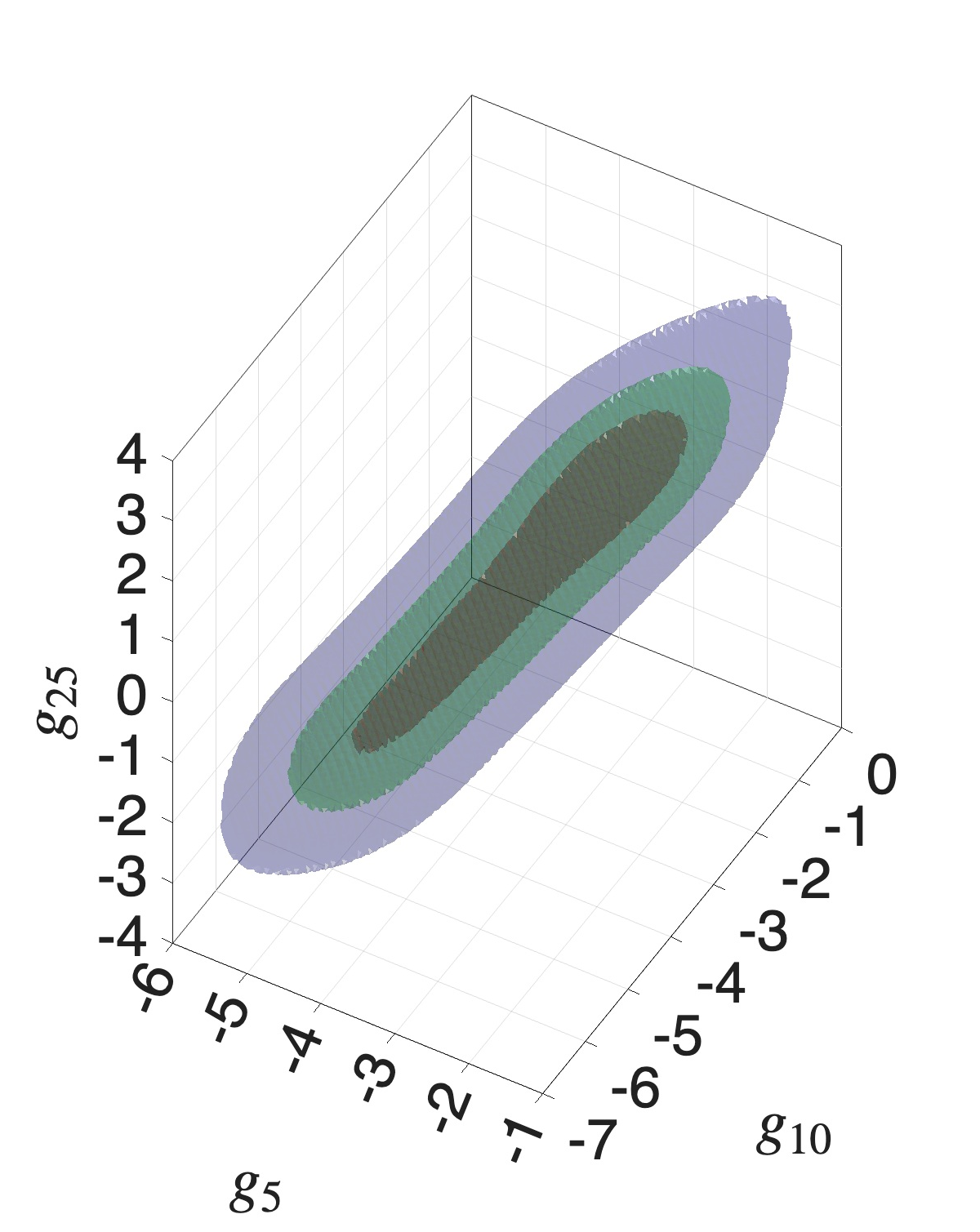}
    \end{subfigure}
    \hfill
    \begin{subfigure}[h]{0.32\linewidth}
      \centering
      \includegraphics[width=\linewidth]{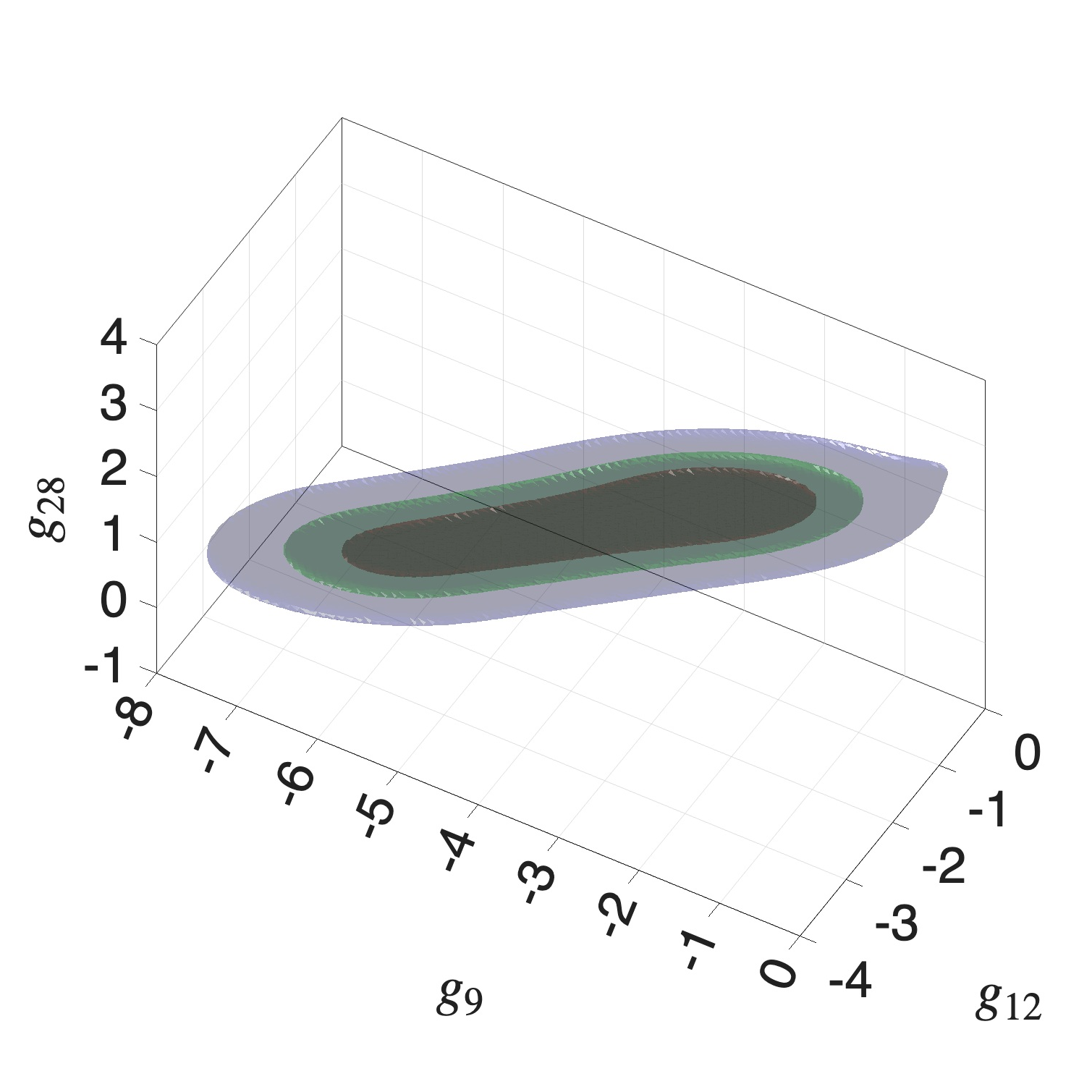}
    \end{subfigure}
    \caption{Nonlinear operator \eqref{nonlinear}. Comparison of the 3-point PDFs generated by Gaussian copula (first row) vs. the  benchmark 3-point PDFs obtained from MC (second row).  Shown are $\{0.005, 0.1, 0.4\}$ level sets. }
    \label{fig:3-point_PDF_results_Nonlinop}
  \end{figure}

\noindent 
Finally, in Figures~\ref{fig:3-point_PDF_results_Linop} and~\ref{fig:3-point_PDF_results_Nonlinop}, we demonstrate that the accuracy of the Gaussian copula approximation remains robust for three-point PDFs.

\section{Summary}
\label{sec:summary}
We presented new methods for uncertainty propagation in feed-forward neural network architectures with leaky ReLU activation functions, subject to random perturbations in the input vectors. In particular, we derived computable approximate analytical expressions for the probability density function (PDF) of the neural network output and its statistical moments.
To obtain these expressions, we introduced a suitable linearization of the leaky ReLU activation functions, which renders the formulas for both the PDF and the moments analytically tractable. We applied these formulas to feed-forward neural networks trained to approximate linear and nonlinear integro-differential operators, and found that the resulting statistical predictions (PDFs and statistical moments) remain remarkably accurate even for large perturbation amplitudes in the input vectors.
The accuracy of the PDFs and statistical moments obtained from the linearized leaky ReLU network approximation depends on how the approximation error propagates through the network layers (see Appendix~\ref{app:error}). We found that the error is proportional to the amplitude of the input perturbation—\textcolor{blue}{in agreement with the error analysis in Appendix~\ref{app:errorBounds}}—and tends to decrease as the number of network layers \( L \) increases. This trend suggests that the linearized leaky ReLU approximation becomes increasingly effective for deeper networks. 
\textcolor{blue}{
At the same time, while the proposed uncertainty quantification method demonstrates strong performance in capturing output variability for both linear and nonlinear operators, it also exhibits limitations in regimes where the linearized leaky ReLU approximation breaks down. In particular, the numerical results shown in Figure~\ref{fig:NonlinOps} and Figure~\ref{fig:NonlinOp_ErrResid} indicate a slight loss of 
accuracy at higher noise levels, especially in cases 
where input perturbations lead to significant 
changes in activation patterns.}

We also studied the accuracy of Gaussian copula models for approximating the \emph{full} joint PDF \( p(\bm{g}) \) of the neural network output. As discussed in Section~\ref{sec:copula}, a Gaussian copula can be constructed using the marginal PDFs \( p(g_j) \) and the correlation function \( \mathbb{E}\{g_i g_j\} \), both of which are accurately approximated by the analytical expressions derived in~\eqref{eq:FINAL_PDF} and~\eqref{corr}. This enables a fully analytical construction of the Gaussian copula surrogate model for the PDF of the neural network output. Our results show that this model yields highly accurate approximations of two-point and three-point PDFs, indicating that the Gaussian copula serves as an excellent surrogate for the joint distribution of the neural network output.
The methods we proposed in this work can be extended to more complex neural operator architectures, \textcolor{blue}{such as ResNets (see \ref{app:ResNets})}, Fourier Neural Operators (FNOs) \cite{FNO2021}, or DeepONets \cite{Paris2021,GK2021}. These architectures can be viewed as generalizations of the MLP considered in this study~\cite{boulle_mathematical_2023, kovachki_neural_2021}.

\section*{Acknowledgements}
\noindent
This work was supported by the U.S. Department of Energy (DOE) under grant ``Resolution-invariant deep learning for accelerated propagation of epistemic and aleatory uncertainty in multi-scale energy storage systems, and beyond,'' contract number DE-SC0024563.

\appendix
\section{Derivation of the PDF of the neural network output}
\label{app:PDF}
In this appendix, we derive an analytical expression for the probability density function (PDF) of each component of the neural network output, under the assumption that the random input vector is uniformly distributed (see~\eqref{PDF}). To simplify the calculation, we linearize the leaky ReLU activation functions around the mean of the input vector \( \bm{\mu} \) (see \eqref{P1} and \eqref{P2}).
Let us first consider the case of a neural network with a single layer. Substituting equations~\eqref{PDF} and~\eqref{P1} into~\eqref{16}, we obtain the following expression for the PDF of \( g_j \)
  \begin{align}
    p(g_j) &\simeq \Cexpmarg \Cbeta \int_{-\infty}^{\infty} \int_{\betacube} e^{ia_j\left(g_j - \bm A_1 \cdot [\neu(\Wparm_1\bm{\mu} + \bparm_1) + \bm J_1 \bm z] \right)} d\bm z da_j, \nonumber \\
    &= \Cexpmarg \Cbeta \int_{-\infty}^{\infty} e^{ia_j[g_j - \bm A_j \cdot \neu(\Wparm_1\bm{\mu} + \bparm_1)]}
    \left[\int_{\betacube}
    e^{ ia_j\bm A_j\bm{J}_1\noise}d\noise \right] da_j. \label{eq:pdf_ithComponent_splitIntegral}
  \end{align}
where $\bm A_j$ is the $j$-th row of the matrix $\bm A$  (output weights), and $\bm J_1$  is the matrix defined in \eqref{J1}. Upon definition of the $N_x$-dimensional row vector
\begin{equation}
\bm q_j=\bm A_j \bm J_1,
\end{equation}
we have 
  \begin{align}
    \int_{\betacube}
    e^{ ia_j\bm q_j\cdot\noise}d\noise
    = & \prod_{n = 1}^{N_x}\int_{-\beta}^{\beta} e^{-ia_j q_{jn}z_n}dz_n\nonumber\\
    =& \prod_{n= 1}^{N_x} 
    \frac{e^{ia_j \beta q_{jn}} -e^{-ia_j \beta q_{jn}} }{ia_jq_{jn}} \nonumber\\
     =& \prod_{n= 1}^{N_x} \frac{2\beta\sin(a_j \beta q_{jn})}{a_jq_{jn}}\nonumber\\
  =& \prod_{n= 1}^{N_x} 2\beta\text{sinc}(a_j \beta q_{jn}),  
    \label{B3}
  \end{align}
where $\text{sinc}(\cdot)$ is the sinc function. Substituting \eqref{B3} into \eqref{eq:pdf_ithComponent_splitIntegral} yields 
 \begin{align}
p(g_j) \simeq \Cexpmarg \Cbeta \int_{-\infty}^{\infty} e^{ia_j[g_j - \bm A_j \cdot \neu(\Wparm_1\bm{\mu} + \bparm_1)]} \prod_{n= 1}^{N_x} 2\beta\text{sinc}(a_j \beta q_{jn}) da_j.
 \label{eq:t}
\end{align}
Finally, by using the Fourier convolution theorem we obtain
 \begin{align}
p(g_j)  \simeq \frac{1}{2\pi}\left(\prod_{n= 1}^{N_x}\frac{2\pi}{q_{jn}}\right) \Pi_{\beta q_{j1}}[\eta(g_j)]*\cdots * \Pi_{\beta q_{jN_x}}(\eta(g_j))
 \label{eq:FINAL_PDF0}
\end{align}
where 
\begin{equation}
\eta (g_j)= g_j - \bm A_j \cdot \neu(\Wparm_1\bm{\mu} + \bparm_1),
\end{equation}
$*$ denotes a standard convolution operation, and $\Pi_{\beta q_{j1}}(\cdot)$ is the rectangular function\footnote{In equation \eqref{r}, $H(x)$ denotes the Heaviside step function.}
\begin{equation}
\Pi_{\beta q_{j1}}(x)  = H(x+\beta q_{j1})- H(x-\beta q_{j1}).
\label{r}
\end{equation}
In practice, one may compute \( p(g_j) \) using its inverse Fourier transform representation, which involves the product of $\text{sinc}()$ functions in \eqref{eq:t}, rather than through the iterated convolution of rectangular functions, which can be more computationally expensive.

\section{Error analysis for neural networks with linearized leaky ReLU activation functions}
\label{app:error}

In this appendix, we develop a thorough error analysis to quantify the effects of linearizing leaky ReLU activation functions on the neural network output. This analysis provides insight into why the approximate analytical formulas for the marginal probability density function (PDF) we obtained in Section \ref{sec:PDFapprox}, and the formulas for the statistical moments in Section \ref{sec:moments}, remain accurate even in the presence of large perturbation amplitudes in the input vectors \( \bm{f}=\bm \mu +\bm z\), i.e., large perturbation 
amplitudes in $\bm z$. \textcolor{blue}{The theoretical framework presented in this appendix is based on a linearization\footnote{\textcolor{blue}{The leaky ReLU activation function  \eqref{eq:leakyReLU} is piecewise linear, and therefore the notion of ``linearization'' should be interpreted within this context. Specifically, \eqref{eq:leakyReLU} has slope 1 for $x > 0$ and slope $\alpha$ for $x < 0$, with a discontinuity in the derivative at $x = 0$. Hence, linearizing $\phi(x)$ at $x=a$ corresponds to approximating $\phi(x)$ with the linear function $\phi'(a) x$ for all $x\in\mathbb{R}$.}} of the activation function $\phi(\cdot)$, and can be also applied to the standard ReLU activation function.}

Hereafter, we {analyze} the error behavior as it is pushed-forward into the neural net. To this end, let us first recall the linear approximation of first neural network layer, as discussed in section  \ref{sec:PDFapprox}
\begin{align}
 \neu(\Wparm_1(\mean + \noise) +\bparm_1)  \simeq \neu(\bm h_1) + 
 \bm J_1(\bm h_1)\noise,
 \label{leakyexp}
\end{align}
where 
\begin{equation}
\bm h_1 = \Wparm_1\mean  +\bparm_1,\qquad \text{and}\qquad  \bm J_1(\bm h_1) = \bm{\phi}'(\bm h_1)\boxdot\Wparm_1. 
\end{equation}
Equation \eqref{leakyexp} can be equivalently written as\footnote{In equation \eqref{leakyexp1} $\circ$ denotes the Hadamard matrix product. }
\begin{align}
 \neu(\bm h_1+ \Wparm_1\bm z)  \simeq \phi'(\bm h_1) \circ \left[\bm h_1+\bm W_1 \bm z\right].
\label{leakyexp1}
\end{align}
\textcolor{blue}{This expression is exact when the sign of \( \bm{h}_1 \) matches that of \( \bm{h}_1 + \bm{W}_1 \bm{z} \) component-wise. In such cases, the perturbation \( \bm{z} \) is not large enough to shift any component of \( \bm{h}_1 + \bm{W}_1 \bm{z} \) across the kink of the leaky ReLU function, and thus the linear expansion \eqref{leakyexp1} remains exact.}
Given the "half" linearity of the leaky ReLU activation function, it is straightforward to compute the exact residual in \eqref{leakyexp1}, i.e., the \emph{exact error} incurred by replacing \( \neu(\bm{W}_1(\bm{\mu} + \bm{z}) + \bm{b}_1) \) with the linear approximation \( \neu(\bm{h}_1) \circ \left[ \bm{h}_1 + \bm{W}_1 \bm{z} \right] \). We have

 \begin{align}
 {\phi}(\bm{h}_1 + \Wparm_1\noise) 
    = \phi'(\bm h_{1}) \circ [\bm h_1 + \Wparm_1\bm z] + \underbrace{\left(\phi'(\bm{h}_1 + \Wparm_1\noise) - \phi' (\bm h_1) \right)\circ[\bm h_1 + \Wparm_1\bm z]}_{\Delta \phi_1(\bm{h}_1)}.
    \label{DeltaPhi1}
  \end{align}
We can now proceed recursively across the neural network and quantify the 
error in the output of subsequent layers, which has two components: i) the error in the output of the previous layer passed through as input to the current layer, and error in the current layer output due to leaky ReLU linearization approximation. To quantify these errors we define recursively for $n=2, \ldots,  L$ and $\bm z_1=\bm z$ 
\begin{align}
\bm h_n = &\bm W_n \phi'(\bm h_{n-1})\circ \bm h_{n-1} + \bm b_{n}, \label{hn}\\
\bm z_n = &\phi'(\bm h_{n-1})\circ \bm W_{n-1}\bm z_{n-1}+\Delta \phi_{n-1}(\bm h_{n-1}),
\label{zn}
\end{align}
where 
\begin{equation}
\Delta \phi_{n-1}(\bm h_{n-1})= \left[\phi'(\bm h_{n-1}+\bm W_{n-1}\bm z_{n-1}) - \phi'(\bm h_{n-1})\right]\circ \left(\bm h_{n-1}+\bm W_{n-1} z_{n-1}\right).
\end{equation}
This allows us to write the exact output of the $L$-th layer as 
\begin{equation}
\phi(\bm h_L+\bm W_L \bm z_{L}) = \phi'(\bm h_L)\circ \left[\bm h_L+\bm W_L \bm z_L\right]+\Delta \phi_{L}(\bm h_{L}),
\end{equation}
where 
\begin{equation}
\Delta \phi_{L}(\bm h_{L}) = \left[\phi'(\bm h_L+\bm W_L\bm z_L) - \phi'(\bm h_L)\right]\circ \left(\bm h_L+\bm W_L z_L\right). 
\end{equation}
Correspondingly, the {\em exact} error in the neural network output layer due to the linearized leaky ReLU approximation throughout the network is 
\begin{equation}
\bm e= \left|\bm A\Delta \phi_{L}(\bm h_{L})  \right|,
\label{eq:error}
\end{equation}
where $\bm A$ is the output matrix.

\textcolor{blue}{
\subsection{Error bounds}
\label{app:errorBounds}
While the error formula \eqref{eq:error} is exact, its dependence on the number of hidden layers $L$, the number of neurons $N$, the slope parameter $\alpha$ of the activation function, weights and biases, and statistical properties of the added noise is not immediately transparent. Therefore, it is useful to derive deterministic and probabilistic upper bounds for the error that make these dependencies explicit in terms of the parameters of the neural network and input noise. 
\subsubsection{Deterministic error bound for networks with one layer}
Consider a neural network with only one layer. 
We are interested in bounding the \( i \)-th component of the error output \eqref{eq:error}
\begin{equation}
e_i = \left| \bm{A}_{i} \cdot \Delta \phi_1(\bm{h}_1) \right|,
\label{eL1}
\end{equation}
where \( \bm{A}_i \) denotes the \( i \)-th row of the output matrix \( \bm{A} \).
The linearization error due to the leaky ReLU activation function (see equation~\eqref{DeltaPhi1}) is
\begin{equation}
\Delta \phi_1(\bm{h}_1) = \left[ \phi'\left(\bm{h}_1 + \bm{W}_1 \bm{z}\right) - \phi'\left(\bm{h}_1\right) \right] \circ \left( \bm{h}_1 + \bm{W}_1 \bm{z} \right).
\label{Dphi}
\end{equation}
Let
\[
\bm{\gamma}_1 = \phi'\left(\bm{h}_1 + \bm{W}_1 \bm{z}\right) - \phi'\left(\bm{h}_1\right),
\]
so that equation~\eqref{Dphi} becomes
\[
\Delta \phi_1(\bm{h}_1) = \bm{\gamma}_1 \circ \left( \bm{h}_1 + \bm{W}_1 \bm{z} \right).
\]
Each component of \( \bm{\gamma}_1 \) satisfies \( \gamma_{1j} \in \{0, \pm(1 - \alpha)\} \), depending on whether the sign of  $h_{1j}$ flips under the perturbation $(\bm W_1\bm z)_j$, i.e., 
\[
\gamma_{1j} \neq 0 \quad \Leftrightarrow \quad \operatorname{sign}(h_{1j}) \neq \operatorname{sign}\left(h_{1j} + \left(\bm{W}_1 \bm{z}\right)_j \right).
\]
Define the flip set
\begin{equation}
\mathcal{Z}_1 = \left\{ j : \operatorname{sign}(h_{1j}) \neq \operatorname{sign}\left(h_{1j} + \left(\bm{W}_1 \bm{z}\right)_j \right) \right\},
\end{equation}
which contains the indices where a sign flip occurs.
Since only flipped entries contribute to \( \Delta \phi_1 \), we have
\begin{equation}
\left| \Delta \phi_1(h_{1j}) \right| = 
\begin{cases}
(1 - \alpha) \left| h_{1j} + \left( \bm{W}_1 \bm{z} \right)_j \right| & \text{if } j \in \mathcal{Z}_1, \\
0 & \text{otherwise}.
\end{cases}
\end{equation}
This implies
\begin{equation}
\left\| \Delta \phi_1(\bm{h}_1) \right\|_2^2 = (1 - \alpha)^2 \sum_{j \in \mathcal{Z}_1} \left| h_{1j} + \left( \bm{W}_1 \bm{z} \right)_j \right|^2.
\label{eq:dphi-norm}
\end{equation}
Using the triangle inequality,
\[
\left| h_{1j} + \left( \bm{W}_1 \bm{z} \right)_j \right| \le \left| h_{1j} \right| + \left| \left( \bm{W}_1 \bm{z} \right)_j \right|.
\]
Whenever a sign flip occurs (i.e., \( j \in \mathcal{Z}_1 \)), we necessarily have
\[
\left| \left( \bm{W}_1 \bm{z} \right)_j \right| \ge \left| h_{1j} \right|,
\]
which implies
\[
\left| h_{1j} + \left( \bm{W}_1 \bm{z} \right)_j \right| = \left| \left| \left( \bm{W}_1 \bm{z} \right)_j \right| - \left| h_{1j} \right| \right| \le \left| \left( \bm{W}_1 \bm{z} \right)_j \right|.
\]
Substituting this into equation~\eqref{eq:dphi-norm}, we obtain
\begin{align}
\left\| \Delta \phi_1(\bm{h}_1) \right\|_2^2 
&\le (1 - \alpha)^2 \sum_{j \in \mathcal{Z}_1} \left| \left( \bm{W}_1 \bm{z} \right)_j \right|^2 \nonumber \\
&\le (1 - \alpha)^2 \left\| \bm{W}_1 \bm{z} \right\|_2^2.
\label{bbo}
\end{align}
Finally, by the Cauchy--Schwarz inequality, we bound the output error \eqref{eL1} as
\begin{equation}
e_i  \le \left\| \bm{A}_i \right\|_2  \left\| \Delta \phi_1(\bm{h}_1) \right\|_2.
\label{cs}
\end{equation}
Substituting \eqref{bbo} into \eqref{cs} yields
\begin{equation}
e_i \le (1 - \alpha) \left\| \bm{A}_i \right\|_2 \left\| \bm{W}_1 \bm{z} \right\|_2.
\end{equation}
Since \( \bm{z} \sim \mathcal{U}[-\beta, \beta]^{N_x} \) (uniform with i.i.d. components), we have
\begin{equation}
\left\| \bm{z} \right\|_2 \le \sqrt{N_x} \beta.
\label{zb}
\end{equation}
Combining these estimates yields the final error bound
\begin{equation}
e_i \le (1 - \alpha)\sqrt{N_x} \left\| \bm{A}_i \right\|_2 \left\| \bm{W}_1 \right\|_2   \beta.
\label{deb}
\end{equation}
This bound depends explicitly on the network parameters and the perturbation amplitude \( \beta \), and satisfies \( e_i = 0 \) when \( \bm{z} = \bm{0} \), as expected.
\subsubsection{Deterministic error bound for networks with $L$ layers}
The deterministic error bound \eqref{deb} can be extended 
to neural networks with \( L \) layers. To this end, consider the \( i \)-th 
component of the output error \eqref{eq:error}
\begin{equation}
e_i = \left| \bm{A}_i \cdot \Delta \phi_L(\bm{h}_L) \right|,
\end{equation}
where the residual at the output layer is given by
\[
\Delta \phi_L(\bm{h}_L) = \left[ \phi'\left( \bm{h}_L + \bm{W}_L \bm{z}_L \right) - \phi'\left( \bm{h}_L \right) \right] \circ \left( \bm{h}_L + \bm{W}_L \bm{z}_L \right).
\]
Applying the Cauchy--Schwarz inequality, we obtain
\begin{equation}
e_i \leq \left\| \bm{A}_i \right\|_2 \left\| \Delta \phi_L(\bm{h}_L) \right\|_2.
\label{eLay}
\end{equation}
To estimate the $2$-norm of $\Delta \phi_L(\bm{h}_L)$, we introduce the flip sets
\begin{equation}
\mathcal{Z}_q = \left\{ j : \operatorname{sign}(h_{qj}) \neq \operatorname{sign}\left( h_{qj} + \left( \bm{W}_q \bm{z}_q \right)_j \right) \right\}, \qquad q = 1, \ldots, L.
\end{equation}
As in the single-layer case, the absolute value of each component of the residual $\Delta \phi_L(\bm{h}_L)$ is either zero or equal to $(1 - \alpha) \left| h_{Lj} + (\bm{W}_L \bm{z}_L)_j \right|$, depending on whether the index $j$ belongs to the flip set $\mathcal{Z}_L$. In the worst-case scenario, where a sign flip occurs in every  component, we obtain 
\begin{align}
\left\| \Delta \phi_L(\bm{h}_L) \right\|_2 
&\le (1 - \alpha) \left\| \bm{W}_L \bm{z}_L \right\|_2 \nonumber \\
&\le (1 - \alpha) \left\| \bm{W}_L \right\|_2 \left\| \bm{z}_L \right\|_2.
\label{dphi}
\end{align}
We now derive a bound for \( \left\| \bm{z}_L \right\|_2 \). 
From the recursive expression (see Eq.~\eqref{zn}),
\[
\bm{z}_{n+1} = \phi'\left( \bm{h}_n \right) \circ \left( \bm{W}_n \bm{z}_n \right) + \Delta \phi_n(\bm{h}_n),
\]
and using the bound on \( \left\| \Delta \phi_n(\bm{h}_n) \right\|_2 \) in Eq.~\eqref{dphi}, 
we obtain
\begin{align}
\left\| \bm{z}_{n+1} \right\|_2 
&\le \left\| \phi'\left( \bm{h}_n \right) \right\|_\infty \left\| \bm{W}_n \right\|_2 \left\| \bm{z}_n \right\|_2 
+ \left\| \Delta \phi_n(\bm{h}_n) \right\|_2 \nonumber \\
&\le \left( \left\| \phi'\left( \bm{h}_n \right) \right\|_\infty + (1 - \alpha) \right) \left\| \bm{W}_n \right\|_2 \left\| \bm{z}_n \right\|_2.
\label{rec}
\end{align}
Next, define the coefficient\footnote{\textcolor{blue}{Recall that 
\[
\left\| \phi'\left( \bm{h}_n \right) \right\|_\infty = \max_{j = 1, \ldots, N} \left| \phi'\left( h_{nj} \right) \right|,
\]
so that \( \left\| \phi'\left( \bm{h}_n \right) \right\|_\infty\) is equal to either $1$ or $\alpha$ depending on whether any component of \( \bm{h}_n \) is positive.}}
\begin{equation}
\lambda_n = \left( \left\| \phi'\left( \bm{h}_n \right) \right\|_\infty + (1 - \alpha) \right) \left\| \bm{W}_n \right\|_2.
\end{equation}
Using this definition and the inequality \eqref{rec} we obtain
\begin{equation}
\left\| \bm{z}_L \right\|_2 \le \left( \prod_{n = 1}^{L - 1} \lambda_n \right) \left\| \bm{z}_1 \right\|_2.
\label{Lrecursion}
\end{equation}
Inserting the last inequality into \eqref{eLay} and using \eqref{zb}, 
we obtain the final error bound
\begin{equation}
e_i \le (1 - \alpha) \sqrt{N_x} \left\| \bm{A}_i \right\|_2 \left\| \bm{W}_L \right\|_2 \left( \prod_{j = 1}^{L - 1} \lambda_j \right) \beta, \qquad L\geq 2.
\label{boundL}
\end{equation}
This bound depends on the network parameters and the perturbation amplitude \( \beta \), and it satisfies \( e_i = 0 \) when \( \bm{z} = \bm{0} \), as expected. 
}

\textcolor{blue}{
\subsubsection{Probabilistic error bound for networks with $L$ layers}
In this section, we establish a probabilistic upper bound on the components of the output error \eqref{eq:error} for a neural network with $L$ layers. The analysis begins with the deterministic inequality
\begin{equation}
e^2_i \le K^2 \left\| \bm{z}_1 \right\|^2_2, 
\qquad \text{where} \quad 
K = (1 - \alpha) \left\| \bm{A}_i \right\|_2 \left\| \bm{W}_L \right\|_2 \prod_{n = 1}^{L - 1} \lambda_n,
\label{eq:prob_Kbound}
\end{equation}
which follows directly from the output error bound \eqref{eLay}, the residual 
estimate \eqref{dphi}, and the recursive norm bound \eqref{Lrecursion}.
Since \( \bm{z}_1 = \bm{z} \sim \mathcal{U}[-\beta, \beta]^{N_x} \) with i.i.d.\ components, the distribution of \( \left\| \bm{z}_1 \right\|_2^2 \) can, in principle, be computed explicitly. This would allow us to evaluate the exact probability that \( \left\| \bm{z}_1 \right\|_2^2 \) is below a prescribed threshold, thus yielding a probabilistic bound on the output error.
In this section, we instead pursue a different route based on Bernstein-type concentration inequalities for Lipschitz functions of bounded independent variables, following the framework presented in \cite{Boucheron2016}. Such concentration inequalities provide bounds on the probability that \( \left\| \bm{z}_1 \right\|_2^2 \) deviates from its expected value by a specified amount \( t > 0 \). We begin by computing
\begin{equation}
\mathbb{E}\left[\left\| \bm{z}_1 \right\|^2_2\right] =  N_x\frac{\beta^2}{3} \qquad\text{and}\qquad  
\mathrm{Var}\left[\left\| \bm{z}_1 \right\|^2_2\right] = N_x \frac{4\beta^4}{45}.
\end{equation}
Using Bernstein's inequality for sums of bounded independent random variables 
(see \cite[Corollary 2.11]{Boucheron2016}), we obtain the following 
probabilistic bound 
\begin{equation}
\mathbb{P} \left( \left\| \bm{z}_1 \right\|_2^2 - \frac{N_x \beta^2}{3} \ge t \right) 
\le \exp\left(- \frac{t^2}{\displaystyle \frac{8 N_x \beta^4}{45} + \frac{2N_x}{3} \beta^2 t} \right)\qquad \quad \text{for all}\quad  t>0.
\label{eq:bernstein}
\end{equation}
Letting the right-hand side equal \( \delta \in (0,1) \), we 
solve for \( t \) and obtain with probability at least \( 1 - \delta \),
\begin{equation}
\left\| \bm{z}_1 \right\|_2^2 \le \frac{N_x \beta^2}{3} + \beta^2 \left( \sqrt{ \frac{8 N_x^2}{45} \log\left( \frac{1}{\delta} \right) } + \frac{2 N_x}{3} \log\left( \frac{1}{\delta} \right) \right).
\label{eq:z1_bound}
\end{equation}
Inserting this bound into the deterministic inequality \eqref{eq:prob_Kbound}, we obtain the probabilistic error bound
\begin{equation}
e_i^2 
\le K^2 \beta^2 \left( \frac{N_x}{3} 
+ \sqrt{ \frac{8 N_x^2}{45} \log\left( \frac{1}{\delta} \right) }
+ \frac{2 N_x}{3} \log\left( \frac{1}{\delta} \right) \right),
\label{eq:final_prob_bound}
\end{equation}
with probability at least \( 1 - \delta \), where \( K \) is defined in \eqref{eq:prob_Kbound}.
This bound improves upon the deterministic bound by capturing typical behavior (rather than worst-case) of the input perturbation norm.
For example, setting \(\delta = 1/2\) (i.e., $50\%$ probability) in the Bernstein inequality \eqref{eq:final_prob_bound}, we obtain
\begin{align} 
e_i^2 &\le K^2 \beta^2 \left( \frac{N_x}{3} 
+ \sqrt{ \frac{8 N_x^2}{45} \cdot \log(2) }
+ \frac{2 N_x}{3} \log(2) \right) \\
&\approx K^2 \beta^2 \left( \frac{N_x}{3} + 0.35 \sqrt{N_x} + 0.462 \right).
\label{eq:prob_bound_delta_half}
\end{align}
This result shows that, even with modest confidence (\(\delta = 0.5\)), the probabilistic bound \eqref{eq:final_prob_bound} improves upon the worst-case deterministic estimate \( e_i^2 \le K^2 \beta^2 N_x \), which holds with probability one. 
Note that as $\delta\rightarrow 0$, the bound in \eqref{eq:final_prob_bound} goes to infinity, even if we know $ K^2 \beta^2 N_x$ holds with probability one, i.e., for every realization of $\bm z_1$. This behavior is typical of concentration inequalities, which trade off tightness of the bound for high-probability guarantees. The smaller the failure probability \( \delta \), the looser the bound becomes, even when the underlying variable is uniformly bounded.}

\textcolor{blue}{
\section{Application to ResNets}
\label{app:ResNets}
In this section, we apply the forward uncertainty propagation framework developed in this paper to a neural network architecture with skip connections, specifically ResNet \cite{ResNets}. Our goal is to derive analytical expressions for the PDF and the statistical moments of the neural network output by leveraging a linear approximation of the leaky ReLU activation function.
The input-output mapping of a ResNet architecture with $L$ layers is given by
\begin{equation}
  \bm g = \bm f + 
\bm A \phi \left( \bm{W}_L \phi\left( \bm{W}_{L-1}\phi\left( \cdots \phi\left( \bm{W}_1 \bm f + \bm{b}_1 \right) \cdots\right) + \bm{b}_{L-1} \right) + \bm{b}_L \right).  \label{def:resnet_nn}
\end{equation}
To approximate the output distribution and its statistical moments, 
we introduce the linearized leaky ReLU model described in Section \ref{sec:linerizedLeakyReLU}, along with the representation~\eqref{f} of the random input vector $\bm f$. Under this approximation, the ResNet output \eqref{def:resnet_nn} can be written as
\begin{equation}
  \bm g \simeq \bm A \bm R_L + \bm A \bm J_L \bm z + \bm \mu + \bm z,
  \label{eq:linearized_resnet_output}
\end{equation}
where $\bm J_L$ is given in \eqref{JL}, and $\bm R_L$ defined as
\begin{equation}
  \bm R_L = \phi\left( \bm{W}_L \, \phi\left( \bm{W}_{L-1} \, \phi\left( \cdots \phi\left( \bm{W}_1 \bm \mu + \bm{b}_1 \right) \cdots \right) + \bm{b}_{L-1} \right) + \bm{b}_L \right).
\end{equation}
Following the same derivation as in \ref{app:PDF}, it is straightforward to obtain an analytical expression for the PDF of each component of the output vector \( \bm{g} \) defined in~\eqref{eq:linearized_resnet_output}. We have 
  \begin{equation}
 p(g_j) = \frac{1}{2\pi}\frac{1}{(2\beta)^{N_x}} \int_{-\infty}^{\infty}  e^{i a_j (g_j - ( \bm A_j \cdot \bm R_L + \mu_j) ) }
 \left[\int_{ [-\beta, \beta]^{N_x}} e^{- i a_j \bm A_{j}  \bm J_L \bm z + z_j} d\bm z \right] d a_j. \label{eq:marginalUnifPDF}
  \end{equation}
Upon definition of
\begin{equation}
  \tilde{\bm q}_j = \bm A_j \bm J_L + \bm e_j^T,
  \label{tildeq}
\end{equation}
where $\bm e^T_j$ is the $j$-th row of a $N_x\times N_x$ identity matrix, 
we can simplify the integral in \eqref{eq:marginalUnifPDF} as 
  \begin{align}
    \int_{[-\beta, \beta]^{N_x}} e^{- i a_j \tilde{\bm q}_j \bm z} d\bm z = & \prod_{n = 1}^{N_x}\int_{-\beta}^{\beta} e^{-ia_j \tilde{q}_{jn}z_n}dz_n\nonumber\\
      =& \prod_{n= 1}^{N_x} 2\beta_n\text{sinc}(a_j \beta_n \tilde{q}_{jn}).
      \label{C7}
  \end{align}
A substitution of \eqref{C7} into \eqref{eq:marginalUnifPDF} yields 
 \begin{align}
p(g_j) = \frac{1}{2\pi}\left(\prod_{n= 1}^{N_x}\frac{2\pi}{\tilde{q}_{jn}}\right) \Pi_{\beta_n \tilde{q}_{j1}}[\tilde\eta(g_j)]*\cdots * \Pi_{\beta_n \tilde{q}_{jN_x}}(\tilde\eta(g_j)),
\end{align}
where $ \Pi_{\beta_n \tilde{q}_{jN_x}}(\cdot)$  is defined in \eqref{r} and  
\begin{equation}
\tilde{\eta} (g_j)= g_j - \left(\bm A_j \cdot \bm R_L + \mu_j\right).
\end{equation}
The (approximate) statistical moments of the output $\bm g$ can be readily computed using the linearized expression~\eqref{eq:linearized_resnet_output} and the PDF of the perturbation \eqref{PDF}, following the same approach outlined in Section~\ref{sec:moments}. We obtain,
\begin{align}
  \mathbb{E}\{g_j\} = &\bm A_j \cdot \bm R_L + \mu_j, \label{eq:mean_ResNet}\\
  \operatorname{Var}(g_j) =& \frac{1}{3} (\bm \beta \circ \tilde{\bm q}_j) \cdot (\bm \beta \circ \tilde{\bm q}_j),\label{eq:var_ResNet} \\
 \operatorname{Cov}(g_i, g_j)  = &\frac{1}{3}  (\bm \beta \circ \tilde{\bm q}_i)\cdot\left(\bm \beta\circ \tilde{\bm q}_j \right), \label{eq:cov_ResNet}
\end{align}
where $\bm \beta=(\bm \beta_1,\ldots, \bm \beta_{N_x})$, $\tilde{\bm q}_j$ is defined in \eqref{tildeq}, and $\circ$ denotes the Hadamard product.
}

\bibliographystyle{plain}
\bibliography{UQNN}

\end{document}